\documentclass[a4paper,fleqn]{cas-sc}

\usepackage[latin9]{inputenc}
\usepackage{url}
\usepackage{amsmath}
\usepackage{graphicx}
\usepackage[authoryear,longnamesfirst]{natbib}

\usepackage{units}
\usepackage{textcomp}
\usepackage{mathrsfs}
\usepackage[autostyle=true]{csquotes} 
 
\makeatletter

\DeclareTextSymbolDefault{\textquotedbl}{T1}
\providecommand{\tabularnewline}{\\}

\def\tsc#1{\csdef{#1}{\textsc{\lowercase{#1}}\xspace}}
\tsc{WGM}
\tsc{QE}
\tsc{EP}
\tsc{PMS}
\tsc{BEC}
\tsc{DE}
\makeatother

\begin{document}

\let\WriteBookmarks\relax
\def\floatpagepagefraction{1}
\def\textpagefraction{.001}
\shorttitle{Study and improvement of search algorithms in two-players perfect information games}
\shortauthors{Q. Cohen-Solal}
%\begin{frontmatter}

\title [mode = title]{Study and improvement of search algorithms in two-players perfect information games}                      
\tnotemark[1]%,2

\tnotetext[1]{
This work was supported in part by the French government under management of Agence Nationale de la Recherche as part of the "Investissements d'avenir" program, benchmark ANR19-P3IA-0001 (PRAIRIE 3IA Institute).
This project was provided with computer and storage resources by GENCI at CINES and IDRIS thanks to the grant 20XX-A0161011461 on the supercomputer Jean Zay and Adastra (V100 and MI2050x partitions). }

%\tnotetext[2]{The second title footnote which is a longer text matter
%   to fill through the whole text width and overflow into
%   another line in the footnotes area of the first page.}

\author[1]{Quentin Cohen-Solal}[type=editor,
                        auid=000,bioid=1,
                        %prefix=Sir,
                        role=Researcher,
                        orcid=0000-0003-4828-2708]
\cormark[1]
%\fnmark[1]
\ead{quentin.cohen-solal@dauphine.psl.eu}
%\ead[url]{www.jkkrishnan.in}

\credit{All}

\affiliation[1]{organization={LAMSADE, Université Paris Dauphine - PSL, CNRS},
                addressline={Place de Lattre de Tassigny}, 
                city={Paris},
%               citysep={}, % Uncomment if no comma needed between city and postcode
                postcode={75016}, 
                %state={Kerala},
                country={France}}

\cortext[cor1]{Corresponding author}
%\cortext[cor2]{Principal corresponding author}
%\fntext[fn1]{This is the first author footnote, but is common to third
%  author as well.}
% \fntext[fn2]{Another author footnote, this is a very long footnote and
%   it should be a really long footnote. But this footnote is not yet
%   sufficiently long enough to make two lines of footnote text.}

%\nonumnote{This note has no numbers. In this work we demonstrate $a_b$
%  the formation Y\_1 of a new type of polariton on the interface
%  between a cuprous oxide slab and a polystyrene micro-sphere placed
%  on the slab.
%  }

\begin{abstract}
Games, in their mathematical sense, are everywhere (game industries,
economics, defense, education, chemistry, biology, ...). Search algorithms
in games are artificial intelligence methods for playing such games.
Unfortunately, there is no study on these algorithms that evaluates
the generality of their performance. We propose to address this gap
in the case of two-player zero-sum games with perfect information.
Furthermore, we propose a new search algorithm and we show that, for
a short search time, it outperforms all studied algorithms on all
games in this large experiment and that, for a medium search time,
it outperforms all studied algorithms on $17$ of the $22$ studied
games. 
\end{abstract}
%\begin{graphicalabstract}
%\includegraphics{figs/cas-grabs.pdf}
%\end{graphicalabstract}

\begin{highlights} 

\item Broad experimental comparison on search algorithms in games. 

\item A new variant of a game search algorithm surpassing all previous approaches for a short search time and having the best average performance for a medium search time.

\end{highlights}


\maketitle

\noindent\begin{minipage}[t]{1\columnwidth}%
\global\long\def\tmax{t_{\mathrm{max}}}%

\global\long\def\negation{\mathop{\neg}}%

\global\long\def\Max#1#2{{\displaystyle \mathop{\mathrm{max}}_{#2}\left(#1\right)}}%
\global\long\def\max#1{\mathrm{max}\left(#1\right)}%

\global\long\def\Min#1#2{{\displaystyle \mathop{\mathrm{min}}_{#2}\left(#1\right)}}%
\global\long\def\min#1{\mathrm{min}\left(#1\right)}%

\global\long\def\rollout#1{\mathrm{rollout}\left(#1\right)}%

\global\long\def\et{\ \wedge\ }%

\global\long\def\ou{\ \vee\ }%

\global\long\def\terminal{\mathrm{t}}%

\global\long\def\alphabeta{\alpha\beta}%

\global\long\def\pvs#1{\mathrm{PVS}^{#1}}%

\global\long\def\mtdf#1{\mathrm{MTD(f)}^{#1}}%

\global\long\def\mctsh#1{\mathrm{MCTS}_{\mathrm{h}}^{#1}}%

\global\long\def\mcts#1{\mathrm{MCTS}^{#1}}%

\global\long\def\mctsip#1#2{\mathrm{MCTS}_{\gamma=#2}^{#1}}%

\global\long\def\mctsipm#1{\mathrm{MCTS}_{\alpha\beta}^{#1}}%

\global\long\def\mctsic#1{\mathrm{MCTS}_{\mathrm{max}}^{#1}}%

\global\long\def\joueur{\mathrm{j}}%

\global\long\def\joueurUn{\mathrm{1}}%

\global\long\def\joueurDeux{\mathrm{2}}%

\global\long\def\fbin{\mathrm{f_{b}}}%

\global\long\def\Actions#1{\mathrm{actions}\left(#1\right)}%

\global\long\def\zip{\mathcal{\mathrm{zip}}}%

\global\long\def\etats{\mathcal{S}}%

\global\long\def\mv#1#2{v_{#1,#2}}%

\global\long\def\bv#1#2{v_{#1,#2}}%

\global\long\def\select#1#2{n_{#1,#2}}%

\global\long\def\sselect#1{n_{#1}}%

\global\long\def\sign#1#2{\overline{#1}^{#2}}%

\global\long\def\vsign#1#2{\overline{\mv{#1}{#2}}}%

\global\long\def\bvsign#1#2{\ddddot{\bv{#1}{#2}}}%

\global\long\def\vinf#1#2{v_{#1,#2}^{-}}%

\global\long\def\vinfsign#1#2{\overline{\vinf{#1}{#2}}}%

\global\long\def\vsup#1#2{v_{#1,#2}^{+}}%

\global\long\def\vsupsign#1#2{\overline{\vsup{#1}{#2}}}%

\global\long\def\vsinf#1{v_{#1}^{-}}%

\global\long\def\vssup#1{v_{#1}^{+}}%

\global\long\def\vresol#1#2{c_{#1,#2}}%

\global\long\def\svresol#1{c_{#1}}%

\global\long\def\resol#1#2{r_{#1,#2}}%

\global\long\def\sresol#1{r_{#1}}%

\global\long\def\vresolsign#1#2{\overline{\vresol{#1}{#2}}}%

\global\long\def\svresolsign#1{\overline{\svresol{#1}}}%

\global\long\def\mtdflower{f_{-}}%

\global\long\def\mtdfupper{f_{+}}%

\global\long\def\terminalrandom{\mathrm{t_{r}}}%
\global\long\def\hfb{\mathrm{b_{t}}}%
\global\long\def\hfp{\mathrm{p_{t}}}%
\global\long\def\hadapt{f_{\theta}}%
\global\long\def\hadaptnum#1{f_{\theta_{#1}}}%
\global\long\def\itreeset{\mathbf{T_{i}}}%
\global\long\def\treeset{\mathbf{T}}%
\global\long\def\rootset{\mathbf{R}}%
\global\long\def\hterminal{f_{\mathrm{t}}}%
\global\long\def\rnap{\mathrm{p_{rna}}}%
\global\long\def\rnab{\mathrm{b_{rna}}}%
\global\long\def\ubfm{\mathrm{UBFM}}%
\global\long\def\ubfmt{\ubfm_{\mathrm{s}}}%
\global\long\def\argmax{\operatorname*{\mathrm{arg\,max}}}%
\global\long\def\ubfmref{\mathrm{UBFM}_{\mathrm{ref}}}%
\global\long\def\ubfmc{\mathrm{UBFM}_{\mathrm{c}}}%
\global\long\def\ubfms{\mathrm{UBFM}_{\mathrm{s}}}%
\global\long\def\ubfmcs{\mathrm{UBFM}_{\mathrm{c-s}}}%
\global\long\def\argmin{\operatorname*{\mathrm{arg\,min}}}%
\global\long\def\liste#1#2{\left\{  #1\,|\,#2\right\}  }%
\global\long\def\fterminal{\hterminal}%
\global\long\def\fadapt{\hadapt}%
\global\long\def\id{\mathrm{ID}}%
\global\long\def\minimum{\operatorname*{\mathrm{min}}}%

\global\long\def\pdiscret{\delta}%

\global\long\def\ended#1{\mathrm{ended}\left(#1\right)}%

\global\long\def\corum{\mathrm{CORUM}}%

\global\long\def\corsum{\mathrm{CORSUM}}%

\global\long\def\firstplayer#1{\mathrm{first\_player\left(#1\right)}}%
\global\long\def\orsum{\mathrm{ORSUM}}%
\end{minipage}

\section{Introduction}

Games have numerous applications, far beyond the obvious ones (the
video game and board game industries) and the slightly less obvious
ones (economics, defense, and also education through serious games).
In fact, all computational problems can naturally be reformulated
in terms of games. Game search algorithms are therefore general-purpose
artificial intelligence techniques for problem-solving. For instance,
the AlphaZero algorithm \citep{silver2018general}, which is a learning
and search algorithm for two-player perfect-information games, has
achieved state-of-the-art performance in numerous fields. For example,
AlphaZero's application has led to the discovery of a more efficient
matrix multiplication algorithm \citep{fawzi2022discovering} as well
as a more efficient sorting algorithm \citep{mankowitz2023faster}.
These two examples involve reformulations into single-player games,
on which two-player game algorithms apply. However, some problems
can even be reformulated as strictly two-player games (those possessing
the structure of zero-sum two-player perfect information games). Examples
include manufacturing problems, such as molecule retrosynthesis \citep{guo2024retrosynthesis}
, and theorem proving \citep{kaliszyk2018reinforcement}.

Numerous search algorithms have been proposed and studied. However,
these studies often compare only a very small number of search algorithms
on a very limited set of games. So we don't know which algorithm is
the best in general (if it exists). We don't even have general relative
results such as \enquote{this algorithm is better in general than
this other}. Moreover, the search algorithm comparisons typically
rely on a single heuristic evaluation function per game. This raises
questions about the generality of such studies: using a stronger heuristic
evaluation function could potentially alter the results.

Furthermore, the recent development of reinforcement learning methods
without human knowledge, capable of achieving very high levels of
gameplay, has significantly shifted the landscape. This happened thanks
to the irruption of AlphaZero, but also thanks to Athénan \citep{2020learning},
a more recent framework whose learning performance is superior to
AlphaZero \citep{cohen2023minimax}. A few years ago, in board game
AI tournaments, such as the Computer Olympiad, methods based on partially
ad hoc and highly specialized techniques were champion algorithms.
They are now outperformed and replaced by general-purpose algorithms
(primarily AlphaZero and Athénan). In fact, Athénan is henceforth
the best algorithm having participated in the Computer Olympiad. No
other program has exceeded 5 gold medals in the same year. While Athénan
won 5 gold medals in its first participation in 2020, and much more
in the years after: 11 in 2021, 5 in 2022, 16 in 2023, 11 in 2024
(plus 6 previous uncontested gold medals) \citep{cohen2021descent,cohen2023athenan}.
Faced with this paradigm shift, one may wonder whether the weak and/or
partial results of the literature remain unchanged. Indeed, the use
of higher-level game state evaluation functions, coming from reinforcement
learning without human knowledge, could change the balance of power
of the different algorithms proposed until now.

In addition, we are interested in algorithms that can be applied directly.
Indeed, many algorithms in the literature require tedious and costly
tuning phases, complicating or even limiting their applications. Some
of these algorithms even require expensive human research to be applied.
Ideally, a good search algorithm should have no parameter. If this
is not possible, it must have default parameters whose optimization
would be negligible, at least to remain more efficient than other
algorithms.

We thus conduct an extensive experimental study on the main search
algorithms for two-player, perfect-information, zero-sum games. The
selected algorithms are evaluated on 22 games, using around fifty
heuristic evaluation functions per game, to ensure statistically robust
results and maintaining a high level of generality. Furthermore, we
propose an improvement to one of the studied algorithms, namely Unbounded
Minimax \citep{2020learning,korf1996best}, the search algorithm of
Athénan, and experimentally demonstrate that this variant is the most
effective algorithm among all those evaluated and for all tested games
for a short search time and that it is the algorithm with the best
average performance for a medium search time.

\section{Generality about search algorithms}

Game search algorithms are used to determine the best possible strategy
within the limits of available resources. For this, these algorithms
explore the different possible continuations of a game match. This
exploration consists in constructing a game tree, that is, a tree
where the nodes are the game states and a state $c$ is a child of
another state $s$ if there is an action $a$ in $s$ allowing to
move from $s$ to $c$. It is generally not possible to construct
the complete game tree. In Chess, there are approximately $10^{120}$
possible matches. A search algorithm will therefore build a partial
game tree. The states of the game tree are labeled by values indicating
how interesting a state is. For example, end-game states are associated
with a value indicating the winner ($1$: first player wins, $0$:
draw, $-1$: first player loses). Since at the beginning of the match,
partial game trees do not contain endgame states and their scores,
a search algorithm therefore generally uses an evaluation function
for evaluating non-endgame states. These evaluation functions can
be constructed manually or learned automatically by an additional
algorithm.

\section{Improved Unbounded Minimax}

\subsection{Unbounded Minimax algorithm }

Unbounded (Best-First) Minimax ($\ubfm$) is a variant of the basic
search algorithm in two-player zero-sum perfect information games,
called Minimax. The Minimax algorithm anticipates $d$ rounds in advance
all possible game developments from the current state. It evaluates
the states encountered by assuming that the first player seeks to
maximize his score and that the second player seeks to minimize his
opponent's score. It is not possible in practice to anticipate all
possible moves $d$ rounds in advance for a large $d$ since the number
of game states to analyze evolves exponentially as a function of $d$.
Unbounded Minimax ($\ubfm$) corrects this defect since it performs
a non-uniform search in best-first order. It does not explore all
game states, only those it considers most important, which allows
it to perform very deep searches locally. In fact, it is not even
parameterized by the number of rounds to analyze: it extends the sequences
of possible next actions in best-first order,
and can stop its search at any time % \citep{2020learning}
 (see Section \ref{subsec:Minimax-non-born=00003D0000E9-de-r=00003D0000E9f=00003D0000E9rence} for details).

\subsection{Unbounded Minimax with safe decision }

We now describe the variant of the Unbounded Minimax algorithm that
we propose in this article, that we denote by $\ubfms$. It  consists
in combining the search of the classic $\ubfm$ with that we call
the \emph{safe decision strategy.} With this strategy, after performing
the same search of $\ubfm$, the action to be played is different.
In other words, this variant change the selection criterion of the
action to play after the search. With $\ubfm$, the played action 
is the action leading to the child state of best value. With the safe decision strategy, the played action
 is the \emph{safest action}, that is to say the most
played action during the search. Said differently, the action that
is chosen after the search is the one that maximizes the number of
selection of this action from the root. This algorithm is described
in Figure~\ref{alg:safe-Unbounded-Best-First-Mininmax}.

Note that the safest action is the action for which we have the best visibility concerning
the possible futures (it is the action whose subtree was the most
extensive).

Note also that $\mcts{}$ has the same decision strategy.

Remark that, since this algorithm explores in best value first, the safest action, i.e. the most played action,
is therefore the one which was most often the best during the search. 
This decision thus performs a kind of vote. This decision criterion is therefore even more relevant than in the case of $\mcts{}$.

\section{Main search algorithms of Literature}

The Minimax algorithm with Alpha-Beta pruning \citep{knuth1975analysis},
called more shortly Alpha-Beta or $\alphabeta$, is an improvement
of the Minimax algorithm that reduces the search space. For this,
it avoids exploring states that do not influence the value of the
root state of the tree of possible futures of the match (by using
the properties of min and max). More precisely, the algorithm has
two bounds $\alpha$ and $\beta$ that define the interval in which
the value of the root is searched. This interval narrows as new states
are analyzed during the search. All actions leading to states whose
value is not in the interval are pruned. It is an exact pruning, i.e.
it determines the same strategy as Minimax without this pruning rule,
but thanks to it the computation is faster. Note that this algorithm
suffers from the same problem as Minimax, namely that the number of
states to analyze still explodes exponentially with the number of
rounds to anticipate.

The $k$-best pruning technique \citep{baier2018mcts} was used to
mitigate the phenomenon of exponential evolution with respect to the
number of anticipated rounds. Note that, with Minimax, the number of states explored
is of the order of the average number of actions per round to the
power of the number of anticipated rounds. Therefore, reducing the
number of actions makes it possible to decrease the number of states
to be analyzed. The $k$-best pruning technique thus consists in restricting
the search to the $k$ best actions for each state analyzed. Since
obviously we cannot know the best actions in advance, the best actions
identified during the previous search are used. This pruning rule
drastically accelerates the search, and therefore makes it possible
to significantly increase the number of anticipated rounds. However,
this is obviously an inexact pruning. The winning strategies may well
be found in the parts of the tree pruned by this technique. Another
flaw of this technique is that it depends on one parameter: the number
of actions to consider, which must therefore generally be optimized
for each context. Minimax with Alpha-Beta and $k$-best pruning is
denoted $k$-best $\alphabeta$.

The Principal Variation Search algorithm \citep{pearl1980scout} (also
called the Scout algorithm), abbreviated $\pvs{}$, is a variant of
the Alpha-Beta algorithm. The algorithm assumes that the first child
state evaluated is the best of the child states. Each of the other
children is evaluated by a restricted search that uses this assumption
so that this restricted search is faster than the full search it replaces.
If a child evaluated by a restricted search is not worse than the
best of the previous children, then a full search is performed on
that child.

Memory Test Driver \citep{plaat1996best}, $\mtdf{}$, is an algorithm
that uses Alpha-Beta as a sub-procedure. This algorithm iteratively
applies Alpha-Beta with $\beta=\alpha+1$, which performs very powerful
pruning. It then uses the information from this search to perform
the next Alpha-Beta search with a better value of $\alpha$, so that
the algorithm finds $\alpha$ such that the value of the root is between
$\alpha$ and $\beta=\alpha+1$.

The Monte Carlo Tree Search algorithm \citep{Coulom06,browne2012survey}
($\mcts{}$ for short) is an algorithm similar to Unbounded Minimax.
Like the latter, it extends the most interesting action sequences
iteratively as long as it has time left. Thus, it is anytime and can
perform deep local searches. However, they have fundamental differences.
For Unbounded Minimax, an action is more interesting than another,
if it lead to a state of better value. While for $\mcts{}$, an action
is better than another, if the sum of its victory statistic and an
exploration term is larger. These victory statistics are determined
during the search. More precisely, every time that an actions sequence
is extended, $\mcts{}$ performs a random endgame simulation continuing
this sequence. It then uses this information to update the victory
statistics of the actions of the sequence. Thus, $\mcts{}$ uses this
statistic for the states evaluation and therefore does not use a (learned)
evaluation function. The exploration term helps manage the exploration/exploitation
dilemma \citep{mandziuk2010knowledge} and thus favors states that
are less explored than others. It depends on a constant $C$ that
must be tuned for each context.

However, $\mcts{}$ rarely has a high level of play. Many researches
have therefore attempted to inject knowledge into it \citep{baier2018mcts}.
The culmination of this research path is the search algorithm used
within AlphaZero. The latter replaces the evaluation of states based
on random endgames victory statistics by the classic use of state
evaluation functions. It also uses a learned probability distribution
for actions. This search algorithm cannot be compared within this
article since it requires a learned policy to be used. Indeed, we
place ourselves in the context of the state of the art, namely that
of Athénan based on the algorithm Descent which generates the evaluation
functions without actions probabilities \citep{2020learning}. Anyway,
Athénan (and therefore its search algorithm Unbounded Minimax) and
AlphaZero (and therefore its version of $\mcts{}$) have already been
compared \citep{cohen2023minimax}: Athénan is much more efficient
than AlphaZero. However, there remains one question. Can $\mcts{}$
be better than Unbounded Minimax with Athénan's evaluation functions?
This amounts to using AlphaZero's $\mcts{}$ without a learned policy,
which corresponds to an algorithm already proposed \citep{ramanujan2011trade},
which we will denote by $\mctsh{}$.

\section{Experimental comparison of the main search algorithms}

We now focus on the performance of the main search algorithms of the
literature.

\subsection{Summary of the experimental protocol}

We use $\ubfm$ as benchmark adversary using a lossless parallelization
procedure, called Child Batching \citep{2020learning} (child Batching
consists in evaluating in parallel the children states of states).
We evaluate each algorithm on 22 recurring games at the Computer Olympiad
(see Table~\ref{tab:list-of-games}). Each algorithm faces the benchmark
adversary twice (once as first player and once as second player) and
this for each pair of evaluation functions. We use about fifteen evaluation
functions. This corresponds to approximately 450 matches for each
evaluated algorithm. This evaluation is repeated 6 times with a different
set of evaluation functions. Each winning percentage of a search algorithm on a game
is therefore the average over about 2700 matches. The evaluation functions
were generated using Descent, the Athénan's evaluation function learning
algorithm \citep{2020learning}. Each algorithm uses the same search
time: 10 seconds per action. The performance of a search algorithm
is the average of the scores obtained at the end of its matches (1
for a victory, 0 for a draw and -1 for a defeat). We conducted this
experiment in two stages. First, we evaluated the basic algorithms
without Child Batching (because Child Batching cannot be used with
$\mcts{}$). Second, we evaluated all algorithms, except the Monte
Carlo algorithms, with Child Batching. Note that whether in the first
or second stage, the benchmark adversary always uses Child Batching.
Thus, the algorithms of the two stages can be directly compared. At
each stage, we give for each algorithm: the average performance over
the 22 games, the associated 5\% stratified bootstrapping confidence
interval (lower and upper bound) and the performance for each game
with its 5\% confidence radius.

\subsection{First stage of the comparison: without Child Batching}

During the first stage of the comparison, we evaluate the following
algorithms without Child Batching: $\mcts{}$, $\mctsh{}$, $\alphabeta$,
$\ubfm$, and $\ubfms$. The algorithms performances are described
in Table~\ref{tab:res-sans-batching} (mean over the games) and Table~\ref{tab:Performance-main-10s-without-child-batching}
(details). Both Monte Carlo algorithms obtain significantly worse
results in all games. So even if we could use Child Batching with
them, these algorithms would still be outdated. Moreover, $\ubfms$
obtains the best results on average but also on all games. $\ubfm$
is the second best algorithm in average.

\subsection{Second stage of the comparison: with Child Batching}

Then, we apply the same study but each algorithm uses this time Child
Batching. We evaluate the following algorithms: $\alphabeta$, $\mtdf{}$,
$\pvs{}$, $k$-best $\alphabeta$, $\ubfm$, and finally $\ubfms$.
The algorithms performances are described in Table~\ref{tab:res-avec-batching}
(mean over the games) and Table~\ref{tab:Performance-main-10s-with-child-batching}
(details). $\ubfms$ obtains the best average performance and it is
the best algorithm over $17$ of the $22$ studied games. Over $4$
of the $5$ games, $\ubfms$ is the second best algorithm. In addition, over
the $5$ games, the best algorithm is $k$-best $\alphabeta$ , and
thus it is a parametrized algorithm that requires a costly tuning
phase to be used.  $\ubfm$ is the second best algorithm in average.

\subsection{Comparison in a short time}

We also repeated the first and second stage of the experiment with
a search time per action of 1 second. This time, in addition to be
the best algorithm on average, $\ubfms$ is also the best algorithm
for each game (see Table~\ref{tab:Performance-main-1s-without-child-batching}
and Table~\ref{tab:Performance-main-1s-with-child-batching}).
$\ubfm$ is the second best algorithm in average.

\section{Conclusion}

We conducted a large and robust experiment evaluating the main search
algorithms in the literature. We showed that Unbounded Minimax achieves
better average results than other algorithms, while this algorithm
is not used in practice (except recently in the context of Athénan)
or considered state-of-the-art of game search algorithms by the literature. 

Furthermore, we proposed a variant of Unbounded Minimax, using the
safe decision, and we showed that this constitutes an improvement
of the algorithm in all cases. This algorithm is even the best algorithm
on all games with a short search time. Moreover, it is the best
algorithm on 17 of the 22 games and the best algorithm on average,
when using a medium search time.

In \citep{2020learning}, Athénan, the architecture combining Unbounded Minimax and Descent (the learning algorithm), was shown to be superior to AlphaZero and thus  to be the new state of the art to master games without using human knowledge. However, it was unclear whether this superiority was solely due to Descent or due to Descent and Unbounded Minimax. Indeed, 
this study does not show whether
modifying Athénan using another search algorithm could yield  better results. This article bridges the gap: Unbounded Minimax with safe decision is thus the best algorithm on average according to this study. It is better in all cases studied in this article when time is short and better in the vast majority of cases for a medium search time. This article also shows that it is possible to obtain better performance in particular cases by considering another search algorithm for Athénan, in particular $k$-best Alpha-Beta.

The pseudo-codes and detailed descriptions of the used algorithms,
the optimization procedures of the parameter-based algorithms, the
description of games, and other details about the performed experiments
are available in Appendix.

%%%%%%%%%%%%%%%% MAIN TEXT FIGURES %%%%%%%%%%%%%%%

\begin{figure}
\begin{centering}
\includegraphics[width=0.6\textwidth]{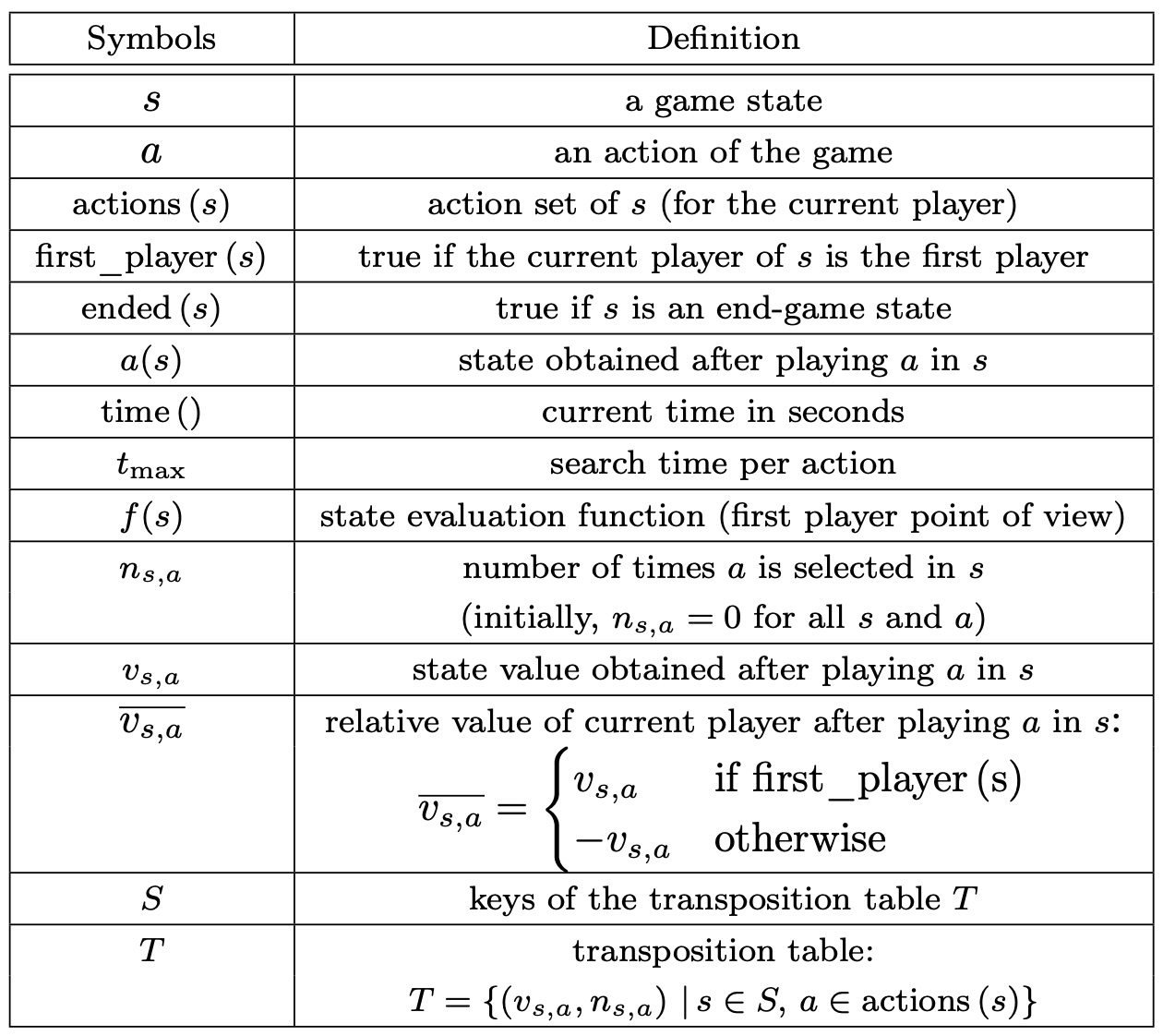} 
\par\end{centering}
\caption{\textbf{Definition of symbols.}\label{fig:def-symbol}}
\end{figure}

{\small{}}
\begin{figure}
\begin{centering}
{\small{}{}\includegraphics[width=0.5\textwidth]{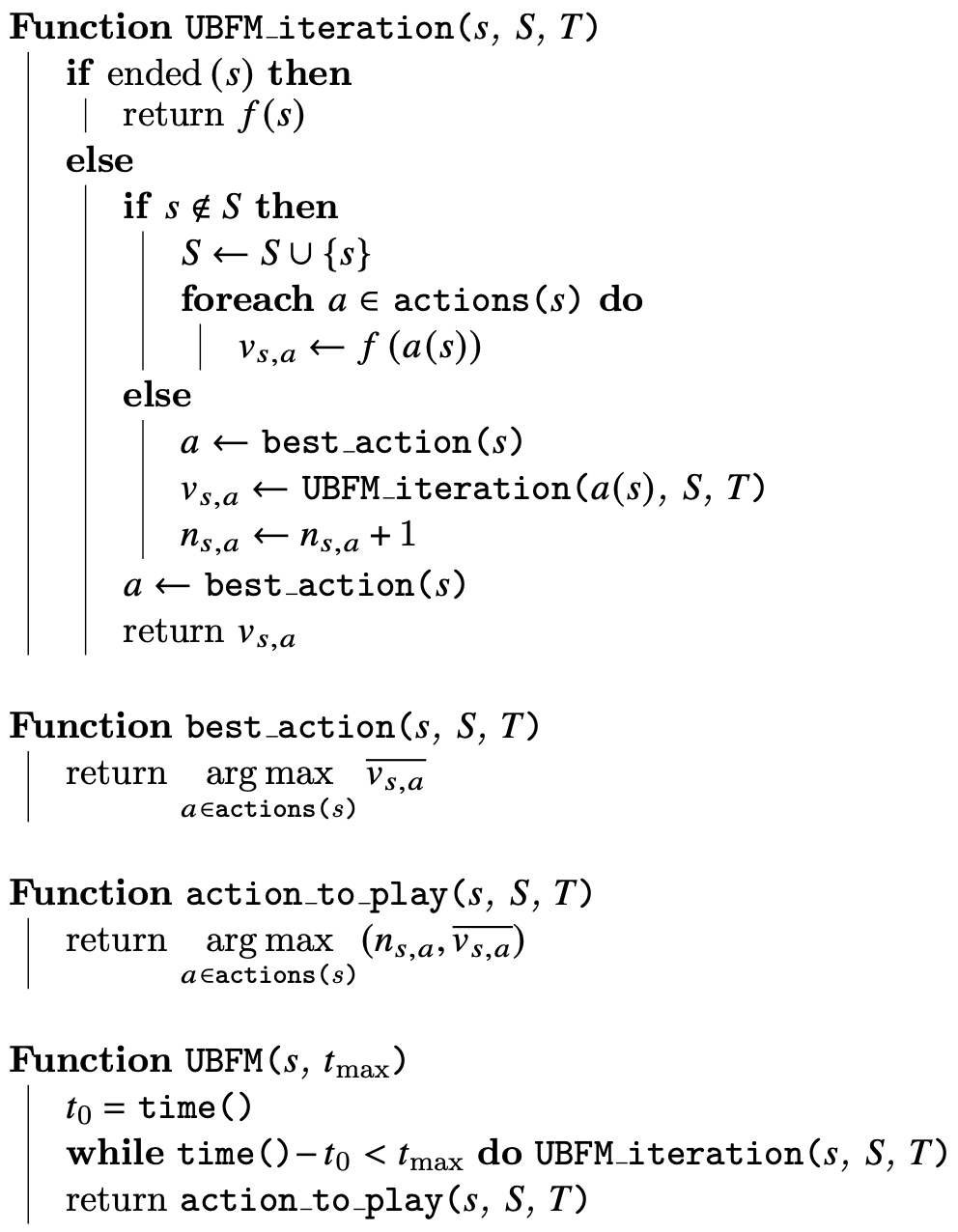}}{\small\par}
\par\end{centering}
{\small{}{}\caption{\textbf{Unbounded Best-First Minimax with safe decision algorithm
($\protect\ubfms$).} It computes the best action to play in the generated
non-uniform partial game tree starting from the root state $s$ with
$\protect\tmax$ as research time 
%(see Table~\ref{tab:Index-of-symbols}
(see Figure~\ref{fig:def-symbol}
for the definitions of symbols ; at any time $T=\protect\liste{\left(\protect\mv sa,\protect\select sa\right)}{s\in S\protect\et a\in\mathrm{actions}(s)}$).
Note: tuples are lexicographically ordered.\label{alg:safe-Unbounded-Best-First-Mininmax}}
}{\small\par}
\end{figure}

\begin{table}
\centering{}\caption{\textbf{Performance of basic search algorithms without Child Batching
and with 10 seconds per action.}\label{tab:res-sans-batching}}
\begin{tabular}{ccc||cc||c||c}
\hline 
 & {\footnotesize{}{}mean} & \multicolumn{2}{c}{{\footnotesize{}{}lower bound}} & \multicolumn{3}{c}{{\footnotesize{}{}upper bound}}\tabularnewline
\hline 
{\footnotesize{}{}$\ubfms$} & {\footnotesize{}{}$-23.02\%$} & \multicolumn{2}{c}{{\footnotesize{}{}$-23.59\%$}} & \multicolumn{3}{c}{{\footnotesize{}{}$-22.45\%$}}\tabularnewline
{\footnotesize{}{}$\ubfm$} & {\footnotesize{}{}$-34.12\%$} & \multicolumn{2}{c}{{\footnotesize{}{}$-34.66\%$}} & \multicolumn{3}{c}{{\footnotesize{}{}$-33.56\%$}}\tabularnewline
{\footnotesize{}{}$\alphabeta$} & {\footnotesize{}{}$-54.33\%$} & \multicolumn{2}{c}{{\footnotesize{}{}$-54.81\%$}} & \multicolumn{3}{c}{{\footnotesize{}{}$-53.85\%$}}\tabularnewline
{\footnotesize{}{}$\mcts{}$} & {\footnotesize{}{}$-93.55\%$} & \multicolumn{2}{c}{{\footnotesize{}{}$-93.38\%$}} & \multicolumn{3}{c}{{\footnotesize{}{}$-93.08\%$}}\tabularnewline
{\footnotesize{}{}$\mctsh{}$} & {\footnotesize{}{}$-77.38\%$} & \multicolumn{2}{c}{{\footnotesize{}{}$-77.72\%$}} & \multicolumn{3}{c}{{\footnotesize{}{}$-77.03\%$}}\tabularnewline
\end{tabular}
\end{table}

\begin{table}
\begin{centering}
\caption{\textbf{Performance of main search algorithms with Child Batching
and 10 seconds per action.}\label{tab:res-avec-batching}}
\par\end{centering}
\centering{}%
\begin{tabular}{ccc||cc||c}
\hline 
 & {\footnotesize{}{}mean} & \multicolumn{2}{c}{{\footnotesize{}{}lower bound}} & \multicolumn{2}{c}{{\footnotesize{}{}upper bound}}\tabularnewline
\hline 
{\footnotesize{}{}$\ubfms$} & {\footnotesize{}{}$8.64\%$} & \multicolumn{2}{c}{{\footnotesize{}{}$8.06\%$}} & \multicolumn{2}{c}{{\footnotesize{}{}$9.23\%$}}\tabularnewline
{\footnotesize{}{}$\ubfm$} & {\footnotesize{}{}$0.08\%$} & \multicolumn{2}{c}{{\footnotesize{}{}$-0.51\%$}} & \multicolumn{2}{c}{{\footnotesize{}{}$0.68\%$}}\tabularnewline
{\footnotesize{}{}$k$-best $\alphabeta$} & {\footnotesize{}{}$-11.78\%$} & \multicolumn{2}{c}{{\footnotesize{}{}$-12.37\%$}} & \multicolumn{2}{c}{{\footnotesize{}{}$-11.18\%$}}\tabularnewline
{\footnotesize{}{}$\mtdf{}$} & {\footnotesize{}{}$-39.69\%$} & \multicolumn{2}{c}{{\footnotesize{}{}$-40.21\%$}} & \multicolumn{2}{c}{{\footnotesize{}{}$-39.16\%$}}\tabularnewline
{\footnotesize{}{}$\alphabeta$} & {\footnotesize{}{}$-41.79\%$} & \multicolumn{2}{c}{{\footnotesize{}{}$-42.3\%$}} & \multicolumn{2}{c}{{\footnotesize{}{}$-41.25\%$}}\tabularnewline
{\footnotesize{}{}$\pvs{}$} & {\footnotesize{}{}$-42.3\%$} & \multicolumn{2}{c}{{\footnotesize{}{}$-42.3\%$}} & \multicolumn{2}{c}{{\footnotesize{}{}$-42.82\%$}}\tabularnewline
\end{tabular}
\end{table}

\begin{table}
{\small{}\centering{}{}\caption{\textbf{Performance of basic search algorithms without Child Batching
and with $10$ seconds per action} \textbf{for the studied games.
}Performances are in percentage.\label{tab:Performance-main-10s-without-child-batching}}
}%
\begin{tabular}{cccccccccccc}
\hline 
 & {\tiny{}{}Amazons} & {\tiny{}{}Arimaa} & {\tiny{}{}Ataxx} & {\tiny{}{}Breaktrhough} & {\tiny{}{}Brazilian} & {\tiny{}{}Canadian} & {\tiny{}{}Clobber} & {\tiny{}{}Connect6} & {\tiny{}{}International} & {\tiny{}{}Chess} & {\tiny{}{}Go}\tabularnewline
 &  &  &  &  & {\tiny{}{}Draughts} & {\tiny{}{}Draughts} &  &  & {\tiny{}{}Draughts} &  & {\small{}${\scriptscriptstyle 9\times9}$}\tabularnewline
\hline 
{\scriptsize{}{}$\ubfms$} & {\tiny{}{}-26 }{\small{}${\scriptscriptstyle {\scriptscriptstyle \pm}}$}{\tiny{}{}
3} & {\tiny{}{}-22 }{\small{}${\scriptscriptstyle {\scriptscriptstyle \pm}}$}{\tiny{}{}
1} & {\tiny{}{}-60 }{\small{}${\scriptscriptstyle {\scriptscriptstyle \pm}}$}{\tiny{}{}
4} & {\tiny{}{}-8 }{\small{}${\scriptscriptstyle {\scriptscriptstyle \pm}}$}{\tiny{}{}
3} & {\tiny{}{}-1 }{\small{}${\scriptscriptstyle {\scriptscriptstyle \pm}}$}{\tiny{}{}
1} & {\tiny{}{}-14 }{\small{}${\scriptscriptstyle {\scriptscriptstyle \pm}}$}{\tiny{}{}
3} & {\tiny{}{}-33 }{\small{}${\scriptscriptstyle {\scriptscriptstyle \pm}}$}{\tiny{}{}
3} & {\tiny{}{}-36 }{\small{}${\scriptscriptstyle {\scriptscriptstyle \pm}}$}{\tiny{}{}
5} & {\tiny{}{}-4 }{\small{}${\scriptscriptstyle {\scriptscriptstyle \pm}}$}{\tiny{}{}
1} & {\tiny{}{}-33 }{\small{}${\scriptscriptstyle {\scriptscriptstyle \pm}}$}{\tiny{}{}
2} & {\tiny{}{}-8 }{\small{}${\scriptscriptstyle {\scriptscriptstyle \pm}}$}{\tiny{}{}
3}\tabularnewline
\end{tabular}
\end{table}
{\small\par}

{\small{}{} }
\begin{table}
{\small{}\centering{}{}\caption{\textbf{Performance of main search algorithms with Child Batching
and $10$ seconds per action for the studied games. }Performances
are in percentage.\label{tab:Performance-main-10s-with-child-batching}}
}%
\begin{tabular}{cccccccccccc}
\hline 
 & {\tiny{}{}Amazons} & {\tiny{}{}Arimaa} & {\tiny{}{}Ataxx} & {\tiny{}{}Breaktrhough} & {\tiny{}{}Brazilian} & {\tiny{}{}Canadian} & {\tiny{}{}Clobber} & {\tiny{}{}Connect6} & {\tiny{}{}International} & {\tiny{}{}Chess} & {\tiny{}{}Go 9}\tabularnewline
\hline 
{\tiny{}{}$\ubfms$} & {\tiny{}{}13 }{\small{}${\scriptscriptstyle {\scriptscriptstyle \pm}}$}{\tiny{}{}
3} & {\tiny{}{}5 }{\small{}${\scriptscriptstyle {\scriptscriptstyle \pm}}$}{\tiny{}{}
1} & {\tiny{}{}10 }{\small{}${\scriptscriptstyle {\scriptscriptstyle \pm}}$}{\tiny{}{}
5} & {\tiny{}{}25 }{\small{}${\scriptscriptstyle {\scriptscriptstyle \pm}}$}{\tiny{}{}
3} & {\tiny{}{}0 }{\small{}${\scriptscriptstyle {\scriptscriptstyle \pm}}$}{\tiny{}{}
1} & {\tiny{}{}4 }{\small{}${\scriptscriptstyle {\scriptscriptstyle \pm}}$}{\tiny{}{}
3} & {\tiny{}{}19 }{\small{}${\scriptscriptstyle {\scriptscriptstyle \pm}}$}{\tiny{}{}
3} & {\tiny{}{}6 }{\small{}${\scriptscriptstyle {\scriptscriptstyle \pm}}$}{\tiny{}{}
5} & {\tiny{}{}0 }{\small{}${\scriptscriptstyle {\scriptscriptstyle \pm}}$}{\tiny{}{}
1} & {\tiny{}{}3 }{\small{}${\scriptscriptstyle {\scriptscriptstyle \pm}}$}{\tiny{}{}
2} & {\tiny{}{}8 }{\small{}${\scriptscriptstyle {\scriptscriptstyle \pm}}$}{\tiny{}{}
3}\tabularnewline
\end{tabular}
\end{table}
{\small\par}

{\small{}{} }
\begin{table}
{\small{}\centering{}{}\caption{\textbf{Performance of basic search algorithms without Child Batching
and with $1$ second per action. }Performances are in percentage.\label{tab:Performance-main-1s-without-child-batching}}
}%
\begin{tabular}{cccccccccccc}
\hline 
 &  &  &  & {\tiny{}{}mean} & \multicolumn{2}{c}{{\tiny{}{}lower bound}} & \multicolumn{2}{c}{{\tiny{}{}upper bound}} &  &  & \tabularnewline
\hline 
 &  &  & {\scriptsize{}{}$\ubfms$} & {\tiny{}{}-38.57} & \multicolumn{2}{c}{{\tiny{}{}-39.12}} & \multicolumn{2}{c}{{\tiny{}{}-38.02}} &  &  & \tabularnewline
 &  &  & {\scriptsize{}{}$\ubfm$} & {\tiny{}{}-43.55} & \multicolumn{2}{c}{{\tiny{}{}-44.08}} & \multicolumn{2}{c}{{\tiny{}{}-43.01}} &  &  & \tabularnewline
 &  &  & {\scriptsize{}{}$\alphabeta$} & {\tiny{}{}-61.69} & \multicolumn{2}{c}{{\tiny{}{}-62.16}} & \multicolumn{2}{c}{{\tiny{}{}-61.23}} &  &  & \tabularnewline
 &  &  & {\scriptsize{}{}$\mcts{}$} & {\tiny{}{}-95.13} & \multicolumn{2}{c}{{\tiny{}{}-95.23}} & \multicolumn{2}{c}{{\tiny{}{}-95.02}} &  &  & \tabularnewline
 &  &  & {\scriptsize{}{}$\mctsh{}$} & {\tiny{}{}-78.47} & \multicolumn{2}{c}{{\tiny{}{}-78.83}} & \multicolumn{2}{c}{{\tiny{}{}-78.12}} &  &  & \tabularnewline
\hline 
 & {\tiny{}{}Amazons} & {\tiny{}{}Arimaa} & {\tiny{}{}Ataxx} & {\tiny{}{}Breaktrhough} & {\tiny{}{}Brazilian} & {\tiny{}{}Canadian} & {\tiny{}{}Clobber} & {\tiny{}{}Connect6} & {\tiny{}{}International} & {\tiny{}{}Chess} & {\tiny{}{}Go}\tabularnewline
 &  &  &  &  & {\tiny{}{}Draughts} & {\tiny{}{}Draughts} &  &  & {\tiny{}{}Draughts} &  & {\small{}${\scriptscriptstyle 9\times9}$}\tabularnewline
\end{tabular}
\end{table}
{\small\par}

{\small{}{} }
\begin{table}
{\small{}\centering{}{}\caption{\textbf{Performance of main search algorithms with Child Batching
and $1$ second per action. }Performances are in percentage.\label{tab:Performance-main-1s-with-child-batching}}
}%
\begin{tabular}{cccccccccccc}
\hline 
 &  &  &  & {\tiny{}{}mean} & \multicolumn{2}{c}{{\tiny{}{}lower bound}} & \multicolumn{2}{c}{{\tiny{}{}upper bound}} &  &  & \tabularnewline
\hline 
 &  &  & {\tiny{}{}$\ubfms$} & {\tiny{}{}7.49} & \multicolumn{2}{c}{{\tiny{}{}6.88}} & \multicolumn{2}{c}{{\tiny{}{}8.08}} &  &  & \tabularnewline
 &  &  & {\tiny{}{}$\ubfm$} & {\tiny{}{}1.22} & \multicolumn{2}{c}{{\tiny{}{}0.59}} & \multicolumn{2}{c}{{\tiny{}{}1.83}} &  &  & \tabularnewline
 &  &  & {\tiny{}{}$k$-best $\alphabeta$} & {\tiny{}{}-14.34} & \multicolumn{2}{c}{{\tiny{}{}-14.95}} & \multicolumn{2}{c}{{\tiny{}{}-13.72}} &  &  & \tabularnewline
 &  &  & {\tiny{}{}$\mtdf{}$} & {\tiny{}{}-44.35} & \multicolumn{2}{c}{{\tiny{}{}-44.89}} & \multicolumn{2}{c}{{\tiny{}{}-43.82}} &  &  & \tabularnewline
 &  &  & {\tiny{}{}$\alphabeta$} & {\tiny{}{}-46.48} & \multicolumn{2}{c}{{\tiny{}{}-47.02}} & \multicolumn{2}{c}{{\tiny{}{}-45.96}} &  &  & \tabularnewline
 &  &  & {\tiny{}{}$\pvs{}$} & {\tiny{}{}-42.13} & \multicolumn{2}{c}{{\tiny{}{}-42.68}} & \multicolumn{2}{c}{{\tiny{}{}-41.59}} &  &  & \tabularnewline
\hline 
 & {\tiny{}{}Amazons} & {\tiny{}{}Arimaa} & {\tiny{}{}Ataxx} & {\tiny{}{}Breaktrhough} & {\tiny{}{}Brazilian} & {\tiny{}{}Canadian} & {\tiny{}{}Clobber} & {\tiny{}{}Connect6} & {\tiny{}{}International} & {\tiny{}{}Chess} & {\tiny{}{}Go 9}\tabularnewline
\end{tabular}
\end{table}

 \newpage

\appendix
%dummy comment inserted by tex2lyx to ensure that this paragraph is not empty%dummy comment inserted by tex2lyx to ensure that this paragraph is not empty

\section{Appendix}

\subsection{Plan of Appendix}

We present in this document everything that is necessary to understand
the scope of the article and to be able to reproduce its experiments.
It also details the tuning of the parameter-based algorithms (specifying
in particular the results for each tested parameter value).

We begin by discussing search algorithms broadly in Section~\ref{subsec:Algorithmes-de-recherche}.
Next, we specify the notations and symbols common to the algorithms
evaluated in this article (Section~\ref{subsec:Symboles-et-notations}).
Then, in Section~\ref{sec:Techniques-transversales}, we present
transversal techniques improving the performance of search algorithms,
which we uses in the experiments of this paper. In Section~\ref{sec:Proc=00003D0000E9dure-d'=00003D0000E9valuation-commune},
we detail the common evaluation procedure of the different search
algorithms studied. This includes the machine used, the games, the
evaluation functions, the adversary benchmark algorithm, the evaluation
procedure and the performance criterion.

Finally, we present in Section~\ref{subsec:Algorithms-and-their-tuning}
the studied algorithms and their tuning. 

\subsection{Games and search algorithms\label{subsec:Algorithmes-de-recherche}}

\subsubsection{Generality}

Game search algorithms are algorithms that explore the different possible
continuations of a game match in order to determine the best possible
strategy within the limits of available resources. The algorithm constructs
a game tree, that is, a tree where the nodes are the game states and
a state $c$ is a child of another state $s$ if there is an action
in $s$ allowing to move from $s$ to $c$. It is generally not possible
to construct the complete game tree, that is, to construct all possible
strategies. For example, in Chess, there are approximately ${10}^{120}$
possible matches. A branch of the complete game tree (whose root is
the initial state of the game) is therefore a complete game match.
A leaf state of the complete tree is an endgame state. Since resources
are limited, a search algorithm will therefore build a partial game
tree, that is, it will restrict itself to a subset of strategies.
The states of the game tree are labeled by values indicating how interesting
a state is. The more interesting a state is, the more likely it is
to be part of the strategy finally chosen. For example, end-game states
are associated with a value indicating the winner. This value is generally
1 if the first player is the winner, 0 in the event of a draw, and
-1 if the second player is the winner; remember that we are in the
context of two-player, perfect information, zero-sum games, that is,
strictly competitive games. This value can also be the game score
when it has one (the case 1,0,-1 is in fact the most basic score function).
In theory, a search algorithm does not need anything else to determine
the optimal strategy. However, a search algorithm will only have access
to the scores of the endgame states towards the end of the match.
At the beginning of the match, the partial game trees do not contain
any or not enough of them. Indeed, at the beginning, the partial trees
cannot be large enough. A search algorithm therefore generally uses
an evaluation function for non-endgame states, called \emph{heuristic}.
This heuristic evaluation function thus makes it possible to evaluate
the non-endgame states by a value supposed to approximate the true
value of this state induced by the values of its successor endgame
states.

Overviews of this research area are available here~\citep{bouzy2020artificial,millington2009artificial}.

\subsubsection{Minimax at bounded depth}

The basic algorithm is the minimax algorithm. It associates with each
state a theoretical value, called the \emph{minimax value}, which
corresponds to the endgame score reached from this state assuming
that the players play optimally. The first player seeks to maximize
his score. Thus, the value of a state in which the first player must
play is the maximum of the values of its child states. Conversely,
the second player seeks to minimize the score of the first player.
Thus the value of a state in which the second player must play is
the minimum of the values of its child states. The minimax value of
each state is thus calculated recursively. Since this algorithm cannot
be applied to the complete game tree, it is applied to the partial
game tree. The practical application of this algorithm then behaves
as if the game stopped at the leaves of the partial game tree (called
the \emph{search horizon}). The practical value of a state is then
the \emph{partial minimax value}, that is to say the value of the
leaf of the partial tree reached by assuming that the players play
optimally considering that the score of the non-endgame leaves is
thus given by the heuristic evaluation function.

The practical application of the minimax is parameterized by the search
depth $d$ which defines how many rounds in advance the algorithm
anticipates the different possible evolutions of the match. It thus
constructs the partial game tree of the possible continuations of the match
whose depth is $d$. After the construction of this partial game tree,
the action played by this algorithm is the action of the root leading
to the child state of best value. If the player of the root is the
first player, the state of best value is the one of highest value.
If the player of the root is the second player, the state of best
value is the state of lowest value.

This algorithm has been improved with the so-called alpha-beta pruning
technique.~\citep{knuth1975analysis} which allows to restrict the
state space and thus the search space. This restriction is done by
pruning certain areas of the game tree which the algorithm is able
to prove are not part of the optimal strategy, using the properties
of min and max.

The advent of Alpha-Beta is the Deep Blue Chess program \citep{campbell2002deep},
first program to beat a world champion at Chess. However, this success
has some limitations. On the one hand, Deep Blue's success comes largely
from the power of the supercomputer used to calculate two hundred
million game states per second. On the other hand, Deep Blue's heuristic
evaluation function was designed by hand from knowledge dedicated
to Chess, making the application of this method in another context
impossible without numerous adaptations and without any guarantee
of success. The design of evaluation functions is indeed a particularly
complex task that must be adapted for each game. Much work has
been done on the automatic design of evaluation functions~\citep{mandziuk2010knowledge}.
One of the first successes was on the game Backgammon \citep{tesauro1995temporal},
which has allowed it to approach the level of the best players in
this game. However, although most of these approaches are general,
they have not allowed to obtain a high level of play on other games,
such as Go or Hex. In fact, even the hand-crafted design of evaluation
functions has not succeeded in obtaining a high level of play in Go
and Hex.

\subsubsection{Monte Carlo Tree Search}

Due to these limitations, a completely different approach was proposed:
the Monte Carlo Tree Search algorithm (a.k.a. MCTS)~\citep{Coulom06,browne2012survey}.
This algorithm was the first to give promising results to the game
of Go and to the game Hex. Unlike Minimax, it has two advantages.
The first is that it does not need a heuristic evaluation function
to work. MCTS evaluates states by victory statistics based on endgame
simulations. The second advantage is that it creates a non-uniform
search tree, allowing it to focus on the most promising strategies.
We will say that such an algorithm is of unbounded depth. Indeed,
one of the weaknesses of the minimax is that it builds a uniform game
tree, that is to say that it is an algorithm with bounded depth. In
other words, one of the limits of Minimax is that the number of states
explored and therefore the complexity of this algorithm evolve exponentially
according to the depth of the game tree, that is to say according
to the number of anticipated rounds. It is therefore not possible
in practice to plan very many tours in advance (for time reasons but
sometimes also for memory reasons).

The Monte Carlo Tree Search was then improved by injecting knowledge
into it, in particular to guide the exploration of the tree and/or
end-game simulations \citep{browne2012survey}. This was first done
using supervised learning \citep{clark2015training}, then by supervised
learning and by reinforcement learning \citep{silver2016mastering}.
The apogee of Monte Carlo Tree Search occurred with the use of reinforcement
learning only, specifically with the AlphaZero algorithm \citep{silver2018general},
able to achieve state-of-the-art level on many games without any human
knowledge. This latest success, however, suffers from a significant
limitation. Its training requires a colossal amount of resources to
reach a state-of-the-art level. For example, in the game of Chess,
it took about 2,293,760,000,000 state evaluations to surpass the benchmark
program called Stockfish, which with a single high-end graphics card
would take more than 20 years to achieve.

\subsubsection{Minimax at unbounded depth}

In fact, other unbounded depth algorithms have been proposed before
it~\citep{korf1994best,schaeffer1990conspiracy,berliner1981b}. For
example, there is Unbounded (Best-First) Minimax~\citep{korf1996best}.
This one had some success in Othello and then fell into oblivion.
This algorithm, like MCTS, performs a best-first search, but its search
is based on a heuristic evaluation function. More precisely, this
algorithm performs a very selective search guided by its heuristic.
However, at that time the heuristic evaluation functions were of a
more or less average level. However, with this algorithm, if the heuristic
evaluation function is bad, the partial game tree will be bad, in
addition to the fact that the values of this tree will be bad. Its
performance will then be lower than with a classic minimax. It has
therefore not been used or very little since. This algorithm was rediscovered
a few years ago as one of the basic components of the Athénan learning
and search algorithm~\citep{2020learning} and has even been improved.
It has thus benefited from the heuristic evaluation functions generated
by the learning part of Athénan: the zero-knowledge reinforcement
learning algorithm \emph{Descent}. Ultimately, this combination has
made it possible to exceed the state-of-the-art level on many games
\citep{2020learning,cohen2023minimax} and in particular to win many
gold medals at the Computer Olympiad (since its appearance, in 5 competitions,
Athénan has won 48 gold medals, 8 silver and 4 bronze, and is the
current title holder of 17 games~\citep{cohen2021descent,cohen2023athenan}).
It has notably beaten numerous reimplementations of AlphaZero as well
as the old state-of-the-arts of various games in the competition.
This superiority of Athénan over AlphaZero has even been studied experimentally,
proving to have a learning rate more than 30 times faster and even
being able to equal the performance of AlphaZero on certain games
when AlphaZero is equipped with 100 times more graphics cards~\citep{cohen2023minimax}.

\begin{comment}
Note that before the introduction of Athénan and AlphaZero, there
was no best search algorithm. It was different for each game and was
each time a different variant of MCTS or a different variant of minimax
with bounded depth. However, the article \citep{2020learning} showed
that for $9$ games, Unbounded Minimax with safe decision is better
than Unbounded Minimax, and that Unbounded Minimax is better than
Alpha-Beta. The study of this paper extends \citep{2020learning}
by including the missing search algorithms, using much more efficient
evaluation functions (from 20 to 70 days of learning using a graphics
card instead of 48 hours with only with 4 CPUs) and more than two
times more games.
\end{comment}

\subsubsection{Search algorithms excluded from this study}

Note that we have excluded from the study of this article some algorithms
that are little studied and used in practice. First, there is Conspiracy
Number Search and its variants \citep{mcallester1988conspiracy,schaeffer1990conspiracy}
which are unbounded depth-first search algorithms. Unlike Unbounded
Minimax, they construct the game tree in a non-uniform manner so as
to maximize the number of states in the tree whose value change is
required for the value of the root to change. We are not aware of
any convincing positive results concerning this family of algorithms
that would justify including it in this study. One other algorithm
was excluded, the B{*} algorithm~\citep{berliner1981b}. This algorithm
requires an evaluation function that returns two values: a lower bound
and an upper bound. Although one could convert a single-valued heuristic
evaluation function into a two-valued function, this procedure is
parametric. Thus, in addition to the ad-hoc nature of this procedure,
this parameterization harms the generality of its application. Its
study in this context where we seek general results is therefore not
relevant, at least for the moment. Not until we have a reinforcement
learning method for heuristic evaluation functions that provides a
good quality lower and upper bound.

The search algorithm of AlphaZero \citep{silver2017mastering,silver2016mastering}
is a variant of MCTS which was also excluded from this study. This
algorithm uses a neural network to obtain a policy, i.e. a probability
distribution over actions. The AlphaZero search algorithm cannot therefore
be compared within this article since it uses a policy learned by
reinforcement and the other algorithms do not use one. This potential
comparison is outside the scope of this article since having a policy
learning algorithm is a big assumption. Obtaining a state-of-the-art
policy is very expensive. For example, training AlphaZero to play
Chess required 2293760000000 state evaluations, which with a single
powerfull graphics card would take over 20 years to achieve. Furthermore,
we place ourselves here within the context of the state of the art
of two-player game algorithms, namely Athénan, which does not use
a policy. Athénan only uses a learned function for evaluating states,
generated by the Descent algorithm (see section \ref{subsec:Fonctions-d'=00003D0000E9valuation-utilis=00003D0000E9s}).
Note that AlphaZero and Athénan have in any case already been compared
in their entirety \citep{cohen2023minimax}, and Athénan has much
better results. Note that the MCTS variant which uses a learned evaluation
function for states in the manner of the search algorithm of AlphaZero,
but which does not use a learned policy, is evaluated in this article
(see Section~\ref{subsec:Neural-value-based-MCTS}).

Finally, we also excluded hybridizations of the Minimax algorithm
and the Monte Carlo Tree Search algorithm~\citep{baier2018mcts}.
These algorithms are based on a Monte Carlo Tree Search that uses
at different levels and at each iteration an Alpha-Beta search to
guide or replace the endgame simulations and/or bias the probabilities.
They were excluded for the following reasons. Theses algorithms have
many parameters (up to 11 parameters, 8 for their best hybridization),
so they are very expensive to tune: a robust and complete experimental
study seems to be out of reach, except by using more than colossal
computing power. They have not proven to be more efficient than alpha-beta
on the three games studied except for the Breakthrough game. The state
of the art at Breakthrough is Unbounded Minimax. These algorithms
seem to require a cheap evaluation function (since two searches are
nested and MCTS requires many iterations to converge, MCTS is usually
used with an iteration number between 1000 and 10,000, sometimes even
up to around 100,000~\citep{baier2018mcts}), but in the context
of the state of the art, evaluation functions (which are neural networks)
are very expensive. The context of their study is that of cheap and
low-level evaluation functions: it is therefore very likely that in
the context of this paper, minimax obtains much stronger results.
Their best hybridization is the only one among those evaluated that
has a level that exceeds the performance of alpha-beta on the Breakthrough
game.

We sought to verify the hypothesis of the need for a cheap evaluation
function of their best hybridation. We performed for each game of
this paper, a match against itself using their best hybridation of
Minimax and MCTS. This algorithm is called MCTS-IP-M-k-IR-M-k. Its
parameters are taken in order to reduce the minimax search time as
much as possible (depth 1) to maximize the number of iterations. There
is an excess of the authorized 10 seconds of search per action and
a very low number of iterations. The average data obtained throughout
these matches are available in Table \ref{tab:best-hybrid-stats}.
Thus any other choice of parameters leads to too low a number of iterations
and an excess of the search time.

As a side node, we nevertheless applied the commun evaluation protocol
of the experiments of this paper (see Section~\ref{sec:Proc=00003D0000E9dure-d'=00003D0000E9valuation-commune})
on their best hybridization with their best parameter set optimized
for the Breakthrough game (Table 3 of their paper \citep{baier2018mcts}).
However, we limited this study to the Breakthrough game. This mini-experiment
is given only as indication in order to experimentally justify the
previous arguments. The results of this mini-experiment are described
in Table~\ref{tab:Performances-at-Breakthrough-best-hybrid} (we
tested several values for the exploration coefficient $C$ of the
MCTS part (see Section~\ref{subsec:Algorithme-MCTS}) since its value
was not provided in their paper). Their algorithm lost every matches
(it is as bad as the base MCTS version on this game, see Table~\ref{tab:Tuning-MCTS-10s-without-child-batching}).

In summary, these hybridations are in general inapplicable in the
current state of knowledge, in the context of the state of the art.
Research on them would have to continue so that they can maybe become
truly applicable and efficient. A change in the state of the art regarding
evaluation functions could, however, make them applicable.

\subsubsection{Game resolution}

Note that there are other search algorithms dedicated not to finding
the best possible strategy under resource constraints (which is the
subject of this article) but to fully solving a game. This is only
possible with very small games or at the end of the match, in order
to find the optimal strategy at that time. We can obviously apply
minimax type algorithms for this resolution problem. But dedicated
algorithms have also been proposed: the Proof-Number Search algorithm
and its variants~\citep{allis1994proof,kishimoto2012game}, adaptations
of the Monte Carlo Tree Search~\citep{cazenave2021monte,winands2008monte}
as well as others~\citep{van2002games}.

\subsection{Symbols, methods and notations\label{subsec:Symboles-et-notations}}

Table~\ref{tab:Index-of-symbols} describes the symbols, methods
and notations used within this article.

\subsection{Transversal techniques\label{sec:Techniques-transversales} }

In this section, we present different transversal techniques that
are combined with some algorithms evaluated in this article.
These techniques are \emph{transposition tables} (Section~\ref{subsec:Table-de-transposition}),
\emph{iterative deepening} (Section~\ref{subsec:Iterative-Deepening}),
\emph{move ordering} (Section~\ref{subsec:Ordering-of-actions}), \emph{discretization
of decimal evaluation functions} (Section~\ref{subsec:Discr=00003D0000E9tisation-de-fonctions},
\emph{Child Batching} (Section~\ref{subsec:Child-Batching}), \emph{semi-completed
evaluation functions} (Section~\ref{subsec:Semi-Completed-Evaluation}),
\emph{states resolution} (Section~\ref{subsec:Completed-Evaluation}), as
well as \emph{undoing} (Section~\ref{subsec:Undoing}).

\subsubsection{Transposition table \label{subsec:Table-de-transposition}}

A transposition table \citep{greenblatt1988greenblatt} is a hash
function that associates with a game state all the information collected
about this state during the search. This avoids recalculating the
information of a state when it is encountered later during the search.
This allows the game tree to be transformed into a directed acyclic
graph, which therefore simplifies the state space. All the algorithms
in this article use transposition tables. Note that the transposition
tables are kept from one round to the next.

\subsubsection{Iterative Deepening\label{subsec:Iterative-Deepening}}

The\emph{ iterative deepening }technique \citep{korf1985depth} allows
one to use a search algorithm set by a search depth with a time limit.
It sequentially performs increasing depth searches as long as there
is time. All algorithms in this article parameterized by a search
depth use this technique.

\subsubsection{Move ordering\label{subsec:Ordering-of-actions}}

The \emph{move ordering} technique \citep{fink1982enhancement} is
a technique that involves exploring the children of a state in a particular
order: the most interesting moves are explored first. Move ordering,
combined with iterative deepening, consists in exploring the actions
in the order of their values provided by the previous search. It accelerates
the next search (the one at an incremented depth), because a good
actions order allows for larger alpha-beta prunings. In this paper,
all alpha-beta algorithms with bounded depth or using an alpha-beta
algorithm with bounded depth use move ordering.

\subsubsection{Discretization of decimal evaluation functions\label{subsec:Discr=00003D0000E9tisation-de-fonctions}}

Here we describe a classical and basic technique to transform a decimal
evaluation function into an integer-valued evaluation function. A
minority of the algorithms evaluated in this paper require the use
of an integer-valued evaluation function (namely MTD(f) and PVS).
This simply consists in multiplying the decimal value by a constant
greater than 1 and then rounding~\citep{mandziuk2004alpha}. Here
we adopt a slightly more complicated procedure, since we first normalize
the evaluation functions so that they are values in $[-1,1]$ . This
normalization is thus different for each game but also for each evaluation
function based on a different reinforcement heuristic (see Section~\ref{subsec:Fonctions-d'=00003D0000E9valuation-utilis=00003D0000E9s}
for the definition of a reinforcement heuristic). Indeed, the range
of values of the evaluation function are determined from the reinforcement
heuristic used for its learning (the learned evaluation function and
its reinforcement heuristic have the same arrival domain). The classical
discretization procedure is then applied, which allows to have an
analogous discretization for each game. The normalization-discretization
formula that we used is the following: 
\[
D_{f}(s)=\mathrm{round}\left(\frac{\pdiscret\cdot f\left(s\right)}{\mathrm{max}\left(\left|M\right|,\left|m\right|\right)}\right)
\]
with $\pdiscret$ a discretization constant (a parameter of this procedure),
$s$ a game state, $M$ the maximum practical value of the reinforcement
heuristic $h_{f}$ used for learning $f$ and $m$ the minimum practical
value of the reinforcement heuristic $h_{f}$ used for learning $f$.
More formally, $M=\max V$ and $m=\min V$ with $V=\liste{h_{f}\left(t\right)}{t\in T}$
where $T$ is a set of random endgame states containing $10^{6}$
elements.

When an algorithm requires an integer-valued evaluation function,
we therefore use the function $D_{f}$ instead of the evaluation function
$f$.

\subsubsection{Child Batching\label{subsec:Child-Batching}}

The technique we call Child Batching consists in evaluating in parallel
the children of a state when its children must be evaluated sequentially
by an evaluation function. We have not found any reference to this
technique in the literature, except in the article~\citep{2020learning},
although it would be surprising if it had not been used earlier to
improve Alpha-Beta.

As we will see, this technique is particularly interesting with the
use of graphics cards and neural networks, since the tensor representations
of  child states can be concatenated or \textquotedbl batched\textquotedbl{}
to be evaluated by the neural network in parallel simultaneously on
a single graphics card. Note that, Child Batching is  not applicable with Monte Carlo algorithms because  these algorithms never evaluate the children of a same state
sequentially.

Remark that Child Batching is a lossless, natural, and fixed parallelization
(the degree of parallelization corresponds to the number of children,
it is not possible to increase it). Conversely, there have been many
studies on the parallelization of search algorithms with an unfixed
degree of parallelization (variable number of threads). These approaches
notably performs parallelizations on the entire algorithm and not
only on the evaluation of the children of a state. However, these
parallelizations suffer from a loss because unnecessary calculations
are performed (which is not the case with Child Batching). The parallelization
(with loss) of alpha-beta has notably been studied \citep{singhal2019comparative,powley1990parallel,finkel1982parallelism,hsu1989large}.
Three types of parallelizations in the framework of Monte Carlo Tree
Search have been studied: root parallelization, leaf parallelization,
tree parallelization \citep{chaslot2008parallel}. There is only one
study on parallelization of Unbounded Minimax \citep{shoham2002parallel},
it corresponds to the root parallelization of Monte Carlo Tree Search.

Also note that in the context of state evaluations by neural networks,
performing parallelization to a greater degree than Child Batching
seems to be particularly expensive. When this is not the case, for
example, for games with a very small branching factor, it would then
be interesting to combine Child Batching with global parallelization,
but that is the subject of another study.

\subsubsection{Semi-Completed Evaluation\label{subsec:Semi-Completed-Evaluation}}

In \citep{2020learning}, the semi-completed evaluation technique
is proposed. With Athénan, reinforcement learning of a heuristic evaluation
function $\fadapt$ is based on an endgame state evaluation function
$\fterminal$. The technique of semi-completed evaluation consists
of using the function $f\left(s\right)=\begin{cases}
\fterminal\left(s\right) & \text{if }\ended s\\
\fadapt\left(s\right) & \text{otherwise}
\end{cases}$ instead of the function $\fadapt$ to evaluate states. This technique
is used with all the algorithms in this paper.

\subsubsection{Integrated Basic Solver\label{subsec:Completed-Evaluation}}

Search algorithms using the learned evaluation function $\fadapt$
(or even the corresponding semi-completed evaluation function) do
not always make the best decision because they do not take into account
the fact that some states are \emph{solved} (a state is solved when
its minimax value is that of the complete game subtree starting from
that state). For example, an algorithm may decide to play an unsolved
state that is perhaps losing rather than a solved winning one. This
happens when the unsolved state has a partial minimax value higher
or equal than the \emph{complete} minimax value of the solved state~\citep{2020learning}.
In fact, even some algorithms fail to solve a game with an infinite
search time for similar reasons~\citep{cohen2021completeness,2020learning}.

We use a simple solver combined with each algorithm in this paper.
Each time a state is analyzed, we check if it is solved by looking
whether all its children are solved or whether one of its children
is solved winning for the current player. At the end of the search,
a solved winning state is always played and if possible a solved losing
state is never played. This basic solving algorithm is described in
Figure~\ref{alg:common_solver}. With Monte Carlo Tree Search, there
is a more complex built-in solver~\citep{winands2008monte}. We tested
it, the performance decreased. It is because of the use of another
decision criterion (\emph{secure decision}~\citep{winands2008monte}).
We tested it with the standard MCTS decision criterion, its performance
is indistinguishable with that of our basic solver (the confidence
intervals overlap strongly). Anyway, in the experiments of this paper,
using a solver improves performance by at most a few percent.

\subsubsection{Undoing\label{subsec:Undoing}}

A naive implementation of search algorithms involves copying game
states when exploring the game tree. This is expensive in terms of
space and time. One way to get around this problem is to have an $\mathrm{undo}\left(s\right)$
method that undoes the last move played. This trick is used with all
the algorithms in this article.

\subsection{Common evaluation protocol\label{sec:Proc=00003D0000E9dure-d'=00003D0000E9valuation-commune}}

In this section, we present the common framework for the experiments
in this paper. We begin by presenting the machines used. Then we detail
the games used as benchmarks to compare the different search algorithms.
(Section~\ref{subsec:Jeux}). Next, we present the evaluation functions
used by the search algorithms (Section~\ref{subsec:Fonctions-d'=00003D0000E9valuation-utilis=00003D0000E9s}).
Afterwards, we present the standard Unbounded Minimax algorithm, the
benchmark adversary algorithm that we have chosen for comparing all algorithms
with respect to their performance against this benchmark adversary
(Section~\ref{subsec:Minimax-non-born=00003D0000E9-de-r=00003D0000E9f=00003D0000E9rence}).
Later, we detail the procedure for evaluating the different search
algorithms (Section~\ref{subsec:=00003D0000C9valuation-par-tournois}).
Finally, we define what the performance of a search algorithm is in
this paper, refining the win rate to take into account the impact
of draws (Section~\ref{subsec:Gain-d'un-algorithme}).

\subsubsection{Computer used}

During the experiments in this article, two types of machines were
used.

The evaluation functions were learned on the V100 partition of the
Jean-Zay supercomputer. A node is based on 2 Intel Cascade Lake 6248
processors (20 cores at 2.5 GHz), i.e. 40 cores per node, 192 GB of
memory, 4 Nvidia Tesla V100 SXM2 32 GB GPUs. Note that learning an
evaluation function uses a quarter node (and thus only on GPU).

The matches were carried out with the MI250X partition of the Adastra
supercomputer. A node is based on 1 AMD Trento EPYC 7A53 (64 cores
2.0 GHz processor), 256 Gio of DDR4-3200 MHz CPU memory, 4 Slingshot
200 Gb/s NICs, 8 GPUs devices (4 AMD MI250X accelerator, each with
2 GPUs) with a total of 512 Gio of HBM2. Note that 8 matches are carried
out in parallel per GPU.

\subsubsection{Games\label{subsec:Jeux}}

We now briefly present the 22 games on which our experiments are performed,
namely: Amazons, Arimaa, Ataxx, Breaktrhough, Clobber, Connect6, Draughts
(International, Brazilian, and Canadian), Chess, Go (board size 9x9),
Outer Open Gomoku, Hex (board size 11x11, 13x13, and 19x19), Havannah,
Lines of Action, Othello (board size 8x8 and 10x10), Santorini, Surakarta, and
Xiangqi. All of these games are present and recurring at the Computer
Olympiad (except Chess which has its own event), the worldwide multi-games
event in which computer programs compete against each other. In addition,
all these games and their rules are included in Ludii~\citep{Piette2020Ludii},
a free general game system. Additional information on these games
can also be found at \url{https://boardgamegeek.com}.

\paragraph{Amazons}

Amazons is a game of movement and blocking. On their turn, a player
moves one of their pieces in a straight line in any direction, similar
to the queen in Chess. After moving, the player places a neutral piece
in any direction from the new position of the moved piece, again similar
to the queen's movement in Chess. The neutral pieces block movement
and placement of other pieces. The objective is to be the last player
able to make a move.

\paragraph{Arimaa}

Arimaa is a game with similarities with Chess. Each player has a different
set of pieces (identical between the two players). The pieces types
are fully ordered. A higher piece can move (push or pull) a lower
piece. If a piece is on a trap square without being adjacent to a
friendly piece, it is removed from the board. A player wins when their
weakest piece reaches the other side of the board.

\paragraph{Ataxx}

Ataxx is originally a video game. On their turn, a player can move
one of their pieces with a distance of one or two. If it is moved
by one, it duplicates itself. In all cases, the other player's pieces
adjacent to the moved piece are converted: they now belong to the
player who has just played.

\paragraph{Breakthrough}

Breakthrough is a game of movement and capture, where the goal is
to be the first to get one of their pieces to the opposite side of
the board. Pieces can only move forward one square, either straight
ahead or diagonally. Captures are made by moving diagonally onto a
square occupied by an opponent's piece.

\paragraph{Draughts}

The game of Draughts consists in moving their pieces diagonally by
one, or more if captures are possible by jumping iteratively over
opponent's pieces. If there is no capture, a piece can only move forward
(by one). The objective is to take all the opposing pieces. When one
of their pieces reaches the opposite board side (of the opponent),
this piece is promoted into a king piece. It can then move in a straight
line as many squares as desired and in all diagonal directions (even
backwards). When a capture is possible, it is obligatory.

The three used version Draughts differ only in their board size (and
their initial number of pieces): Brazilian Draughts has an 8x8 board,
International Draughts has a 10x10 board, Canadian Draughts has a
12x12 board.

\paragraph{Clobber}

Clobber is a game of movement and capture, where the objective is
to be the last player able to make a move. A player can make a move
by orthogonally moving one of their pieces onto an adjacent square
occupied by an opponent's piece. This move results in the capture
and removal of the opponent's piece from the board.

\paragraph{Connect6}

Connect6 is a connection game where the goal is to align at least
six pieces of the same color. On each turn, a player places two pieces
of their color on the board. During the first turn, the first player
places only one piece.

\paragraph{Chess}

The Chess game is played on a $8\times 8$ square board. Each player has several types
of pieces, each with a particular movement. A piece can end its movement
on the case of an opponent's piece, the latter is then removed from
the board. The aim of the game is to immobilize a particular piece
of the opponent, called the king. This piece cannot move if its movement
brings it to a case where it could be captured by the opponent during
their next turn.

\paragraph{Go}

The game of Go is played by placing one of their pieces on the board
in turn. When a related group of their pieces is completely surrounded
by the opponent's pieces, it is removed from the board. The player
who wins is the one who has the most pieces on the board at the end
of the game.

\paragraph{Outer Open Gomoku}

Outer Open Gomoku is a connection game where the goal is to align
at least five pieces of the same color. On each turn, a player places
a piece of their color on the board. During the first turn, the first
player is restricted to placing their piece at most two spaces away
from the edges of the board.

\paragraph{Hex}

Hex is a game played on an $n\times n$ hexagonal
board, with common sizes being $11\times11$, $13\times13$, and $19\times19$.
Each player takes turns placing a stone of their color on an empty
cell. The objective is to be the first to connect the two opposite
sides of the board corresponding to their color. Despite its simple
rules, Hex involves complex tactics and strategies. The game has a
vast number of possible states and actions per state, comparable to
Go. For example, on an $11\times11$ board, the number of possible
states surpasses that of Chess (Table~$6$ of \citep{van2002games}).
The first player always has a winning strategy \citep{berlekamp2018winning},
although it is only known for board sizes up to $10\times10$ (game
is weakly solved up to this size \citep{powley1990parallel}
%; the reference can be found in \citep{2020learning}
).
Determining the outcome of a particular state is PSPACE-complete \citep{bonnet2016complexity}.

A common variant of Hex includes the swap rule, where the second player
can choose, on their first turn, to swap roles and sides with the
first player. This rule helps balance the game by mitigating the first
player's advantage and is typically used in competitions. It is still
used at the Computer Olympiad, and we use it in this paper.

\paragraph{Havannah}

Havannah is a game similar to Hex. The differences are as follows.
It is played on a hexagonal shaped board with hexagonal pieces (unlike
a diamond board with hexagonal pieces for Hex). Edges are not associated
with any player. To win, simply connect three edges together or two
corners together but also make a loop (each time with only their pieces).
The swap rule is also used.

\paragraph{Lines of Action}

Lines of Action is a game focused on movement and connection. On their
turn, a player moves one of their pieces in a straight line, 
as many squares as there are pieces in that direction (pieces of any
player). A piece cannot pass through an opponent's piece, but it
can capture it by landing directly on its square. The objective is
to connect all of their pieces into a single continuous group.

\paragraph{Othello}

Othello, also known as Reversi, is a game of territory and encirclement,
with the goal of having more pieces on the board than their opponent.
On their turn, a player places one of their pieces on the board, but
only if it creates an encirclement; otherwise, they must pass their
turn. An encirclement occurs when a continuously aligned group of
opponent's pieces is surrounded at both ends by the player's pieces
(one being the newly placed piece and the other an existing piece).
When this happens, all encircled opponent pieces are replaced by their
pieces.

\paragraph{Santorini}

Santorini is a three-dimensional game of building and movement. The
objective is to be the first player to reach the 3rd floor of a building.
On their turn, a player moves one of their pieces to an adjacent square,
then either places the first floor on an empty adjacent square or
adds one floor to an existing structure (as long as no player's piece
occupies it). A piece can only move up one floor at a time but can
descend any number of floors. Moves cannot be made to a square with
four floors, and construction cannot occur on a square that already
has four floors (the fourth floor is the roof). A player who cannot make a valid move loses. Advanced
mode, which includes power cards, is not used in the experiments of
this article.

\paragraph{Surakarta}

Surakarta is a game of movement and capture, similar to Draughts,
with the objective of capturing all the opponent's pieces. On their
turn, a player can either move a piece to an adjacent empty square
or capture an opponent's piece by following specific conditions.
The capture mechanism in Surakarta is unique, involving a special
movement circuit designed exclusively for capturing, enabling long-distance
captures. When the game ends, the winner is the one with the most
pieces, for example in the event of a premature end due to repetition.

\paragraph{Xiangqi}

Xiangqi also called Chinese Chess is a game similar to Chess, its
rules are very close. Its main differences are different movements
for the pieces and the presence of a special area on the board (the
palace) with the following rule: the king cannot leave its palace.

\subsubsection{Evaluation functions used\label{subsec:Fonctions-d'=00003D0000E9valuation-utilis=00003D0000E9s}}

We now present the evaluation functions used by each of the search
algorithms studied in this article.

The evaluation functions were generated using Descent, the Athénan's
evaluation function learning algorithm (6). There are six sets of
evaluation functions. Each set contains the same number of evaluation
functions. The details of the numbers of evaluation functions per
game are described in the Table~\ref{tab:number_heuristic}. The
details of the Athénan hyperparameters used to generate each evaluation
function of the first set are described in Table~\ref{tab:hyperparameter_heuristics},~\ref{tab:hyperparameter_heuristics-1},~\ref{tab:hyperparameter_heuristics-2}~and~\ref{tab:hyperparameter_heuristics-2-1}
(we use the following notations: search time per action $t$, batch
size $B$, duplication factor $\delta$, number of neural network
weights $W$, number of residual blocks in the network $R$, number
of dense blocks in the network $D$ ; see \citep{2020learning} for
the definitions of these Athénan hyperparameters ; we also denote
``reinforcement heuristic used'' by ``R.H'' and the number of
time this hyperparameters combination is used by ``uses''. 
Note that their parameters are heterogeneous because we reuse evaluation functions generated for other contexts to avoid having to generate others since this is very costly.

Each of these evaluation functions is the result of 20 days of training.
The second set of evaluation functions is obtained from the first
set by carrying out $10$ additional training days (keeping the same
hyperparameters: the learning process was simply continued). This
construction is identical for the four other sets: the sixth set is
the result of the continuation of the first set evaluation functions after
a total of 70 training days with Descent.

\subsubsection{Benchmark adversary algorithm\label{subsec:Minimax-non-born=00003D0000E9-de-r=00003D0000E9f=00003D0000E9rence}}

We now detail the algorithm used as a benchmark adversary within the
experiments in this article. We start by motivating the choice of
Unbounded Minimax as the benchmark algorithm, then we present it.

\paragraph{Choice of this referent}

We chose as benchmark algorithm the version of Unbounded Minimax used
by Athénan~\citep{2020learning}, state of the art of two-player
game algorithms with perfect information. 
%However, we take this algorithm without its default option: the safe decision strategy. Note that this algorithm with the safe decision is however still evaluated in this article (see Section~\ref{subsec:Minimax-non-born=00003D0000E9-safe}).
This algorithm %(without safe decision) 
is a very slight variation
of Unbounded Minimax of Korf \& Chickering~\citep{korf1996best},
discovered independently (the difference relates to a particular case
which only impacts performance by $1\%$). Note that Korf \& Chickering's
Unbounded Minimax is called Best-First Minimax Search, but this can
be confusing because there are several Best-First algorithms in the
literature \citep{plaat1996best}, but they are all of limited depth.
A denomination by Unbounded Best-First Minimax is therefore used in
\citep{2020learning}, which we also do in this article. Since there
is no ambiguity in this article, we will talk about Unbounded Minimax
and we denote it by $\ubfm$.

\paragraph{Benchmark Unbounded Minimax algorithm}

We now present the version of Unbounded (Best-First) Minimax, $\ubfm$,
used as the benchmark adversary algorithm, that is to say as the opponent
in matches for evaluating the studied search algorithms. $\ubfm$
iteratively extends the game tree by adding the children of one of
the leaves of the tree. The chosen leaves are the states obtained
after having played one of the best sequences of possible actions
given the current partial knowledge of the game tree. Thus, that algorithm
iteratively extends the \emph{a priori} best sequences of actions.
The best sequences usually change at each extension. Thus, the game
tree is non-uniformly explored by focusing on the \emph{a priori}
most interesting actions without exploring just one sequence of actions.
Note that when the search time is exhausted, the action with the best
value is returned. $\ubfm$ is formally described in Figure~\ref{alg:-Unbounded-Best-First-Mininmax-ref}.

\subsubsection{Matches to evaluate algorithms\label{subsec:=00003D0000C9valuation-par-tournois}}

In this section we describe the entire procedure for evaluating the
search algorithms studied in this article.

We use as benchmark adversary $\ubfm$ (see Section~\ref{subsec:Minimax-non-born=00003D0000E9-de-r=00003D0000E9f=00003D0000E9rence})
using the lossless parallelization procedure Child Batching (see Section~\ref{subsec:Child-Batching}).
The benchmark algorithm uses Child Batching even if the algorithm
evaluated against it does not use it (which allows the performance
of each algorithm to be compared with each other).

We evaluate each algorithm on the 22 selected games (see Section~\ref{subsec:Jeux}).
There are 6 sets of evaluation functions per game, each containing
approximately fifteen evaluation functions (see Section~\ref{subsec:Fonctions-d'=00003D0000E9valuation-utilis=00003D0000E9s}).
The evaluation procedure is repeated for each of these six sets. Each
algorithm faces the benchmark adversary twice (once as first player
and once as second player) and this for each pair of evaluation functions
in the current set of evaluation functions. This corresponds to approximately
450 matches for each evaluated algorithm per repetition and per game.
This evaluation is thus repeated 6 times with a different set of evaluation
functions. Each performance of a search algorithm is therefore the
average over about 2700 matches per game. The exact total number of
matches performed per repetition and in total for each game is described
in Table~\ref{tab:number_of_matches}.

This experimental protocol was applied twice in this article: once
with a search time of 10 seconds and once with a search time of 1
second. Each algorithm evaluated within one this two comparisons use
the same search time, as the benchmark adversary algorithm.

Finally, the performance of the different algorithms are evaluated
with the metrics described in the following section.

\subsubsection{Performance of a search algorithm\label{subsec:Gain-d'un-algorithme}}

We end this section with the description of the performance criteria
used to discriminate between the different search algorithms studied.

To evaluate a search algorithm, we aggregate the result of each of
its matches as follows. The performance on a game of a search algorithm
is the averaged results of performed matches over this game with that
search algorithm: 1 in case of victory, -1 in case of defeat, and
0 in case of a draw. We give the 5\% confidence radius associated
with all of these values. We also give the average of the results
over all games as an overall evaluation. Moreover, in order to have
a global unbiased confidence interval, we provide the stratified bootstrapping
confidence interval (at 5\%).

The percentages detailing performance for each game have been rounded
to the nearest percent to reduce size and improve readability of data
tables.

\subsection{Algorithms and their tuning\label{subsec:Algorithms-and-their-tuning}}

In this section, we describe, formalize and tune within our common
evaluation protocol the different studied algorithms: Minimax Alpha-Beta
pruning (Section~\ref{subsec:Alpha-B=00003D0000EAta}), Principal
Variation Search (Section~\ref{subsec:Principal-Variation-Search}),
MTD(f) (Section~\ref{subsec:MTDf}), Minimax algorithm with Alpha-Beta
and $k$-best pruning (Section~\ref{subsec:k-best-pruning-alpha-b=00003D0000EAta}),
Monte Carlo Tree Search (Section~\ref{subsec:Algorithme-MCTS}) as
well as its variant $\mctsh{}$ which uses a state evaluation function
instead of rollouts (Sections~\ref{subsec:Neural-value-based-MCTS}),
%Unbounded Minimax with safe decision (Section~\ref{subsec:Minimax-non-born=00003D0000E9-safe}), the randomized version of the Unbounded Minimax of the literature, i.e. RBFM (Section~\ref{sec:Minimax-non-born=00003D0000E9-randomis=00003D0000E9}) and our new algorithm $\orsum$, instantiation and modification of RBFM (Section~\ref{sec:Minimax-non-born=00003D0000E9-randomis=00003D0000E9-am=00003D0000E9liorer}). 
%In Section~\ref{sec:softmax_and_greedy}, we describe the two standard actions probability distribution: $\epsilon$-greedy and softmax. The last section focuses in detail on the tuning of RBFM for several probability distributions, showing the interest of our new, ordinal, distribution and justifying for the first time in a robust and general way, the interest of RBFM (Section~\ref{sec:tuning_of_Randomized_Unbounded_Minimax}).

\subsubsection{Alpha-Beta\label{subsec:Alpha-B=00003D0000EAta}}

In this section, we focus on the Alpha-Beta algorithm. First, we describe
it. Then, the transversal techniques used in conjunction with Alpha-Beta
during the experiments are detailed.

\paragraph{Alpha-Beta algorithm \label{subsec:Algorithme-alpha-beta}}

The Minimax algorithm with Alpha-Beta pruning \citep{knuth1975analysis}
or more simply the Alpha-Beta algorithm, denoted also $\alphabeta$,
is an improvement of the Minimax algorithm. It reduces the search
space by avoiding exploring the states which do not influence the
value of the root by using the properties of the min and the max.
More precisely, the algorithm has two bounds $\alpha$ and $\beta$
which define the interval in which the value of the root is found
given the values encountered in the previous states during the search.
At the start of the algorithm, $\alpha=-\infty$ and $\beta=+\infty$.
All actions leading to states whose value is not in the range are
pruned. Figure~\ref{alg:Algorithme-AB} describes the Alpha-Beta
algorithm. This algorithm depends on one parameter: the search depth
$d$.

It is not possible in general to carry out a search on the entire
game tree. For example, in Chess, the number of states is $4.8\times10^{44}$.
This algorithm (like the basic Minimax algorithm) therefore limits
the search to the tree depth $d$, that is to say, it only explores
the possible futures of the match at most $d$ rounds in advance.

\paragraph{Dedicated Alpha-Beta Improvements\label{subsec:Am=00003D0000E9liorations-de-AB}}

We now detail the transversal techniques used with alpha-beta within
the experiments of this article. We use Iterative Deepening (Section~\ref{subsec:Iterative-Deepening})
in order to make this algorithm anytime, which therefore allows the
use of a search time. Move ordering (Section~\ref{subsec:Ordering-of-actions})
is also used to make more efficient pruning. Except when the contrary
is explicitly specified, we still use Child Batching (Section~\ref{subsec:Child-Batching}).
The Alpha-Beta algorithm with Child Batching is described in Figure~\ref{alg:Algorithme-AB-Child-Batching}.
Like all algorithms implemented and evaluated in this paper, Alpha-beta
uses transposition tables. The Alpha-Beta algorithm with transposition
table is described in Figure~\ref{alg:Algorithme-AB-transpositions}.

\paragraph{Alpha-beta tuning\label{subsec:Evaluations-AB}}

This algorithm has no parameters, so there is no tuning.

\subsubsection{Principal Variation Search\label{subsec:Principal-Variation-Search}}

In this section, we focus on the Principal Variation Search algorithm,
a.k.a. $\pvs{}$. First, we describe it. Then, the transversal techniques
used in conjunction with $\pvs{}$ during the experiments are detailed.
Finally, the tuning of $\pvs{}$ is performed within our common evaluation
protocol (see also Section~\ref{sec:Proc=00003D0000E9dure-d'=00003D0000E9valuation-commune}).

\paragraph{$\protect\pvs{}$ algorithm\label{subsec:Algorithme-PVS}}

The Principal Variation Search algorithm (also called the Scout algorithm)~\citep{pearl1980scout,pearl2022asymptotic,marsland1982parallel},
acronym $\pvs{}$, is a variant of the Alpha-Beta algorithm. This
algorithm assumes that the first child evaluated is the best child.
Each of the other children is evaluated by a restricted search that
uses this assumption so that this restricted search is faster than
the full search it replaces. If a child evaluated by a restricted
search is not worse than the best of the previous children, then a
full search is performed on that child. If the ordering of the children
is good enough, this makes the algorithm better than Alpha-Beta. If
the ordering of the children is very bad, the algorithm will be worse
than Alpha-Beta since it will perform many restricted searches for
nothing in addition to the same full searches. Figure~\ref{alg:Algorithme-PVS}
described $\pvs{}$.

\paragraph{$\protect\pvs{}$ Improvements\label{subsec:Am=00003D0000E9liorations-de-PVS}}

We thus detail the transversal techniques used with $\pvs{}$ within
the experiments of this article. We use the same improvement techniques
for $\pvs{}$ as for standard Alpha-Beta, namely: Iterative Deepening
(Section~\ref{subsec:Iterative-Deepening}) in order to make this
algorithm anytime; move ordering (Section~\ref{subsec:Ordering-of-actions})
to perform more efficient pruning (and also obviously to apply PVS since it basically needs a move ordering technique). Finally, since the algorithm is
designed to be used with an integer-valued evaluation function, we
therefore additionally use the evaluation function discretization
technique described in Section~\ref{subsec:Discr=00003D0000E9tisation-de-fonctions}.
The use of $\pvs{}$ is therefore parameterized by a discretization
coefficient $\pdiscret$. We still use Child Batching (Section~\ref{subsec:Child-Batching}).
The use of Child Batching is identical to that of Alpha-Beta (see
Section~\ref{subsec:Am=00003D0000E9liorations-de-AB}).

\paragraph{$\protect\pvs{}$ tuning\label{subsec:Evaluations-PVS}}

We have performed a tuning experiment for the constant $\delta$ of
$\pvs{}$ with a search time of 10 seconds per action with Child Batching,
by testing the following values: $\delta\in\left\{ 10,100,300,1000,3000,10000\right\} $.
Performance for each tested value $\delta$ is in Table~\ref{tab:Tuning-PVS-10s-with-child-batching}.
The best value is $\delta=3000$ on all games. We have also performed
a tuning for the constant $\delta$ of $\pvs{}$ with a search time
of 1 second per action, with Child Batching by testing the following
values: $\delta\in\left\{ 10,30,100,300,1000,3000\right\} $. Performance
for each tested value $\delta$ is in Table~\ref{tab:Tuning-PVS-1s-with-child-batching}.
The used values for $\delta$ in the main comparison tables of the
paper are each time the best identified for each game ($\delta$ is
game-dependent). We denote this by $\delta=*$ in the tuning tables.
Note that this tuning process, i.e. taking $\delta=*$, produces an
upper bound on the real performance. To know the real performance,
it is necessary to apply a cross-validation. However, we do not need
it, since we show that the Unbounded Minimax algorithms have better
performances than an upper bound of the $\pvs{}$ performance and
therefore than the real performance of $\pvs{}$ .

\subsubsection{Memory-enhanced Test Driver: $\protect\mtdf{}$\label{subsec:MTDf}}

In this section, we present, detail the transversal techniques used,
and tune the Memory-enhanced Test Driver algorithm, named more succinctly
$\mtdf{}$.

\paragraph{The $\protect\mtdf{}$ algorithm \label{subsec:Algorithme-MTD(f)}}

Thus, we start by describing the algorithm $\mtdf{}$~\citep{plaat1996best}.
$\mtdf{}$ is a search algorithm using an integer-valued evaluation
function. This algorithm uses alpha-beta as a sub-procedure. We recall
some basics on alpha-beta in order to be able to present their differences.
The alpha-beta algorithm uses two bounds $\alpha$ and $\beta$ to
perform its search, which define the interval in which the value of
the root is sought. The states whose values are not found in this
interval are pruned. The values of $\alpha$ and $\beta$ are defined
by the values of the states already encountered: as the search progresses
this interval becomes more precise. Initially, $\alpha=-\infty$ et
$\beta=\infty$. The idea of $\mtdf{}$ is to fix from the start an
interval of the form $[f-1,f]$ and therefore to apply Alpha-Beta
with $\alpha=f-1$ et $\beta=f$. Alpha-beta with this restricted
interval therefore applies very powerful pruning and thus explores
a drastically reduced game tree. It thus performs this search much
faster than standard Alpha-Beta (i.e. with $\alpha=-\infty$ and $\beta=\infty$
at the start). Obviously, this restricted search, called a zero-window
search, does not always calculate the right value for the root. However,
the value returned by this restricted procedure gives information
about the true value of the root. If the value is indeed in the interval
$[\alpha,\beta]=[f-1,f]$ then the value is correct. The $\mtdf{}$
algorithm is terminated. Otherwise, the returned value is an approximation
of the searched value and $f$ is reset to that value. More precisely,
if the returned value is less than $\alpha$ then the returned value
is an upper bound on the true value and if the returned value is greater
than $\beta$ then it is a lower bound on the true value. In these
last two cases, $\mtdf{}$ applies a new restricted Alpha-Beta search
with the updated interval $[f-1,f]$. This procedure is thus applied
until the value of the root is found. In other words, the space of
possible values of the root is explored in such a way that recursively
applying the restricted search reaches the true value of the root.
$\mtdf{}$ is described in Figure~\ref{alg:Algorithme-MTDf}. Note
that initially, $f$ is the value of the previous $\mtdf{}$ search
which is generally a quite good approximation of the value of the root,
which speeds up the search.

\paragraph{Improvements used with $\protect\mtdf{}$\label{subsec:Am=00003D0000E9liorations-utilis=00003D0000E9es-MTDf}}

We thus detail the transversal techniques used with $\mtdf{}$ within
the experiments of this article. We use the same improvement techniques
for $\mtdf{}$ as for standard Alpha-Beta and for $\pvs{}$, namely:
Iterative Deepening (Section~\ref{subsec:Iterative-Deepening}) in
order to make this algorithm anytime; move ordering (Section~\ref{subsec:Ordering-of-actions})
to perform more efficient pruning. Finally, since the algorithm is
designed to be used with an integer-valued evaluation function, such
as $\pvs{}$, we therefore additionally use the evaluation function
discretization technique described in Section~\ref{subsec:Discr=00003D0000E9tisation-de-fonctions}.
The use of $\mtdf{}$ is therefore, like $\pvs{}$, parameterized
by a discretization coefficient $\pdiscret$. We still use Child Batching.
The use of Child Batching is identical to that of Alpha-Beta (see
Section~\ref{subsec:Am=00003D0000E9liorations-de-AB}).

Note that the use of Iterative Deepening with $\mtdf{}$ is optimized
as follows: the initial value $f$ of $\mtdf{}$ is the value of the
root determined by the previous search of $\mtdf{}$ performed by
Iterative Deepening.

\paragraph{$\protect\mtdf{}$ tuning\label{subsec:Evaluations-de-MTD(f)}}

We have performed a tuning experiment for the constant $\delta$ of
$\mtdf{}$ with a search time of 10 seconds per action with Child
Batching, but without the basic solver (see Section~\ref{subsec:Completed-Evaluation}).

This is the only exception to its non-use in this article. It was
an oversight. It is not an important detail (the solver only improves
performance by a few percentages). The basic solver is though used
with $\mtdf{}$ in the main comparison tables (Tables~\ref{tab:res-avec-batching}~and~\ref{tab:Performance-main-10s-with-child-batching}).

We tested the following values: $\delta\in\left\{ 10,30,100,300,1000,3000\right\} $.
Performance for each tested value $\delta$ is in Table~\ref{tab:Tuning-MTDf-10s-with-child-batching}.
The best value is $\delta=100$ on all games. We have also performed
a tuning for the constant $\delta$ of $\mtdf{}$ with a search time
of 1 second per action, with Child Batching, and with the basic solver
by testing the following values: $\delta\in\left\{ 10,30,100,300,1000,3000\right\} $.
Performance for each tested value $\delta$ is in Table~\ref{tab:Tuning-MTDf-1s-with-child-batching}.
The used values for $\delta$ in the main comparison tables of the
paper are each time the best identified for each game ($\delta$ is
game-dependent). We denote this by $\delta=*$ in the tuning tables.
Note that this tuning process, i.e. taking $\delta=*$, produces an
upper bound on the real performance. To know the real performance,
it is necessary to apply a cross-validation. However, we do not need
it, since we show that the Unbounded Minimax algorithms have better
performances than an upper bound of the $\mtdf{}$ performance and
therefore than the real performance of $\mtdf{}$ .

\subsubsection{$k$-best pruning alpha-beta\label{subsec:k-best-pruning-alpha-b=00003D0000EAta}}

In this section, we focus on improved Alpha-Beta with the $k$-best
pruning technique. First, we detail the algorithm. Then, we tune this
algorithm, varying the number $k$. Note that the same transversal
techniques as Alpha-Beta are used (see Section~\ref{subsec:Am=00003D0000E9liorations-de-AB}).

\paragraph{Alpha-Beta algorithm with $k$-best pruning\label{subsec:Algorithme-kbest-alpha-b=00003D0000EAta}}

The alpha-beta algorithm improved with $k$-best pruning technique~\citep{baier2018mcts}
consists in applying the Alpha-Beta algorithm by pruning the actions
that are not part of the $k$ best actions. Let us remember that the
available actions (method $\Actions s$) are sorted by their value
obtained when searching at the previous lower depth of Iterative Deepening
(since move ordering technique is used). Thus, only the current $k$-best
actions are kept. Its implementation is simple, $\Actions s$ is replaced
by $\Actions s$ restricted to the $k$ first best actions in the
algorithm of Figure~\ref{alg:Algorithme-AB-Child-Batching}.

Note that this modification causes Alpha-Beta to lose its completeness
property, i.e. if we take $d=\infty$, this variant does not necessarily
find the optimal strategy. However, in the practical case, since we
cannot take $d=\infty$, this pruning allows one to take a bigger
$d$ than with standard Alpha-Beta.

\paragraph{$k$-best pruning alpha-beta tuning\label{subsec:Evaluation-de-k-best}}

We have performed a tuning experiment for the constant $k$ of $k$-best
$\alphabeta$ with a search time of 10 seconds per action and with
Child Batching by testing the following values: $k\in\left\{ 2,3,4,5,6,7,10,15,30,\infty\right\} $.
Performance for each tested value $k$ is in Table~\ref{tab:Tuning-k-best-10s-with-child-batching-part-1}
(mean) and in Table~\ref{tab:Tuning-k-best-10s-with-child-batching-part-2}
(details). We have also performed a tuning for the constant $k$ of
$k$-best $\alphabeta$ with a search time of 1 second per action
and with Child Batching. We tested the following values: $k\in\left\{ 2,3,4,5,6,7,8,9,10,11,12,15,30,60\right\} $.
Performance for each tested value $k$ is in Table~\ref{tab:Tuning-k-best-1s-with-child-batching-part-1}
(mean) and in Table~\ref{tab:Tuning-k-best-1s-with-child-batching-part-2}
and Table~\ref{tab:Tuning-k-best-1s-with-child-batching-part-3}
(details). The used values for $k$ in the main comparison tables
of the paper are each time the best identified for each game ($k$
is game-dependent). We denote this by $k=*$ in the tuning tables.
Note that this tuning process, i.e. taking $k=*$, produces an upper
bound on the real performance. To know the real performance, it is
necessary to apply a cross-validation. However, we do not need it,
since we show that the Unbounded Minimax algorithms have better performances
than an upper bound of the $k$-best $\alphabeta$ performance and
therefore than the real performance of $k$-best $\alphabeta$ 
(except for 5 of the games when a search time of 10 seconds is used, where indeed, we cannot rigorously conclude that $k$-best Alpha-Beta is better than Unbounded Minimax algorithms, even though it seems very likely).

\subsubsection{Basic Monte Carlo Tree Search\label{subsec:Algorithme-MCTS}}

\paragraph{$\protect\mcts{}$ algorithm}

The Monte Carlo Tree Search algorithm ($\mcts{}$ for short)~\citep{Coulom06,browne2012survey}
is an unbounded best-first search algorithm that evaluates game states
by their win statistic. It explores the most promising states first
where the interest of a state is the sum of that state's win statistic
plus an exploration term in order to handle the exploration/exploitation
dilemma~\citep{mandziuk2010knowledge}. At each iteration, it goes
down the game tree by choosing the most interesting action until reaching
a state whose one of its children is not in the tree. It then randomly
adds one of the missing children of this state within the tree. Next,
it performs a random simulation of the rest of the match starting
from the new added state. Afterwards, it backpropagates the result
of this simulated match in order to update the victory statistics
of all the states of the tree that are part of the history of that
match. This iteration is repeated as long as there is time. The action
chosen after the search is the one that is most selected at the root
of the tree.

The basic algorithm for the exploration/exploitation dilemma is UCT.
With UCT, the most interesting action $a$ in a state $s$ is the
one that maximizes the following formula: 
\[
\frac{w_{a}}{n_{a}}+C\cdot\sqrt{\frac{\ln\sum_{a'\in\Actions s}n_{a'}}{n_{a}}}
\]
with $n_{a}$ the number of times the action $a$ was selected in
the state $s$ and $w_{a}$ the number of victories for the current
player obtained during the end-of-game simulations carried out passing
through this state. Note that the use of $\mcts{}$ is consequently
parameterized by an exploration coefficient $C$. The theoretical
value of $C$ is $\sqrt{2}$, but in practice smaller values improve
performance. It is often useful to tune the algorithm for each game.
Our implementaton of this algorithm uses transposition tables (as
for other algorithms). Figure~\ref{alg:Algorithme-MCTS} described
the $\mcts{}$ algorithm.

\paragraph{$\protect\mcts{}$ tuning}

We have performed a tuning experiment for the constant $C$ of $\mcts{}$
with a search time of 10 seconds per action by testing the following
values: $C\in\left\{ \sqrt{2},1,0.3,0.1,0.03\right\} $. Performance
for each tested value $C$ is in Table~\ref{tab:Tuning-MCTS-10s-without-child-batching}.
We have also performed a tuning for the constant $C$ of $\mcts{}$
with a search time of 1 second per action by testing the following
values: $C\in\left\{ \sqrt{2},1,0.3,0.1,0.03,0.01,0.003,0.001,0.0003,0.0001\right\} $.
Performance for each tested value $C$ is in Table~\ref{tab:Tuning-MCTS-1s-without-child-batching-part-1}
(mean) and in Table~\ref{tab:Tuning-MCTS-1s-without-child-batching-part-2}
(details). The used values for $C$ in the main comparison tables
of the paper are each time the best identified for each game ($C$
is game-dependent). We denote this by $C=*$ in the tuning tables.
Note that this tuning process, i.e. taking $C=*$, produces an upper
bound on the real performance. To know the real performance, it is
necessary to apply a cross-validation. However, we do not need it,
since we show that the Unbounded Minimax algorithms have better performances
than an upper bound of the $\mcts{}$ performance and therefore than
the real performance of $\mcts{}$. As a reminder, Child Batching
is not used since it cannot be applied (see Section~\ref{subsec:Child-Batching}).

\subsubsection{Neural value based Monte Carlo Tree Search: $\protect\mctsh{}$\label{subsec:Neural-value-based-MCTS}}

Monte Carlo Tree Search is a zero-knowledge algorithm, so it generally
does not achieve high level of play. It has been modified to use a
state evaluation function so that it can use knowledge to achieve
a better level of play \citep{ramanujan2011trade}. We denote this
algorithm $\mctsh{}$.

\paragraph{$\protect\mctsh{}$ algorithm}

This algorithm simply consists in replacing the end-of-game simulations for
evaluating states with the evaluation function $f$. More formally,
this amounts to redefining in the $\mcts{}$ algorithm of Figure~\ref{alg:Algorithme-MCTS}
the method $\rollout s$ by $f\left(s\right)$ (i.e. $\rollout s=f\left(s\right)$).
If the evaluation function is not in $[0,1]$, it is normalized to
belong to $[0,1]$ (analogously to Section~\ref{subsec:Discr=00003D0000E9tisation-de-fonctions}).

\paragraph{$\protect\mctsh{}$ tuning}

We have performed a tuning experiment for the constant $C$ of $\mctsh{}$
with a search time of 10 seconds per action. We tested the following
values: 
\[
C\in\{\sqrt{2},1,0.3,0.1,0.03,0.01,0.003,0.0003\}
\]
Performance for each tested value $C$ is in Table~\ref{tab:Tuning-MCTSh-10s-without-child-batching-part-1}
(mean) and in Table~\ref{tab:Tuning-MCTSh-10s-without-child-batching-part-2}
(details). We have also performed a tuning for the constant $C$ of
$\mctsh{}$ with a search time of 1 second per action by testing the
following values: 
\[
C\in\left\{ \sqrt{2},1,0.3,0.1,0.03,0.01,0.003,0.001,0.0003,0.0001,0.00003,0.00001,0\right\} 
\]
Performance for each tested value $C$ is in Table~\ref{tab:Tuning-MCTSh-1s-without-child-batching-part-1}
(mean) and in Table~\ref{tab:Tuning-MCTSh-1s-without-child-batching-part-2}
and Table~\ref{tab:Tuning-MCTSh-1s-without-child-batching-part-3}
(details). The used values for $C$ in the main comparison tables
of the paper are each time the best identified for each game ($C$
is game-dependent). We denote this by $C=*$ in the tuning tables.
Note that this tuning process, i.e. taking $C=*$, produces an upper
bound on the real performance. To know the real performance, it is
necessary to apply a cross-validation. However, we do not need it,
since we show that the Unbounded Minimax algorithms have better performances
than an upper bound of the $\mctsh{}$ performance and therefore than
the real performance of $\mctsh{}$. As a reminder, Child Batching
is not used since it cannot be applied (see Section~\ref{subsec:Child-Batching}).

{\small{}{}%%%%%%%%%%%%%%%% SUPPLEMENTARY TEXT %%%%%%%%%%%%%%%}{\small\par}}{\small\par}

{\small{}{}%%%%%%%%%%%%%%%% SUPPLEMENTARY FIGURES %%%%%%%%%%%%%%%}{\small\par}}{\small\par}

{\small{}{} }{\small\par}

{\small{}}
\begin{figure}
\begin{centering}
{\small{}{}\includegraphics[width=0.6\textwidth]{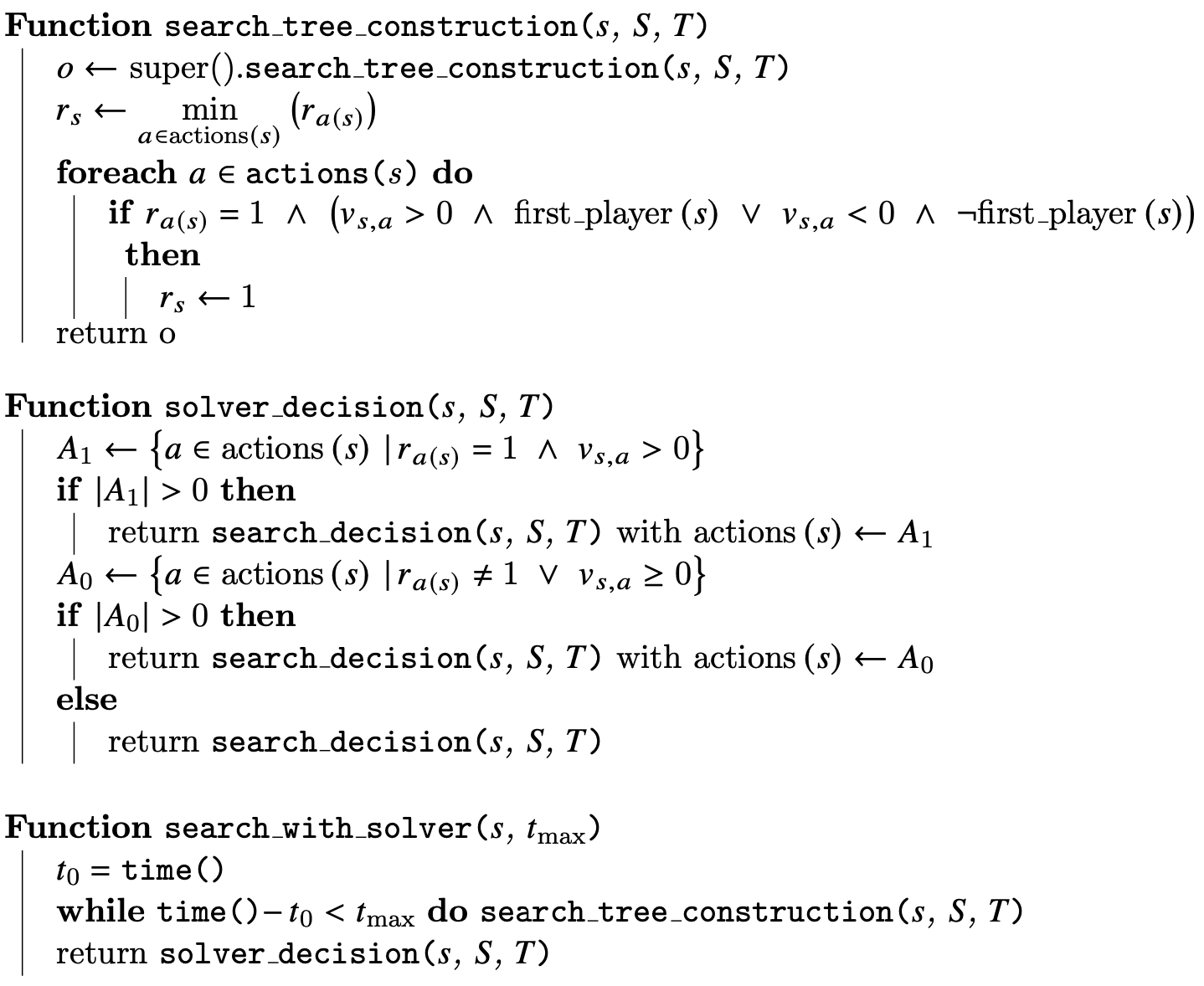}}{\small\par}
\par\end{centering}
\begin{centering}
{\small{} }{\small\par}
\par\end{centering}
{\small{}{}\caption{\textbf{Transformation used to add a basic solver into a search algorithm.}
The transformation is based on a method $\mathrm{search\_tree\_construction\left(\right)}$
that builds the search tree and a method $\mathrm{search\_decision}\left(\right)$
that decides which action to play after building the search tree ($\protect\sresol s$:
resolution value of s ($0$ by default)).\label{alg:common_solver}}
}{\small\par}

{\small{} }{\small\par}
\end{figure}
{\small\par}

{\small{}{} }
\begin{figure}
\begin{centering}
{\small{}{}\includegraphics[width=0.5\textwidth]{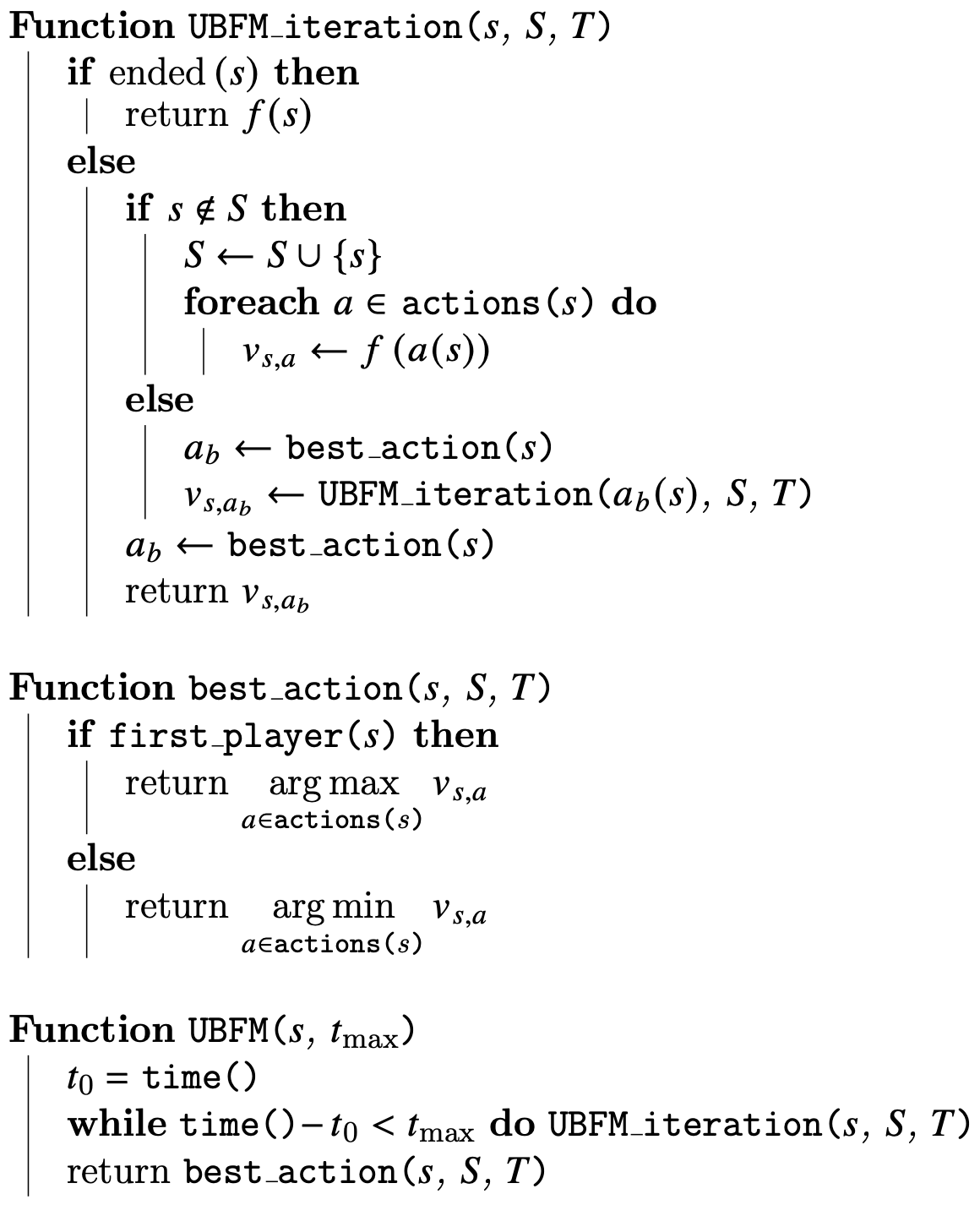}}{\small\par}
\par\end{centering}
\begin{centering}
{\small{} }{\small\par}
\par\end{centering}
{\small{}{}\caption{\textbf{Unbounded Best-First Minimax algorithm $\protect\ubfm$.}
It computes the best action to play in the generated non-uniform partial
game tree starting from the root state $s$ with $\protect\tmax$
as research time (see Table~\ref{tab:Index-of-symbols} for the definitions
of symbols ; at any time $T=\protect\liste{\protect\mv sa}{s\in S\protect\et a\in\mathrm{actions}(s)}$).\label{alg:-Unbounded-Best-First-Mininmax-ref}}
}{\small\par}

{\small{} }{\small\par}
\end{figure}
{\small\par}

{\small{}{} }
\begin{figure}
\begin{centering}
{\small{}{}\includegraphics[width=0.4\textwidth]{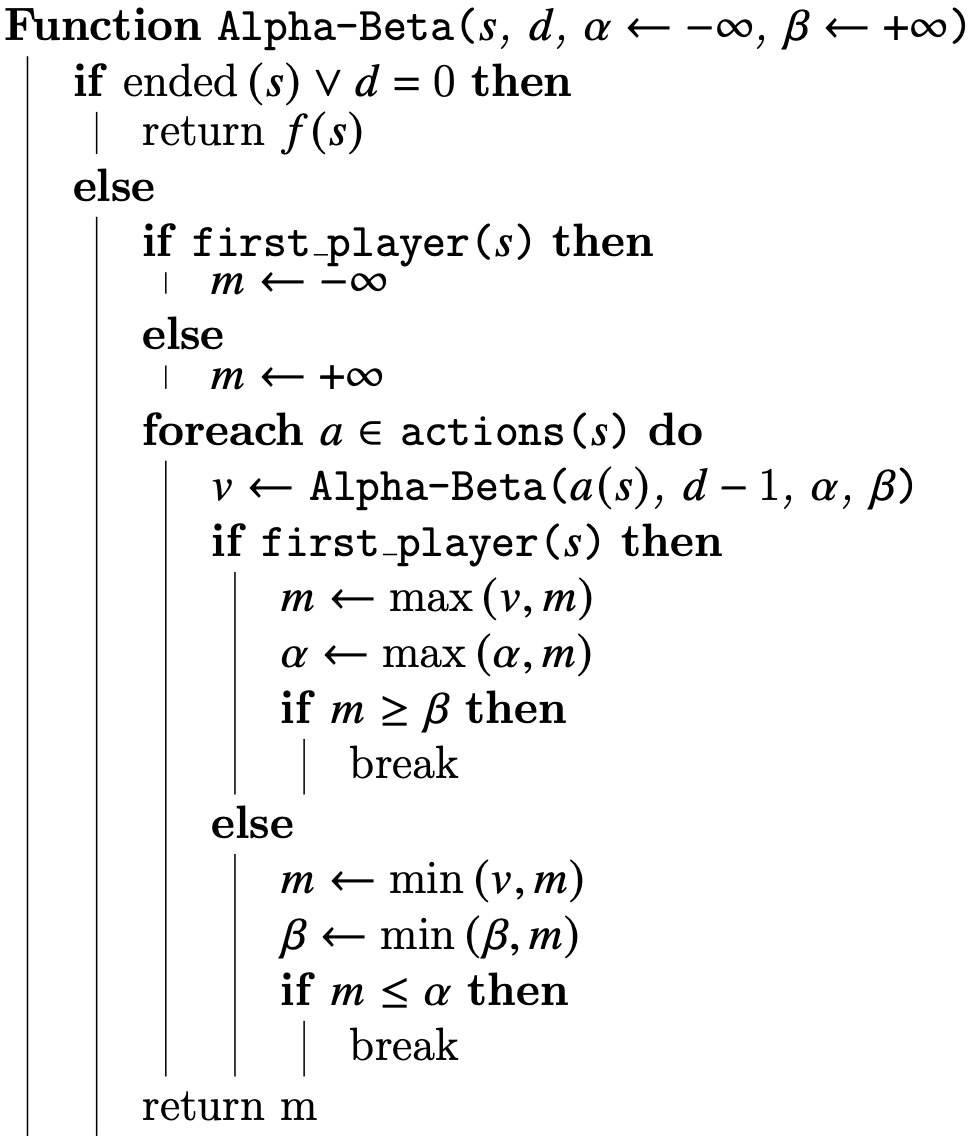}}{\small\par}
\par\end{centering}
\begin{centering}
{\small{} }{\small\par}
\par\end{centering}
{\small{}{}\caption{\textbf{Minimax algorithm with Alpha-Beta pruning. }Parameter $d$
is the search depth.\label{alg:Algorithme-AB}}
}{\small\par}

{\small{} }{\small\par}
\end{figure}
{\small\par}

{\small{}{} }
\begin{figure}
\begin{centering}
{\small{}{}\includegraphics[width=0.4\textwidth]{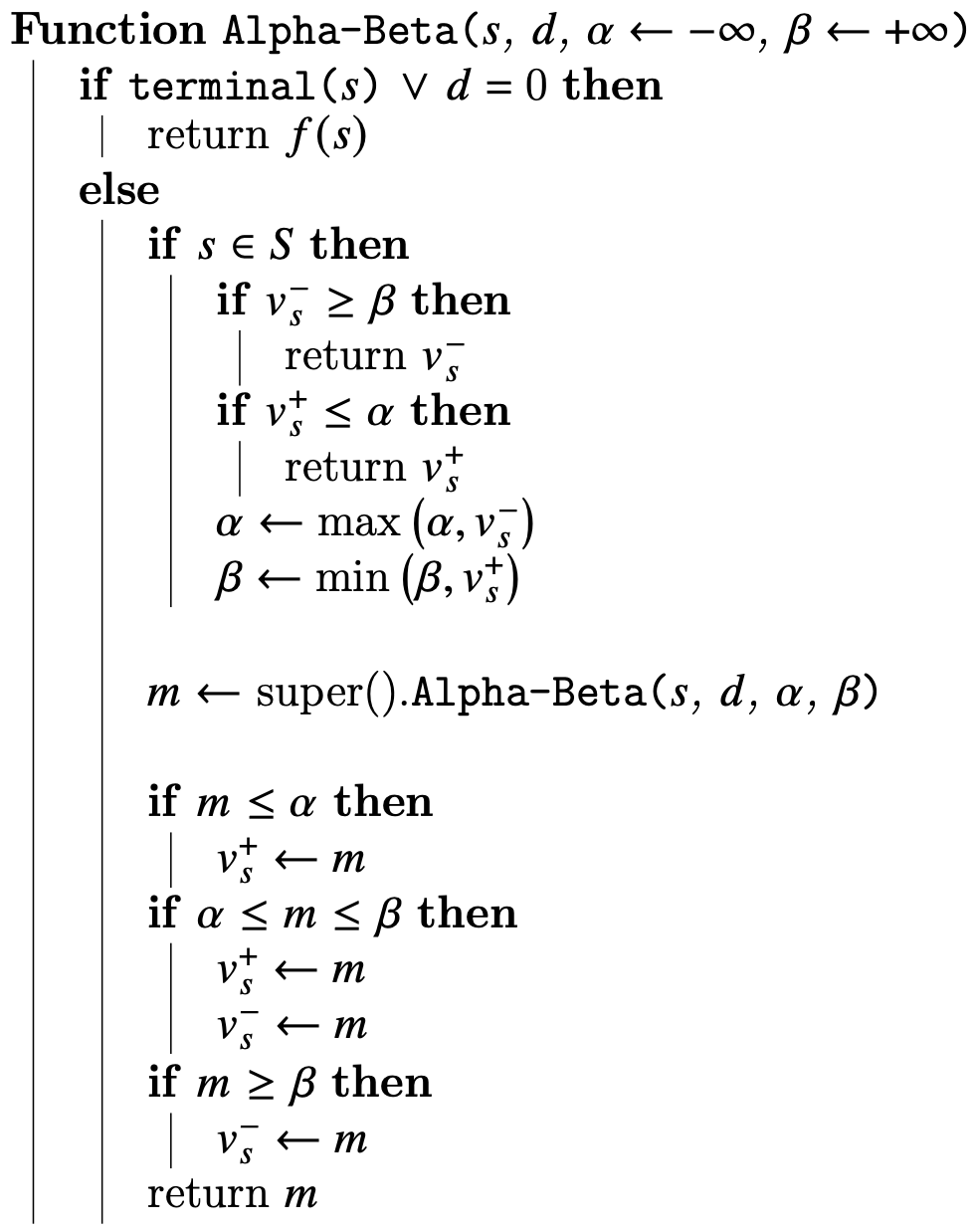}}{\small\par}
\par\end{centering}
\begin{centering}
{\small{} }{\small\par}
\par\end{centering}
{\small{}{}\caption{\textbf{Alpha-beta algorithm with transposition table.} Variable $\ensuremath{\protect\vssup s}$
is the upper bound on the value of \textbf{\emph{$s$}} and $\ensuremath{\protect\vsinf s}$
is the lower bound on the value of \textbf{\emph{$s$}}. This algorithm
extends the basic Minimax algorithm. Initially, $\ensuremath{\protect\vsinf s}=-\infty$
and $\ensuremath{\protect\vsinf s}=+\infty$.\label{alg:Algorithme-AB-transpositions}}
}{\small\par}

{\small{} }{\small\par}
\end{figure}
{\small\par}

{\small{}{} }
\begin{figure}
\begin{centering}
{\small{}{}\includegraphics[width=0.4\textwidth]{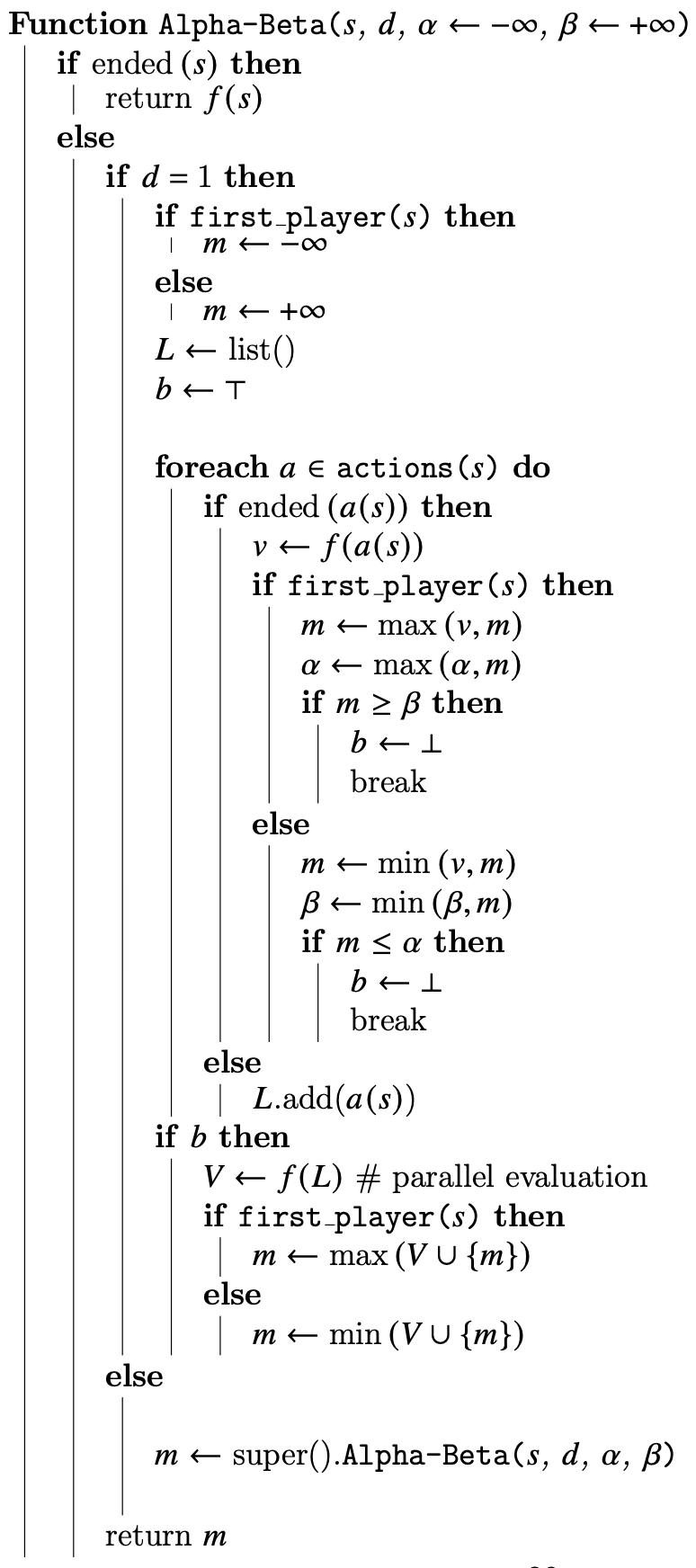}}{\small\par}
\par\end{centering}
\begin{centering}
{\small{} }{\small\par}
\par\end{centering}
{\small{}{}\caption{\textbf{Alpha-Beta algorithm with Child Batching~\label{alg:Algorithme-AB-Child-Batching}}This
algorithm extends the basic Minimax algorithm.}
}{\small\par}

{\small{} }{\small\par}
\end{figure}
{\small\par}

{\small{}{} }
\begin{figure}
\begin{centering}
{\small{}{}\includegraphics[width=0.4\textwidth]{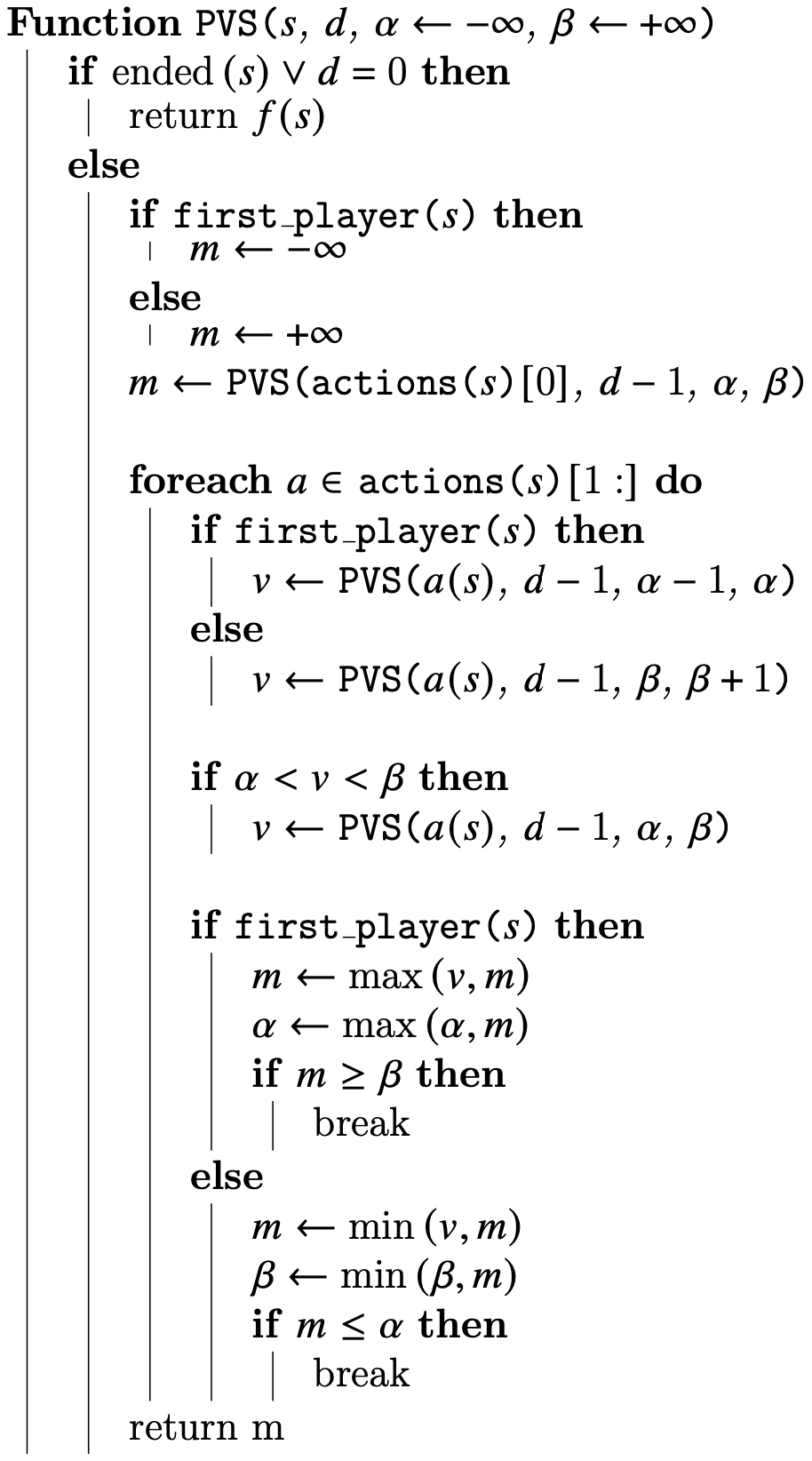}}{\small\par}
\par\end{centering}
\begin{centering}
{\small{} }{\small\par}
\par\end{centering}
{\small{}{}\caption{\textbf{PVS algorithm.} Parameter is $d$ is the search depth.\label{alg:Algorithme-PVS}}
}{\small\par}

{\small{} }{\small\par}
\end{figure}
{\small\par}

{\small{}{} }
\begin{figure}
\begin{centering}
{\small{}{}\includegraphics[width=0.35\textwidth]{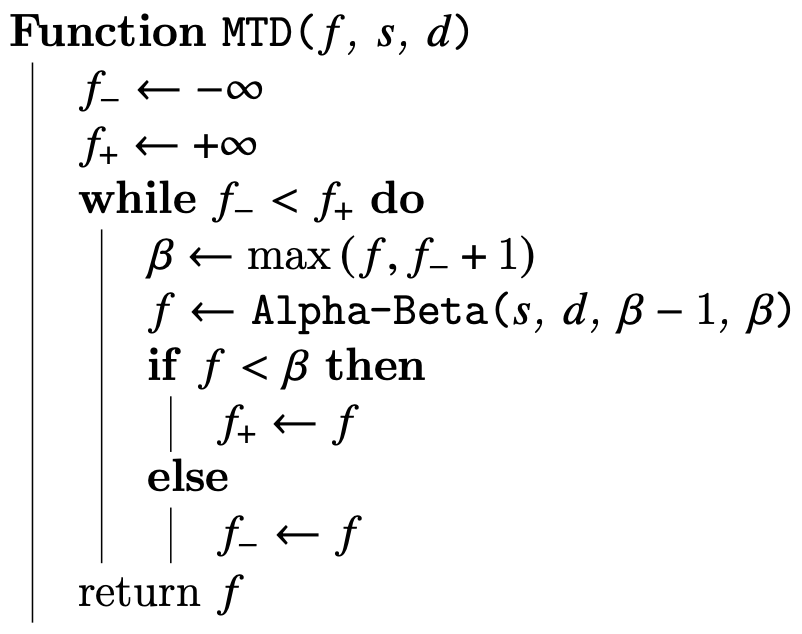}}{\small\par}
\par\end{centering}
\begin{centering}
{\small{} }{\small\par}
\par\end{centering}
{\small{}{}\caption{\textbf{MTD(f) algorithm.\label{alg:Algorithme-MTDf}}}
}{\small\par}

{\small{} }{\small\par}
\end{figure}
{\small\par}

{\small{}{} }
\begin{figure}
\begin{centering}
{\small{}{}\includegraphics[width=0.5\textwidth]{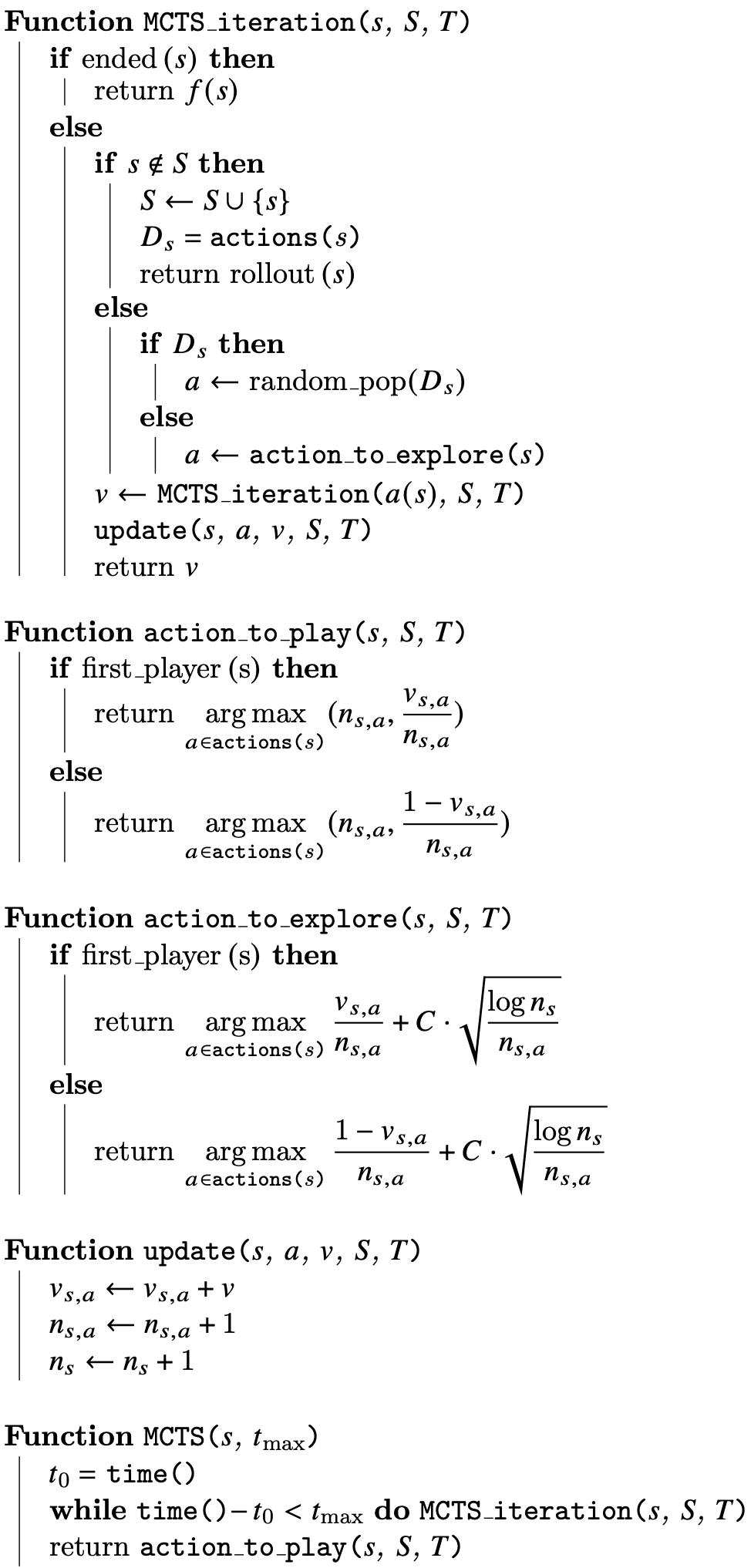}}{\small\par}
\par\end{centering}
\begin{centering}
{\small{} }{\small\par}
\par\end{centering}
{\small{}{}\caption{\textbf{$\protect\mcts{}$ algorithm with transposition table.} Initially,
$\protect\sselect s=0$ for each state $s$.\label{alg:Algorithme-MCTS}}
}{\small\par}

{\small{} }{\small\par}
\end{figure}
{\small\par}

{\small{}{} }{\small\par}

{\small{}{} }{\small\par}

{\small{}{} }
\begin{table}
{\small{}\centering{}{}\caption{\textbf{List of games used during each experiment in this article.}\label{tab:list-of-games}}
}%
\begin{tabular}{c}
\hline 
{\small{}Games}\tabularnewline
\hline 
{\small{}Amazons}\tabularnewline
{\small{}Ataxx}\tabularnewline
{\small{}Breaktrhough}\tabularnewline
{\small{}Brazilian Draughts}\tabularnewline
{\small{}Canadian Draughts}\tabularnewline
{\small{}Clobber}\tabularnewline
{\small{}Connect6}\tabularnewline
{\small{}International Draughts}\tabularnewline
{\small{}Chess}\tabularnewline
{\small{}Go $9\times9$}\tabularnewline
{\small{}Outer-OpenGomoku}\tabularnewline
{\small{}Havannah}\tabularnewline
{\small{}Hex $11\times11$}\tabularnewline
{\small{}Hex $13\times13$}\tabularnewline
{\small{}Hex $19\times19$}\tabularnewline
{\small{}Lines of Action}\tabularnewline
{\small{}Othello {}$10\times10$}\tabularnewline
{\small{}Othello $8\times8$}\tabularnewline
{\small{}Santorini}\tabularnewline
{\small{}Surakarta}\tabularnewline
{\small{}Xiangqi}\tabularnewline
\end{tabular}
\end{table}
{\small\par}

{\small{}{} }

{\small{}}
\begin{table}
\begin{centering}
{\small{}\caption{\textbf{Average number of iterations and search time performed of
MCTS-IP-M-k-IR-M-k.} Average number of iterations and search time
per action performed during a match of MCTS-IP-M-k-IR-M-k against
itself with 10 seconds per action granted. MCTS-IP-M-k-IR-M-k uses
as evaluation function the first of the corresponding game in Table~\ref{tab:hyperparameter_heuristics},
Table~\ref{tab:hyperparameter_heuristics-2}, and Table~\ref{tab:hyperparameter_heuristics-2-1},
and Table~\ref{tab:hyperparameter_heuristics-2} (with 70 days of
learning ; see Section~\ref{subsec:Fonctions-d'=00003D0000E9valuation-utilis=00003D0000E9s}
for details about evaluation functions). Nested Alpha-Beta search
uses Child Batching (see Section~\ref{subsec:Child-Batching}).\label{tab:best-hybrid-stats}}
}{\small\par}
\par\end{centering}
{\small{}\centering{}}%
\begin{tabular}{cccccccccccc}
\hline 
 & {\tiny{}{}Amazons}{\small{} } & {\tiny{}{}Arimaa}{\small{} } & {\tiny{}{}Ataxx}{\small{} } & {\tiny{}{}Breaktrhough}{\small{} } & {\tiny{}{}Brazilian}{\small{} } & {\tiny{}{}Canadian}{\small{} } & {\tiny{}{}Clobber}{\small{} } & {\tiny{}{}Connect6}{\small{} } & {\tiny{}{}International}{\small{} } & {\tiny{}{}Chess}{\small{} } & {\tiny{}{}Go 9}\tabularnewline
\hline 
{\tiny{}{}search time}{\small{} } & {\tiny{}{}14.77}{\small{} } & {\tiny{}{}19.61}{\small{} } & {\tiny{}{}20.5}{\small{} } & {\tiny{}{}17.48}{\small{} } & {\tiny{}{}11.10}{\small{} } & {\tiny{}{}18.37}{\small{} } & {\tiny{}{}14.05}{\small{} } & {\tiny{}{}12.78}{\small{} } & {\tiny{}{}12.54}{\small{} } & {\tiny{}{}14.24}{\small{} } & {\tiny{}{}17.89}\tabularnewline
{\tiny{}{}iterations}{\small{} } & {\tiny{}{}17.70}{\small{} } & {\tiny{}{}3.52}{\small{} } & {\tiny{}{}4.46}{\small{} } & {\tiny{}{}7.82}{\small{} } & {\tiny{}{}23.2}{\small{} } & {\tiny{}{}6.79}{\small{} } & {\tiny{}{}21.92}{\small{} } & {\tiny{}{}10.16}{\small{} } & {\tiny{}{}9.80}{\small{} } & {\tiny{}{}5.53}{\small{} } & {\tiny{}{}3.60}\tabularnewline
\hline 
 & {\tiny{}{}Gomoku}{\small{} } & {\tiny{}{}Havannah}{\small{} } & {\tiny{}{}Hex 11}{\small{} } & {\tiny{}{}Hex 13}{\small{} } & {\tiny{}{}Hex 19}{\small{} } & {\tiny{}{}Lines of A.}{\small{} } & {\tiny{}{}Othello 10}{\small{} } & {\tiny{}{}Othello 8}{\small{} } & {\tiny{}{}Santorini}{\small{} } & {\tiny{}{}Surakarta}{\small{} } & {\tiny{}{}Xiangqi}\tabularnewline
\hline 
{\tiny{}{}search time}{\small{} } & {\tiny{}{}13.06}{\small{} } & {\tiny{}{}18.70}{\small{} } & {\tiny{}{}13.77}{\small{} } & {\tiny{}{}13.05}{\small{} } & {\tiny{}{}23.09}{\small{} } & {\tiny{}{}11.45}{\small{} } & {\tiny{}{}14.49}{\small{} } & {\tiny{}{}11.37}{\small{} } & {\tiny{}{}10.44}{\small{} } & {\tiny{}{}10.82}{\small{} } & {\tiny{}{}19.61}\tabularnewline
{\tiny{}{}iterations}{\small{} } & {\tiny{}{}9.90}{\small{} } & {\tiny{}{}5.39}{\small{} } & {\tiny{}{}7.57}{\small{} } & {\tiny{}{}11.70}{\small{} } & {\tiny{}{}4.28}{\small{} } & {\tiny{}{}15.35}{\small{} } & {\tiny{}{}11.29}{\small{} } & {\tiny{}{}25.2}{\small{} } & {\tiny{}{}37.36}{\small{} } & {\tiny{}{}18.69}{\small{} } & {\tiny{}{}4.38}\tabularnewline
\end{tabular}{\small\par}
\end{table}
{\small\par}

{\small{}}
\begin{table}
{\small{}\centering{}\caption{\textbf{Performance of MCTS-IP-M-k-IR-M-k at Breakthrough.} Performance
(see Section~\ref{subsec:Gain-d'un-algorithme}) against benchmark
Unbounded Minimax with Child Batching (see Section~\ref{subsec:Minimax-non-born=00003D0000E9-de-r=00003D0000E9f=00003D0000E9rence})
at the Breakthrough game of the best hybridation of Minimax and MCTS
(called MCTS-IP-M-k-IR-M-k) from \citep{baier2018mcts} using its
best parameters for Breakthrough (Table 3 of \citep{baier2018mcts})
for different exploration constants $C$ of the MCTS part (minimax
part uses Child Batching). \label{tab:Performances-at-Breakthrough-best-hybrid}}
}%
\begin{tabular}{ccccc}
\hline 
{\small{}$C=1$ } & {\small{}$C=0.1$ } & {\small{}$C=0.01$ } & {\small{}$C=0.001$ } & {\small{}$C=0.0001$}\tabularnewline
\hline 
{\small{}$-100\%\pm0\%$ } & {\small{}$-100\%\pm0\%$ } & {\small{}$-100\%\pm0\%$ } & {\small{}$-100\%\pm0\%$ } & {\small{}$-100\%\pm0\%$}\tabularnewline
\end{tabular}
\end{table}
{\small\par}

{\small{}}
\begin{table}
{\small{}\caption{\textbf{Index of symbols.\label{tab:Index-of-symbols}}}
}{\small\par}

{\small{}\centering{}}{\footnotesize{}{}}{\small{}}%
\begin{tabular}{|c|c|}
\hline 
{\footnotesize{}{}Symbols}{\small{} } & {\footnotesize{}{}Definition}\tabularnewline
\hline 
\hline 
{\small{}$s$ } & {\footnotesize{}{}a game state}\tabularnewline
\hline 
{\small{}$a$ } & {\footnotesize{}{}an action of the game}\tabularnewline
\hline 
{\footnotesize{}{}$\mathrm{actions}\left(s\right)$}{\small{} } & {\footnotesize{}{}action set of $s$}\tabularnewline
\hline 
{\footnotesize{}{}$\mathrm{first\_player}\left(s\right)$}{\small{} } & {\footnotesize{}{}true if the current player of $s$ is the first
player}\tabularnewline
\hline 
{\footnotesize{}{}$\ended s$}{\small{} } & {\footnotesize{}{}true if $s$ is an end-game state}\tabularnewline
\hline 
{\footnotesize{}{}$a(s)$}{\small{} } & {\footnotesize{}{}state obtained after playing $a$ in $s$}\tabularnewline
\hline 
{\footnotesize{}{}$\mathrm{time}\left(\right)$}{\small{} } & {\footnotesize{}{}current time in seconds}\tabularnewline
\hline 
\end{tabular}{\small\par}

{\small{} }{\small\par}
\end{table}
{\small\par}

{\small{}}
\begin{table}
\begin{centering}
{\small{}\caption{\textbf{Details of the number of heuristic evaluation functions used
per game.} Used heuristics for all games per repetition set: 326 and
in total: 1956.\textbf{\label{tab:number_heuristic}}}
}{\small\par}
\par\end{centering}
{\small{}\centering{}}%
\begin{tabular}{cccccccccccc}
\hline 
\tiny{Number in} & {\tiny{}{}Amazons}{\small{} } & {\tiny{}{}Arimaa}{\small{} } & {\tiny{}{}Ataxx}{\small{} } & {\tiny{}{}Breaktrhough}{\small{} } & {\tiny{}{}Brazilian}{\small{} } & {\tiny{}{}Canadian}{\small{} } & {\tiny{}{}Clobber}{\small{} } & {\tiny{}{}Connect6}{\small{} } & {\tiny{}{}International}{\small{} } & {\tiny{}{}Chess}{\small{} } & {\tiny{}{}Go 9}\tabularnewline
\hline 
{\tiny{}{} repetition set}{\small{} } & {\tiny{}{}15}{\small{} } & {\tiny{}{}16}{\small{} } & {\tiny{}{}11}{\small{} } & {\tiny{}{}16}{\small{} } & {\tiny{}{}16}{\small{} } & {\tiny{}{}11}{\small{} } & {\tiny{}{}16}{\small{} } & {\tiny{}{}10}{\small{} } & {\tiny{}{}16}{\small{} } & {\tiny{}{}16}{\small{} } & {\tiny{}{}15}\tabularnewline
{\tiny{}{} total}{\small{} } & {\tiny{}{}90}{\small{} } & {\tiny{}{}96}{\small{} } & {\tiny{}{}66}{\small{} } & {\tiny{}{}96}{\small{} } & {\tiny{}{}96}{\small{} } & {\tiny{}{}66}{\small{} } & {\tiny{}{}96}{\small{} } & {\tiny{}{}60}{\small{} } & {\tiny{}{}96}{\small{} } & {\tiny{}{}96}{\small{} } & {\tiny{}{}90}\tabularnewline
\hline 
 & {\tiny{}{}Gomoku}{\small{} } & {\tiny{}{}Havannah}{\small{} } & {\tiny{}{}Hex 11}{\small{} } & {\tiny{}{}Hex 13}{\small{} } & {\tiny{}{}Hex 19}{\small{} } & {\tiny{}{}Lines of A.}{\small{} } & {\tiny{}{}Othello 10}{\small{} } & {\tiny{}{}Othello 8}{\small{} } & {\tiny{}{}Santorini}{\small{} } & {\tiny{}{}Surakarta}{\small{} } & {\tiny{}{}Xiangqi}\tabularnewline
\hline 
{\tiny{}{} repetition set}{\small{} } & {\tiny{}{}16}{\small{} } & {\tiny{}{}15}{\small{} } & {\tiny{}{}16}{\small{} } & {\tiny{}{}16}{\small{} } & {\tiny{}{}16}{\small{} } & {\tiny{}{}16}{\small{} } & {\tiny{}{}16}{\small{} } & {\tiny{}{}16}{\small{} } & {\tiny{}{}10}{\small{} } & {\tiny{}{}16}{\small{} } & {\tiny{}{}15}\tabularnewline
\end{tabular}{\small\par}
\end{table}
{\small\par}

{\small{}}
\begin{table}
\begin{centering}
{\small{}\caption{\textbf{Part I of Descent hyperparameters for generating the evaluation
functions.}\label{tab:hyperparameter_heuristics}}
}{\small\par}
\par\end{centering}
{\small{}\centering{}}{\tiny{}{}}{\small{}}%
\begin{tabular}{|c|c|c|c|c|c|c|c|c|}
\hline 
{\tiny{}{}game}{\small{} } & {\tiny{}{}$t$}{\small{} } & {\tiny{}{}$B$}{\small{} } & {\tiny{}{}$\delta$}{\small{} } & {\tiny{}{}$W$}{\small{} } & {\tiny{}{}$R$}{\small{} } & {\tiny{}{}$D$}{\small{} } & {\tiny{}{}R.H}{\small{} } & {\tiny{}{}uses}\tabularnewline
\hline 
\hline 
{\tiny{}{}Amazons}{\small{} } & {\tiny{}{}1}{\small{} } & {\tiny{}{}3000}{\small{} } & {\tiny{}{}2}{\small{} } & {\tiny{}{}1E7}{\small{} } & {\tiny{}{}4}{\small{} } & {\tiny{}{}2}{\small{} } & {\tiny{}{}score}{\small{} } & {\tiny{}{}1}\tabularnewline
\hline 
{\tiny{}{}Amazons}{\small{} } & {\tiny{}{}5}{\small{} } & {\tiny{}{}3000}{\small{} } & {\tiny{}{}2}{\small{} } & {\tiny{}{}5E6}{\small{} } & {\tiny{}{}4}{\small{} } & {\tiny{}{}2}{\small{} } & {\tiny{}{}score}{\small{} } & {\tiny{}{}1}\tabularnewline
\hline 
{\tiny{}{}Amazons}{\small{} } & {\tiny{}{}5}{\small{} } & {\tiny{}{}4000}{\small{} } & {\tiny{}{}3}{\small{} } & {\tiny{}{}1E6}{\small{} } & {\tiny{}{}8}{\small{} } & {\tiny{}{}2}{\small{} } & {\tiny{}{}score}{\small{} } & {\tiny{}{}3}\tabularnewline
\hline 
{\tiny{}{}Amazons}{\small{} } & {\tiny{}{}5}{\small{} } & {\tiny{}{}4000}{\small{} } & {\tiny{}{}3}{\small{} } & {\tiny{}{}1E7}{\small{} } & {\tiny{}{}4}{\small{} } & {\tiny{}{}2}{\small{} } & {\tiny{}{}score}{\small{} } & {\tiny{}{}1}\tabularnewline
\hline 
{\tiny{}{}Amazons}{\small{} } & {\tiny{}{}5}{\small{} } & {\tiny{}{}4000}{\small{} } & {\tiny{}{}3}{\small{} } & {\tiny{}{}1E7}{\small{} } & {\tiny{}{}16}{\small{} } & {\tiny{}{}2}{\small{} } & {\tiny{}{}score}{\small{} } & {\tiny{}{}5}\tabularnewline
\hline 
{\tiny{}{}Amazons}{\small{} } & {\tiny{}{}5}{\small{} } & {\tiny{}{}4000}{\small{} } & {\tiny{}{}3}{\small{} } & {\tiny{}{}2E7}{\small{} } & {\tiny{}{}4}{\small{} } & {\tiny{}{}2}{\small{} } & {\tiny{}{}score}{\small{} } & {\tiny{}{}2}\tabularnewline
\hline 
\end{tabular}{\tiny{}{}}{\small{}}%
\begin{tabular}{|c|c|c|c|c|c|c|c|c|}
\hline 
{\tiny{}{}game}{\small{} } & {\tiny{}{}$t$}{\small{} } & {\tiny{}{}$B$}{\small{} } & {\tiny{}{}$\delta$}{\small{} } & {\tiny{}{}$W$}{\small{} } & {\tiny{}{}$R$}{\small{} } & {\tiny{}{}$D$}{\small{} } & {\tiny{}{}R.H}{\small{} } & {\tiny{}{}uses}\tabularnewline
\hline 
\hline 
{\tiny{}{}Brazilian Dr.}{\small{} } & {\tiny{}{}5}{\small{} } & {\tiny{}{}3000}{\small{} } & {\tiny{}{}1}{\small{} } & {\tiny{}{}1E7}{\small{} } & {\tiny{}{}14}{\small{} } & {\tiny{}{}2}{\small{} } & {\tiny{}{}score}{\small{} } & {\tiny{}{}2}\tabularnewline
\hline 
{\tiny{}{}Brazilian Dr..}{\small{} } & {\tiny{}{}5}{\small{} } & {\tiny{}{}4000}{\small{} } & {\tiny{}{}3}{\small{} } & {\tiny{}{}4E7}{\small{} } & {\tiny{}{}8}{\small{} } & {\tiny{}{}2}{\small{} } & {\tiny{}{}score + m.}{\small{} } & {\tiny{}{}3}\tabularnewline
\hline 
{\tiny{}{}Brazilian Dr.}{\small{} } & {\tiny{}{}5}{\small{} } & {\tiny{}{}4000}{\small{} } & {\tiny{}{}3}{\small{} } & {\tiny{}{}4E7}{\small{} } & {\tiny{}{}16}{\small{} } & {\tiny{}{}2}{\small{} } & {\tiny{}{}score + m.}{\small{} } & {\tiny{}{}9}\tabularnewline
\hline 
{\tiny{}{}Breaktrhough}{\small{} } & {\tiny{}{}5}{\small{} } & {\tiny{}{}3000}{\small{} } & {\tiny{}{}1}{\small{} } & {\tiny{}{}1E7}{\small{} } & {\tiny{}{}14}{\small{} } & {\tiny{}{}2}{\small{} } & {\tiny{}{}add. depth}{\small{} } & {\tiny{}{}1}\tabularnewline
\hline 
{\tiny{}{}Breaktrhough}{\small{} } & {\tiny{}{}5}{\small{} } & {\tiny{}{}3000}{\small{} } & {\tiny{}{}2}{\small{} } & {\tiny{}{}1E7}{\small{} } & {\tiny{}{}4}{\small{} } & {\tiny{}{}2}{\small{} } & {\tiny{}{}add. depth}{\small{} } & {\tiny{}{}2}\tabularnewline
\hline 
{\tiny{}{}Breaktrhough}{\small{} } & {\tiny{}{}5}{\small{} } & {\tiny{}{}3000}{\small{} } & {\tiny{}{}2}{\small{} } & {\tiny{}{}1E7}{\small{} } & {\tiny{}{}8}{\small{} } & {\tiny{}{}2}{\small{} } & {\tiny{}{}add. depth}{\small{} } & {\tiny{}{}3}\tabularnewline
\hline 
\end{tabular}{\small\par}

{\small{} }{\small\par}
\end{table}
{\small\par}

{\small{}}
\begin{table}
\begin{centering}
{\small{}\caption{\textbf{Part II of Descent hyperparameters for generating the evaluation
functions.}\label{tab:hyperparameter_heuristics-1}}
}{\small\par}
\par\end{centering}
{\small{}\centering{}}{\tiny{}{}}{\small{}}%
\begin{tabular}{|c|c|c|c|c|c|c|c|c|}
\hline 
{\tiny{}{}game}{\small{} } & {\tiny{}{}$t$}{\small{} } & {\tiny{}{}$B$}{\small{} } & {\tiny{}{}$\delta$}{\small{} } & {\tiny{}{}$W$}{\small{} } & {\tiny{}{}$R$}{\small{} } & {\tiny{}{}$D$}{\small{} } & {\tiny{}{}R.H}{\small{} } & {\tiny{}{}uses}\tabularnewline
\hline 
\hline 
{\tiny{}{}Clobber}{\small{} } & {\tiny{}{}5}{\small{} } & {\tiny{}{}4000}{\small{} } & {\tiny{}{}3}{\small{} } & {\tiny{}{}2E7}{\small{} } & {\tiny{}{}16}{\small{} } & {\tiny{}{}2}{\small{} } & {\tiny{}{}mult. depth}{\small{} } & {\tiny{}{}3}\tabularnewline
\hline 
{\tiny{}{}Clobber}{\small{} } & {\tiny{}{}5}{\small{} } & {\tiny{}{}4000}{\small{} } & {\tiny{}{}3}{\small{} } & {\tiny{}{}3E7}{\small{} } & {\tiny{}{}4}{\small{} } & {\tiny{}{}2}{\small{} } & {\tiny{}{}mult. depth}{\small{} } & {\tiny{}{}1}\tabularnewline
\hline 
{\tiny{}{}Clobber}{\small{} } & {\tiny{}{}5}{\small{} } & {\tiny{}{}4000}{\small{} } & {\tiny{}{}3}{\small{} } & {\tiny{}{}9E6}{\small{} } & {\tiny{}{}16}{\small{} } & {\tiny{}{}2}{\small{} } & {\tiny{}{}mult. depth}{\small{} } & {\tiny{}{}2}\tabularnewline
\hline 
{\tiny{}{}Clobber}{\small{} } & {\tiny{}{}5}{\small{} } & {\tiny{}{}4000}{\small{} } & {\tiny{}{}5}{\small{} } & {\tiny{}{}1E6}{\small{} } & {\tiny{}{}8}{\small{} } & {\tiny{}{}2}{\small{} } & {\tiny{}{}mult. depth}{\small{} } & {\tiny{}{}7}\tabularnewline
\hline 
{\tiny{}{}Clobber}{\small{} } & {\tiny{}{}5}{\small{} } & {\tiny{}{}4000}{\small{} } & {\tiny{}{}5}{\small{} } & {\tiny{}{}3E7}{\small{} } & {\tiny{}{}4}{\small{} } & {\tiny{}{}2}{\small{} } & {\tiny{}{}mult. depth}{\small{} } & {\tiny{}{}1}\tabularnewline
\hline 
{\tiny{}{}Connect6}{\small{} } & {\tiny{}{}1}{\small{} } & {\tiny{}{}4000}{\small{} } & {\tiny{}{}3}{\small{} } & {\tiny{}{}3E6}{\small{} } & {\tiny{}{}16}{\small{} } & {\tiny{}{}2}{\small{} } & {\tiny{}{}mult. depth}{\small{} } & {\tiny{}{}1}\tabularnewline
\hline 
\end{tabular}{\tiny{}{}}{\small{}}%
\begin{tabular}{|c|c|c|c|c|c|c|c|c|}
\hline 
{\tiny{}{}game}{\small{} } & {\tiny{}{}$t$}{\small{} } & {\tiny{}{}$B$}{\small{} } & {\tiny{}{}$\delta$}{\small{} } & {\tiny{}{}$W$}{\small{} } & {\tiny{}{}$R$}{\small{} } & {\tiny{}{}$D$}{\small{} } & {\tiny{}{}R.H}{\small{} } & {\tiny{}{}uses}\tabularnewline
\hline 
\hline 
{\tiny{}{}Havannah}{\small{} } & {\tiny{}{}5}{\small{} } & {\tiny{}{}4000}{\small{} } & {\tiny{}{}2}{\small{} } & {\tiny{}{}4E6}{\small{} } & {\tiny{}{}16}{\small{} } & {\tiny{}{}2}{\small{} } & {\tiny{}{}mult. depth}{\small{} } & {\tiny{}{}4}\tabularnewline
\hline 
{\tiny{}{}Havannah}{\small{} } & {\tiny{}{}5}{\small{} } & {\tiny{}{}4000}{\small{} } & {\tiny{}{}2}{\small{} } & {\tiny{}{}5E5}{\small{} } & {\tiny{}{}8}{\small{} } & {\tiny{}{}2}{\small{} } & {\tiny{}{}mult. depth}{\small{} } & {\tiny{}{}1}\tabularnewline
\hline 
{\tiny{}{}Havannah}{\small{} } & {\tiny{}{}5}{\small{} } & {\tiny{}{}4000}{\small{} } & {\tiny{}{}2}{\small{} } & {\tiny{}{}6E5}{\small{} } & {\tiny{}{}8}{\small{} } & {\tiny{}{}2}{\small{} } & {\tiny{}{}mult. depth}{\small{} } & {\tiny{}{}1}\tabularnewline
\hline 
{\tiny{}{}Havannah}{\small{} } & {\tiny{}{}5}{\small{} } & {\tiny{}{}4000}{\small{} } & {\tiny{}{}2}{\small{} } & {\tiny{}{}6E6}{\small{} } & {\tiny{}{}16}{\small{} } & {\tiny{}{}2}{\small{} } & {\tiny{}{}mult. depth}{\small{} } & {\tiny{}{}2}\tabularnewline
\hline 
{\tiny{}{}Havannah}{\small{} } & {\tiny{}{}5}{\small{} } & {\tiny{}{}4000}{\small{} } & {\tiny{}{}2}{\small{} } & {\tiny{}{}7E5}{\small{} } & {\tiny{}{}8}{\small{} } & {\tiny{}{}2}{\small{} } & {\tiny{}{}mult. depth}{\small{} } & {\tiny{}{}1}\tabularnewline
\hline 
{\tiny{}{}Havannah}{\small{} } & {\tiny{}{}5}{\small{} } & {\tiny{}{}4000}{\small{} } & {\tiny{}{}2}{\small{} } & {\tiny{}{}8E6}{\small{} } & {\tiny{}{}16}{\small{} } & {\tiny{}{}2}{\small{} } & {\tiny{}{}mult. depth}{\small{} } & {\tiny{}{}3}\tabularnewline
\hline 
\end{tabular}{\small\par}

{\small{} }{\small\par}
\end{table}
{\small\par}

{\small{}}
\begin{table}
\begin{centering}
{\small{}\caption{\textbf{Part III of Descent hyperparameters for generating the evaluation
functions.}\label{tab:hyperparameter_heuristics-2}}
}{\small\par}
\par\end{centering}
{\small{}\centering{}}{\tiny{}{}}{\small{}}%
\begin{tabular}{|c|c|c|c|c|c|c|c|c|}
\hline 
{\tiny{}{}game}{\small{} } & {\tiny{}{}$t$}{\small{} } & {\tiny{}{}$B$}{\small{} } & {\tiny{}{}$\delta$}{\small{} } & {\tiny{}{}$W$}{\small{} } & {\tiny{}{}$R$}{\small{} } & {\tiny{}{}$D$}{\small{} } & {\tiny{}{}R.H}{\small{} } & {\tiny{}{}uses}\tabularnewline
\hline 
\hline 
{\tiny{}{}Hex 19x19}{\small{} } & {\tiny{}{}1}{\small{} } & {\tiny{}{}3000}{\small{} } & {\tiny{}{}1}{\small{} } & {\tiny{}{}3E6}{\small{} } & {\tiny{}{}16}{\small{} } & {\tiny{}{}2}{\small{} } & {\tiny{}{}add. depth}{\small{} } & {\tiny{}{}2}\tabularnewline
\hline 
{\tiny{}{}Hex 19x19}{\small{} } & {\tiny{}{}1}{\small{} } & {\tiny{}{}3000}{\small{} } & {\tiny{}{}1}{\small{} } & {\tiny{}{}4E6}{\small{} } & {\tiny{}{}4}{\small{} } & {\tiny{}{}2}{\small{} } & {\tiny{}{}add. depth}{\small{} } & {\tiny{}{}1}\tabularnewline
\hline 
{\tiny{}{}Hex 19x19}{\small{} } & {\tiny{}{}1}{\small{} } & {\tiny{}{}3000}{\small{} } & {\tiny{}{}1}{\small{} } & {\tiny{}{}4E6}{\small{} } & {\tiny{}{}14}{\small{} } & {\tiny{}{}2}{\small{} } & {\tiny{}{}add. depth}{\small{} } & {\tiny{}{}2}\tabularnewline
\hline 
{\tiny{}{}Hex 19x19}{\small{} } & {\tiny{}{}1}{\small{} } & {\tiny{}{}3000}{\small{} } & {\tiny{}{}1}{\small{} } & {\tiny{}{}4E6}{\small{} } & {\tiny{}{}16}{\small{} } & {\tiny{}{}2}{\small{} } & {\tiny{}{}add. depth}{\small{} } & {\tiny{}{}1}\tabularnewline
\hline 
{\tiny{}{}Hex 19x19}{\small{} } & {\tiny{}{}1}{\small{} } & {\tiny{}{}3000}{\small{} } & {\tiny{}{}1}{\small{} } & {\tiny{}{}7E6}{\small{} } & {\tiny{}{}4}{\small{} } & {\tiny{}{}2}{\small{} } & {\tiny{}{}add. depth}{\small{} } & {\tiny{}{}1}\tabularnewline
\hline 
{\tiny{}{}Hex 19x19}{\small{} } & {\tiny{}{}1}{\small{} } & {\tiny{}{}3000}{\small{} } & {\tiny{}{}1}{\small{} } & {\tiny{}{}7E6}{\small{} } & {\tiny{}{}14}{\small{} } & {\tiny{}{}2}{\small{} } & {\tiny{}{}add. depth}{\small{} } & {\tiny{}{}2}\tabularnewline
\hline 
\end{tabular}{\tiny{}{}}{\small{}}%
\begin{tabular}{|c|c|c|c|c|c|c|c|c|}
\hline 
{\tiny{}{}game}{\small{} } & {\tiny{}{}$t$}{\small{} } & {\tiny{}{}$B$}{\small{} } & {\tiny{}{}$\delta$}{\small{} } & {\tiny{}{}$W$}{\small{} } & {\tiny{}{}$R$}{\small{} } & {\tiny{}{}$D$}{\small{} } & {\tiny{}{}R.H}{\small{} } & {\tiny{}{}uses}\tabularnewline
\hline 
\hline 
{\tiny{}{}Oth. 10x10}{\small{} } & {\tiny{}{}5}{\small{} } & {\tiny{}{}4000}{\small{} } & {\tiny{}{}3}{\small{} } & {\tiny{}{}1E7}{\small{} } & {\tiny{}{}16}{\small{} } & {\tiny{}{}2}{\small{} } & {\tiny{}{}score}{\small{} } & {\tiny{}{}1}\tabularnewline
\hline 
{\tiny{}{}Oth. 10x10}{\small{} } & {\tiny{}{}5}{\small{} } & {\tiny{}{}4000}{\small{} } & {\tiny{}{}3}{\small{} } & {\tiny{}{}3E7}{\small{} } & {\tiny{}{}8}{\small{} } & {\tiny{}{}2}{\small{} } & {\tiny{}{}score}{\small{} } & {\tiny{}{}1}\tabularnewline
\hline 
{\tiny{}{}Oth. 10x10}{\small{} } & {\tiny{}{}5}{\small{} } & {\tiny{}{}4000}{\small{} } & {\tiny{}{}3}{\small{} } & {\tiny{}{}3E7}{\small{} } & {\tiny{}{}8}{\small{} } & {\tiny{}{}2}{\small{} } & {\tiny{}{}score + m.}{\small{} } & {\tiny{}{}5}\tabularnewline
\hline 
{\tiny{}{}Oth. 10x10}{\small{} } & {\tiny{}{}5}{\small{} } & {\tiny{}{}4000}{\small{} } & {\tiny{}{}3}{\small{} } & {\tiny{}{}3E7}{\small{} } & {\tiny{}{}16}{\small{} } & {\tiny{}{}2}{\small{} } & {\tiny{}{}score + m.}{\small{} } & {\tiny{}{}1}\tabularnewline
\hline 
{\tiny{}{}Oth. 10x10}{\small{} } & {\tiny{}{}5}{\small{} } & {\tiny{}{}4000}{\small{} } & {\tiny{}{}3}{\small{} } & {\tiny{}{}4E7}{\small{} } & {\tiny{}{}3}{\small{} } & {\tiny{}{}3}{\small{} } & {\tiny{}{}score}{\small{} } & {\tiny{}{}1}\tabularnewline
\hline 
{\tiny{}{}Oth. 10x10}{\small{} } & {\tiny{}{}5}{\small{} } & {\tiny{}{}4000}{\small{} } & {\tiny{}{}3}{\small{} } & {\tiny{}{}4E7}{\small{} } & {\tiny{}{}4}{\small{} } & {\tiny{}{}2}{\small{} } & {\tiny{}{}score}{\small{} } & {\tiny{}{}1}\tabularnewline
\hline 
\end{tabular}{\small\par}

{\small{} }{\small\par}
\end{table}
{\small\par}

{\small{}}
\begin{table}
\begin{centering}
{\small{}\caption{\textbf{Part IV of Descent hyperparameters for generating the evaluation
functions.}\label{tab:hyperparameter_heuristics-2-1}}
}{\small\par}
\par\end{centering}
{\small{}\centering{}}{\tiny{}{}}{\small{}}%
\begin{tabular}{|c|c|c|c|c|c|c|c|c|}
\hline 
{\tiny{}{}game}{\small{} } & {\tiny{}{}$t$}{\small{} } & {\tiny{}{}$B$}{\small{} } & {\tiny{}{}$\delta$}{\small{} } & {\tiny{}{}$W$}{\small{} } & {\tiny{}{}$R$}{\small{} } & {\tiny{}{}$D$}{\small{} } & {\tiny{}{}R.H}{\small{} } & {\tiny{}{}uses}\tabularnewline
\hline 
\hline 
{\tiny{}{}Santorini}{\small{} } & {\tiny{}{}5}{\small{} } & {\tiny{}{}4000}{\small{} } & {\tiny{}{}5}{\small{} } & {\tiny{}{}4E7}{\small{} } & {\tiny{}{}8}{\small{} } & {\tiny{}{}2}{\small{} } & {\tiny{}{}add. depth}{\small{} } & {\tiny{}{}1}\tabularnewline
\hline 
{\tiny{}{}Santorini}{\small{} } & {\tiny{}{}5}{\small{} } & {\tiny{}{}4000}{\small{} } & {\tiny{}{}5}{\small{} } & {\tiny{}{}5E6}{\small{} } & {\tiny{}{}8}{\small{} } & {\tiny{}{}2}{\small{} } & {\tiny{}{}add. depth}{\small{} } & {\tiny{}{}1}\tabularnewline
\hline 
{\tiny{}{}Surakarta}{\small{} } & {\tiny{}{}5}{\small{} } & {\tiny{}{}3000}{\small{} } & {\tiny{}{}3}{\small{} } & {\tiny{}{}4E7}{\small{} } & {\tiny{}{}4}{\small{} } & {\tiny{}{}2}{\small{} } & {\tiny{}{}score}{\small{} } & {\tiny{}{}1}\tabularnewline
\hline 
{\tiny{}{}Surakarta}{\small{} } & {\tiny{}{}5}{\small{} } & {\tiny{}{}3000}{\small{} } & {\tiny{}{}3}{\small{} } & {\tiny{}{}5E6}{\small{} } & {\tiny{}{}4}{\small{} } & {\tiny{}{}2}{\small{} } & {\tiny{}{}score}{\small{} } & {\tiny{}{}1}\tabularnewline
\hline 
{\tiny{}{}Surakarta}{\small{} } & {\tiny{}{}5}{\small{} } & {\tiny{}{}4000}{\small{} } & {\tiny{}{}1}{\small{} } & {\tiny{}{}5E6}{\small{} } & {\tiny{}{}8}{\small{} } & {\tiny{}{}2}{\small{} } & {\tiny{}{}score}{\small{} } & {\tiny{}{}1}\tabularnewline
\hline 
{\tiny{}{}Surakarta}{\small{} } & {\tiny{}{}5}{\small{} } & {\tiny{}{}4000}{\small{} } & {\tiny{}{}3}{\small{} } & {\tiny{}{}1E7}{\small{} } & {\tiny{}{}8}{\small{} } & {\tiny{}{}2}{\small{} } & {\tiny{}{}score}{\small{} } & {\tiny{}{}2}\tabularnewline
\hline 
\end{tabular}{\small\par}

{\small{} }{\small\par}
\end{table}
{\small\par}

{\small{}}
\begin{table}
\begin{centering}
{\small{}\caption{\textbf{Details of the number of matches played per game and per algorithm.
}Number of matches per algorithm for all games: 9852 per repetition
and 59112 in total\textbf{.\label{tab:number_of_matches}}}
}{\small\par}
\par\end{centering}
{\small{}\centering{}}%
\begin{tabular}{cccccccccccc}
\hline 
 & {\tiny{}{}Amazons}{\small{} } & {\tiny{}{}Arimaa}{\small{} } & {\tiny{}{}Ataxx}{\small{} } & {\tiny{}{}Breaktrhough}{\small{} } & {\tiny{}{}Brazilian}{\small{} } & {\tiny{}{}Canadian}{\small{} } & {\tiny{}{}Clobber}{\small{} } & {\tiny{}{}Connect6}{\small{} } & {\tiny{}{}International}{\small{} } & {\tiny{}{}Chess}{\small{} } & {\tiny{}{}Go 9}\tabularnewline
\hline 
{\tiny{}{}per repetition}{\small{} } & {\tiny{}{}450}{\small{} } & {\tiny{}{}512}{\small{} } & {\tiny{}{}242}{\small{} } & {\tiny{}{}512}{\small{} } & {\tiny{}{}512}{\small{} } & {\tiny{}{}242}{\small{} } & {\tiny{}{}512}{\small{} } & {\tiny{}{}200}{\small{} } & {\tiny{}{}512}{\small{} } & {\tiny{}{}512}{\small{} } & {\tiny{}{}450}\tabularnewline
{\tiny{}{}in total}{\small{} } & {\tiny{}{}2700}{\small{} } & {\tiny{}{}3072}{\small{} } & {\tiny{}{}1452}{\small{} } & {\tiny{}{}3072}{\small{} } & {\tiny{}{}3072}{\small{} } & {\tiny{}{}1452}{\small{} } & {\tiny{}{}3072}{\small{} } & {\tiny{}{}1200}{\small{} } & {\tiny{}{}3072}{\small{} } & {\tiny{}{}3072}{\small{} } & {\tiny{}{}2700}\tabularnewline
\hline 
 & {\tiny{}{}Gomoku}{\small{} } & {\tiny{}{}Havannah}{\small{} } & {\tiny{}{}Hex 11}{\small{} } & {\tiny{}{}Hex 13}{\small{} } & {\tiny{}{}Hex 19}{\small{} } & {\tiny{}{}Lines of A.}{\small{} } & {\tiny{}{}Othello 10}{\small{} } & {\tiny{}{}Othello 8}{\small{} } & {\tiny{}{}Santorini}{\small{} } & {\tiny{}{}Surakarta}{\small{} } & {\tiny{}{}Xiangqi}\tabularnewline
\hline 
{\tiny{}{}per repetition}{\small{} } & {\tiny{}{}512}{\small{} } & {\tiny{}{}450}{\small{} } & {\tiny{}{}512}{\small{} } & {\tiny{}{}512}{\small{} } & {\tiny{}{}512}{\small{} } & {\tiny{}{}512}{\small{} } & {\tiny{}{}512}{\small{} } & {\tiny{}{}512}{\small{} } & {\tiny{}{}200}{\small{} } & {\tiny{}{}512}{\small{} } & {\tiny{}{}450}\tabularnewline
\end{tabular}{\small\par}
\end{table}
{\small\par}

{\small{}{} }
\begin{table}
{\small{}\centering{}{}\caption{\textbf{Tuning of $\protect\pvs{}$ with Child Batching and $10$
seconds per action.}\label{tab:Tuning-PVS-10s-with-child-batching}}
}%
\begin{tabular}{cccccccccccc}
\hline 
 &  &  & {\small{}$\delta$ } & {\tiny{}{}mean}{\small{} } & \multicolumn{2}{c}{{\tiny{}{}lower bound}} & \multicolumn{2}{c}{{\tiny{}{}upper bound}} &  &  & \tabularnewline
\hline 
 &  &  & {\small{}$10$ } & {\small{}-50.02 } & \multicolumn{2}{c}{{\small{}-50.02}} & \multicolumn{2}{c}{{\small{}-50.53}} &  &  & \tabularnewline
 &  &  & {\small{}$100$ } & {\small{}-44.31 } & \multicolumn{2}{c}{{\small{}-44.31}} & \multicolumn{2}{c}{{\small{}-44.82}} &  &  & \tabularnewline
 &  &  & {\small{}$300$ } & {\small{}-43.09 } & \multicolumn{2}{c}{{\small{}-43.09}} & \multicolumn{2}{c}{{\small{}-43.62}} &  &  & \tabularnewline
 &  &  & {\small{}$1000$ } & {\small{}-42.72 } & \multicolumn{2}{c}{{\small{}-42.72}} & \multicolumn{2}{c}{{\small{}-43.24}} &  &  & \tabularnewline
 &  &  & {\small{}$3000$ } & {\small{}-42.3 } & \multicolumn{2}{c}{{\small{}-42.3}} & \multicolumn{2}{c}{{\small{}-42.82}} &  &  & \tabularnewline
 &  &  & {\small{}$10^{4}$ } & {\small{}-42.38 } & \multicolumn{2}{c}{{\small{}-42.38}} & \multicolumn{2}{c}{{\small{}-42.9}} &  &  & \tabularnewline
\hline 
{\tiny{}{}$\delta$}{\small{} } & {\tiny{}{}Amazons}{\small{} } & {\tiny{}{}Arimaa}{\small{} } & {\tiny{}{}Ataxx}{\small{} } & {\tiny{}{}Breaktrhough}{\small{} } & {\tiny{}{}Brazilian}{\small{} } & {\tiny{}{}Canadian}{\small{} } & {\tiny{}{}Clobber}{\small{} } & {\tiny{}{}Connect6}{\small{} } & {\tiny{}{}International}{\small{} } & {\tiny{}{}Chess}{\small{} } & {\tiny{}{}Go 9}\tabularnewline
\end{tabular}
\end{table}
{\small\par}

{\small{}{} }
\begin{table}
{\small{}\centering{}{}\caption{\textbf{Tuning of $\protect\mtdf{}$ with Child Batching and $10$
seconds per action.}\label{tab:Tuning-MTDf-10s-with-child-batching}}
}%
\begin{tabular}{cccccccccccc}
\hline 
 &  & {\small{}resol. } & {\small{}$\delta$ } & {\tiny{}{}mean}{\small{} } & \multicolumn{2}{c}{{\tiny{}{}lower bound}} & \multicolumn{2}{c}{{\tiny{}{}upper bound}} &  &  & \tabularnewline
\hline 
 &  & {\small{}no } & {\small{}$10$ } & {\small{}-42.67 } & \multicolumn{2}{c}{{\small{}-43.2}} & \multicolumn{2}{c}{{\small{}-42.15}} &  &  & \tabularnewline
 &  & {\small{}no } & {\small{}30 } & {\small{}-41.83 } & \multicolumn{2}{c}{{\small{}-42.37}} & \multicolumn{2}{c}{{\small{}-41.3}} &  &  & \tabularnewline
 &  & {\small{}no } & {\small{}$100$ } & {\small{}-41.11 } & \multicolumn{2}{c}{{\small{}-41.64}} & \multicolumn{2}{c}{{\small{}-40.58}} &  &  & \tabularnewline
 &  & {\small{}no } & {\small{}$300$ } & {\small{}-41.47 } & \multicolumn{2}{c}{{\small{}-42}} & \multicolumn{2}{c}{{\small{}-40.95}} &  &  & \tabularnewline
 &  & {\small{}no } & {\small{}$1000$ } & {\small{}-41.66 } & \multicolumn{2}{c}{{\small{}-42.18}} & \multicolumn{2}{c}{{\small{}-41.13}} &  &  & \tabularnewline
 &  & {\small{}no } & {\small{}$3000$ } & {\small{}-42.03 } & \multicolumn{2}{c}{{\small{}-42.56}} & \multicolumn{2}{c}{{\small{}-41.5}} &  &  & \tabularnewline
 &  & {\small{}yes } & {\small{}100 } & {\small{}-39.73 } & \multicolumn{2}{c}{{\small{}-40.28}} & \multicolumn{2}{c}{{\small{}-39.21}} &  &  & \tabularnewline
\end{tabular}
\end{table}
{\small\par}

{\small{}{} }
\begin{table}
{\small{}\centering{}{}\caption{\textbf{Tuning of $k$-best $\protect\alphabeta$ with Child Batching
and $10$ seconds per action (mean).}\label{tab:Tuning-k-best-10s-with-child-batching-part-1}}
}%
\begin{tabular}{ccc||cc||c}
\hline 
{\small{}$k$ } & {\tiny{}{}mean}{\small{} } & \multicolumn{2}{c}{{\tiny{}{}lower bound}} & \multicolumn{2}{c}{{\tiny{}{}upper bound}}\tabularnewline
\hline 
{\small{}$\infty$ } & {\small{}-41.79 } & \multicolumn{2}{c}{{\small{}-42.3}} & \multicolumn{2}{c}{{\small{}-41.25}}\tabularnewline
{\small{}30 } & {\small{}-39.28 } & \multicolumn{2}{c}{{\small{}-39.82}} & \multicolumn{2}{c}{{\small{}-38.75}}\tabularnewline
{\small{}15 } & {\small{}-33.92 } & \multicolumn{2}{c}{{\small{}-34.46}} & \multicolumn{2}{c}{{\small{}-33.36}}\tabularnewline
{\small{}10 } & {\small{}-27.07 } & \multicolumn{2}{c}{{\small{}-27.62}} & \multicolumn{2}{c}{{\small{}-26.49}}\tabularnewline
{\small{}7 } & {\small{}-20.89 } & \multicolumn{2}{c}{{\small{}-21.48}} & \multicolumn{2}{c}{{\small{}-20.3}}\tabularnewline
{\small{}6 } & {\small{}-19.01 } & \multicolumn{2}{c}{{\small{}-19.6}} & \multicolumn{2}{c}{{\small{}-18.43}}\tabularnewline
{\small{}5 } & {\small{}-16.97 } & \multicolumn{2}{c}{{\small{}-17.56}} & \multicolumn{2}{c}{{\small{}-16.38}}\tabularnewline
{\small{}4 } & {\small{}-16.15 } & \multicolumn{2}{c}{{\small{}-16.74}} & \multicolumn{2}{c}{{\small{}-15.56}}\tabularnewline
{\small{}3 } & {\small{}-16.8 } & \multicolumn{2}{c}{{\small{}-17.4}} & \multicolumn{2}{c}{{\small{}-16.21}}\tabularnewline
{\small{}2 } & {\small{}-31.58 } & \multicolumn{2}{c}{{\small{}-32.16}} & \multicolumn{2}{c}{{\small{}-31.01}}\tabularnewline
{\small{}$*$ } & {\small{}-11.78 } & \multicolumn{2}{c}{{\small{}-12.37}} & \multicolumn{2}{c}{{\small{}-11.18}}\tabularnewline
\end{tabular}
\end{table}
{\small\par}

{\small{}{} }
\begin{table}
{\small{}\centering{}{}\caption{\textbf{Tuning of $k$-best $\protect\alphabeta$ with Child Batching
and $10$ seconds per action (all games).}\label{tab:Tuning-k-best-10s-with-child-batching-part-2}}
}%
\begin{tabular}{cccccccccccc}
\hline 
 & {\tiny{}{}Amazons}{\small{} } & {\tiny{}{}Arimaa}{\small{} } & {\tiny{}{}Ataxx}{\small{} } & {\tiny{}{}Breaktrhough}{\small{} } & {\tiny{}{}Brazilian}{\small{} } & {\tiny{}{}Canadian}{\small{} } & {\tiny{}{}Clobber}{\small{} } & {\tiny{}{}Connect6}{\small{} } & {\tiny{}{}International}{\small{} } & {\tiny{}{}Chess}{\small{} } & {\tiny{}{}Go 9}\tabularnewline
\hline 
{\tiny{}{}$\infty$}{\small{} } & {\tiny{}{}-60 $\pm$ 2}{\small{} } & {\tiny{}{}-35 $\pm$ 2}{\small{} } & {\tiny{}{}-89 $\pm$ 2}{\small{} } & {\tiny{}{}-41 $\pm$ 3}{\small{} } & {\tiny{}{}4 $\pm$ 1}{\small{} } & {\tiny{}{}4 $\pm$ 3}{\small{} } & {\tiny{}{}-42 $\pm$ 3}{\small{} } & {\tiny{}{}-77 $\pm$ 3}{\small{} } & {\tiny{}{}9 $\pm$ 2}{\small{} } & {\tiny{}{}-36 $\pm$ 2}{\small{} } & {\tiny{}{}2 $\pm$ 3}\tabularnewline
{\tiny{}{}30}{\small{} } & {\tiny{}{}-61 $\pm$ 2}{\small{} } & {\tiny{}{}-36 $\pm$ 2}{\small{} } & {\tiny{}{}-80 $\pm$ 3}{\small{} } & {\tiny{}{}-38 $\pm$ 3}{\small{} } & {\tiny{}{}4 $\pm$ 1}{\small{} } & {\tiny{}{}8 $\pm$ 3}{\small{} } & {\tiny{}{}-30 $\pm$ 3}{\small{} } & {\tiny{}{}-78 $\pm$ 3}{\small{} } & {\tiny{}{}7 $\pm$ 2}{\small{} } & {\tiny{}{}-32 $\pm$ 2}{\small{} } & {\tiny{}{}4 $\pm$ 3}\tabularnewline
{\tiny{}{}15}{\small{} } & {\tiny{}{}-48 $\pm$ 3}{\small{} } & {\tiny{}{}-30 $\pm$ 2}{\small{} } & {\tiny{}{}-78 $\pm$ 3}{\small{} } & {\tiny{}{}-34 $\pm$ 3}{\small{} } & {\tiny{}{}2 $\pm$ 1}{\small{} } & {\tiny{}{}6 $\pm$ 3}{\small{} } & {\tiny{}{}-19 $\pm$ 3}{\small{} } & {\tiny{}{}-75 $\pm$ 3}{\small{} } & {\tiny{}{}8 $\pm$ 1}{\small{} } & {\tiny{}{}-8 $\pm$ 3}{\small{} } & {\tiny{}{}10 $\pm$ 3}\tabularnewline
\end{tabular}
\end{table}
{\small\par}

{\small{}{} }
\begin{table}
{\small{}\centering{}{}\caption{\textbf{Tuning of $\protect\mcts{}$ without Child Batching and $10$
seconds per action.}\label{tab:Tuning-MCTS-10s-without-child-batching}}
}%
\begin{tabular}{cccccccccccc}
\hline 
 &  &  & {\small{}$C$ } & {\tiny{}{}mean}{\small{} } & \multicolumn{2}{c}{{\tiny{}{}lower bound}} & \multicolumn{2}{c}{{\tiny{}{}upper bound}} &  &  & \tabularnewline
\hline 
 &  &  & {\small{}$\sqrt{2}$ } & {\small{}-94.01 } & \multicolumn{2}{c}{{\small{}-94.15}} & \multicolumn{2}{c}{{\small{}-93.86}} &  &  & \tabularnewline
 &  &  & {\small{}1 } & {\small{}-93.55 } & \multicolumn{2}{c}{{\small{}-93.7}} & \multicolumn{2}{c}{{\small{}-93.4}} &  &  & \tabularnewline
 &  &  & {\small{}0.3 } & {\small{}-94.13 } & \multicolumn{2}{c}{{\small{}-94.26}} & \multicolumn{2}{c}{{\small{}-93.99}} &  &  & \tabularnewline
 &  &  & {\small{}0.1 } & {\small{}-95.4 } & \multicolumn{2}{c}{{\small{}-95.51}} & \multicolumn{2}{c}{{\small{}-95.3}} &  &  & \tabularnewline
 &  &  & {\small{}0.03 } & {\small{}-95.61 } & \multicolumn{2}{c}{{\small{}-95.71}} & \multicolumn{2}{c}{{\small{}-95.51}} &  &  & \tabularnewline
 &  &  & {\small{}$*$ } & {\small{}-93.55 } & \multicolumn{2}{c}{{\small{}-93.38}} & \multicolumn{2}{c}{{\small{}-93.08}} &  &  & \tabularnewline
\hline 
{\tiny{}{}$C$}{\small{} } & {\tiny{}{}Amazons}{\small{} } & {\tiny{}{}Arimaa}{\small{} } & {\tiny{}{}Ataxx}{\small{} } & {\tiny{}{}Breaktrhough}{\small{} } & {\tiny{}{}Brazilian}{\small{} } & {\tiny{}{}Canadian}{\small{} } & {\tiny{}{}Clobber}{\small{} } & {\tiny{}{}Connect6}{\small{} } & {\tiny{}{}International}{\small{} } & {\tiny{}{}Chess}{\small{} } & {\tiny{}{}Go 9}\tabularnewline
\end{tabular}
\end{table}
{\small\par}

{\small{}{} }
\begin{table}
{\small{}\centering{}{}\caption{\textbf{Tuning of $\protect\mctsh{}$ without Child Batching and $10$
seconds per action (mean).}\label{tab:Tuning-MCTSh-10s-without-child-batching-part-1}}
}%
\begin{tabular}{ccc||cc||c}
\hline 
{\small{}$C$ } & {\small{}mean } & \multicolumn{2}{c}{{\small{}lower bound}} & \multicolumn{2}{c}{{\small{}upper bound}}\tabularnewline
\hline 
{\small{}$\sqrt{2}$ } & {\small{}-95.74 } & \multicolumn{2}{c}{{\small{}-95.87}} & \multicolumn{2}{c}{{\small{}-95.61}}\tabularnewline
{\small{}1 } & {\small{}-95.53 } & \multicolumn{2}{c}{{\small{}-95.66}} & \multicolumn{2}{c}{{\small{}-95.39}}\tabularnewline
{\small{}0.3 } & {\small{}-93.87 } & \multicolumn{2}{c}{{\small{}-94.04,}} & \multicolumn{2}{c}{{\small{}-93.7}}\tabularnewline
{\small{}0.1 } & {\small{}-89.86 } & \multicolumn{2}{c}{{\small{}-90.07}} & \multicolumn{2}{c}{{\small{}-89.65}}\tabularnewline
{\small{}0.03 } & {\small{}-84.87 } & \multicolumn{2}{c}{{\small{}-85.14}} & \multicolumn{2}{c}{{\small{}-84.59}}\tabularnewline
{\small{}0.01 } & {\small{}-80.75 } & \multicolumn{2}{c}{{\small{}-81.08}} & \multicolumn{2}{c}{{\small{}-80.43}}\tabularnewline
{\small{}0.003 } & {\small{}-79.09 } & \multicolumn{2}{c}{{\small{}-79.43}} & \multicolumn{2}{c}{{\small{}-78.75}}\tabularnewline
{\small{}0.0003 } & {\small{}-78.82 } & \multicolumn{2}{c}{{\small{}-79.16}} & \multicolumn{2}{c}{{\small{}-78.49}}\tabularnewline
{\small{}$*$ } & {\small{}-77.38 } & \multicolumn{2}{c}{{\small{}-77.72}} & \multicolumn{2}{c}{{\small{}-77.03}}\tabularnewline
\end{tabular}
\end{table}
{\small\par}

{\small{}{} }
\begin{table}
{\small{}\centering{}{}\caption{\textbf{Tuning of $\protect\mctsh{}$ without Child Batching and $10$
seconds per action (all games).}\label{tab:Tuning-MCTSh-10s-without-child-batching-part-2}}
}%
\begin{tabular}{cccccccccccc}
\hline 
{\tiny{}{}$C$}{\small{} } & {\tiny{}{}Amazons}{\small{} } & {\tiny{}{}Arimaa}{\small{} } & {\tiny{}{}Ataxx}{\small{} } & {\tiny{}{}Breaktrhough}{\small{} } & {\tiny{}{}Brazilian}{\small{} } & {\tiny{}{}Canadian}{\small{} } & {\tiny{}{}Clobber}{\small{} } & {\tiny{}{}Connect6}{\small{} } & {\tiny{}{}International}{\small{} } & {\tiny{}{}Chess}{\small{} } & {\tiny{}{}Go 9}\tabularnewline
\hline 
{\tiny{}{}$\sqrt{2}$}{\small{} } & {\tiny{}{}-99 $\pm$ 0}{\small{} } & {\tiny{}{}-46 $\pm$ 1}{\small{} } & {\tiny{}{}-100 $\pm$ 0}{\small{} } & {\tiny{}{}-100 $\pm$ 0}{\small{} } & {\tiny{}{}-94 $\pm$ 0}{\small{} } & {\tiny{}{}-99 $\pm$ 0}{\small{} } & {\tiny{}{}-95 $\pm$ 1}{\small{} } & {\tiny{}{}-99 $\pm$ 0}{\small{} } & {\tiny{}{}-98 $\pm$ 0}{\small{} } & {\tiny{}{}-99 $\pm$ 0}{\small{} } & {\tiny{}{}-86 $\pm$ 1}\tabularnewline
{\tiny{}{}1}{\small{} } & {\tiny{}{}-99 $\pm$ 0}{\small{} } & {\tiny{}{}-46 $\pm$ 1}{\small{} } & {\tiny{}{}-100 $\pm$ 0}{\small{} } & {\tiny{}{}-100 $\pm$ 0}{\small{} } & {\tiny{}{}-93 $\pm$ 0}{\small{} } & {\tiny{}{}-99 $\pm$ 0}{\small{} } & {\tiny{}{}-94 $\pm$ 1}{\small{} } & {\tiny{}{}-99 $\pm$ 0}{\small{} } & {\tiny{}{}-98 $\pm$ 0}{\small{} } & {\tiny{}{}-99 $\pm$ 0}{\small{} } & {\tiny{}{}-85 $\pm$ 1}\tabularnewline
{\tiny{}{}0.3}{\small{} } & {\tiny{}{}-99 $\pm$ 0}{\small{} } & {\tiny{}{}-47 $\pm$ 1}{\small{} } & {\tiny{}{}-100 $\pm$ 0}{\small{} } & {\tiny{}{}-100 $\pm$ 0}{\small{} } & {\tiny{}{}-74 $\pm$ 1}{\small{} } & {\tiny{}{}-97 $\pm$ 1}{\small{} } & {\tiny{}{}-93 $\pm$ 1}{\small{} } & {\tiny{}{}-99 $\pm$ 0}{\small{} } & {\tiny{}{}-93 $\pm$ 0}{\small{} } & {\tiny{}{}-99 $\pm$ 0}{\small{} } & {\tiny{}{}-79 $\pm$ 2}\tabularnewline
\end{tabular}
\end{table}
{\small\par}

{\small{}{} }
\begin{table}
{\small{}\centering{}{}\caption{\textbf{Tuning of $\protect\pvs{}$ with Child Batching and $1$ seconds
per action.}\label{tab:Tuning-PVS-1s-with-child-batching}}
}%
\begin{tabular}{cccccccccccc}
\hline 
 &  &  & {\small{}$\delta$ } & {\small{}mean } & \multicolumn{2}{c}{{\small{}lower bound}} & \multicolumn{2}{c}{{\small{}upper bound}} &  &  & \tabularnewline
\hline 
 &  &  & {\small{}3000 } & {\small{}-48.86 } & \multicolumn{2}{c}{{\small{}-49.38}} & \multicolumn{2}{c}{{\small{}-48.33}} &  &  & \tabularnewline
 &  &  & {\small{}1000 } & {\small{}-46.49 } & \multicolumn{2}{c}{{\small{}-47.02}} & \multicolumn{2}{c}{{\small{}-45.96}} &  &  & \tabularnewline
 &  &  & {\small{}300 } & {\small{}-44.15 } & \multicolumn{2}{c}{{\small{}-44.7}} & \multicolumn{2}{c}{{\small{}-43.6}} &  &  & \tabularnewline
 &  &  & {\small{}100 } & {\small{}-42.7 } & \multicolumn{2}{c}{{\small{}-43.24}} & \multicolumn{2}{c}{{\small{}-42.16}} &  &  & \tabularnewline
 &  &  & {\small{}30 } & {\small{}-46.28 } & \multicolumn{2}{c}{{\small{}-46.81}} & \multicolumn{2}{c}{{\small{}-45.75}} &  &  & \tabularnewline
 &  &  & {\small{}10 } & {\small{}-49.38 } & \multicolumn{2}{c}{{\small{}-49.91}} & \multicolumn{2}{c}{{\small{}-48.86}} &  &  & \tabularnewline
 &  &  & {\small{}$*$ } & {\small{}-42.13 } & \multicolumn{2}{c}{{\small{}-42.68}} & \multicolumn{2}{c}{{\small{}-41.59}} &  &  & \tabularnewline
\end{tabular}
\end{table}
{\small\par}

{\small{}{} }
\begin{table}
{\small{}\centering{}{}\caption{\textbf{Tuning of $\protect\mtdf{}$ with Child Batching and $1$
seconds per action.}\label{tab:Tuning-MTDf-1s-with-child-batching}}
}%
\begin{tabular}{cccccccccccc}
\hline 
 &  &  & {\small{}$\delta$ } & {\small{}mean } & \multicolumn{2}{c}{{\small{}lower bound}} & \multicolumn{2}{c}{{\small{}upper bound}} &  &  & \tabularnewline
\hline 
 &  &  & {\small{}10 } & {\small{}-50.39 } & \multicolumn{2}{c}{{\small{}-50.9}} & \multicolumn{2}{c}{{\small{}-49.86}} &  &  & \tabularnewline
 &  &  & {\small{}30 } & {\small{}-48.64 } & \multicolumn{2}{c}{{\small{}-49.16}} & \multicolumn{2}{c}{{\small{}-48.12}} &  &  & \tabularnewline
 &  &  & {\small{}100 } & {\small{}-48.14 } & \multicolumn{2}{c}{{\small{}-48.66}} & \multicolumn{2}{c}{{\small{}-47.62}} &  &  & \tabularnewline
 &  &  & {\small{}300 } & {\small{}-46.11 } & \multicolumn{2}{c}{{\small{}-46.65}} & \multicolumn{2}{c}{{\small{}-45.57}} &  &  & \tabularnewline
 &  &  & {\small{}1000 } & {\small{}-46.18 } & \multicolumn{2}{c}{{\small{}-46.71}} & \multicolumn{2}{c}{{\small{}-45.65}} &  &  & \tabularnewline
 &  &  & {\small{}3000 } & {\small{}-47.62 } & \multicolumn{2}{c}{{\small{}-48.14}} & \multicolumn{2}{c}{{\small{}-47.09}} &  &  & \tabularnewline
 &  &  & {\small{}$*$ } & {\small{}-44.35 } & \multicolumn{2}{c}{{\small{}-44.89}} & \multicolumn{2}{c}{{\small{}-43.82}} &  &  & \tabularnewline
\end{tabular}
\end{table}
{\small\par}

{\small{}{} }
\begin{table}
{\small{}\centering{}{}\caption{\textbf{Tuning of $k$-best $\protect\alphabeta$ with Child Batching
and $1$ seconds per action (mean).}\label{tab:Tuning-k-best-1s-with-child-batching-part-1}}
}%
\begin{tabular}{ccc||cc||c}
\hline 
{\small{}$k$ } & {\small{}mean } & \multicolumn{2}{c}{{\small{}lower bound}} & \multicolumn{2}{c}{{\small{}upper bound}}\tabularnewline
\hline 
{\small{}2 } & {\small{}-16.17 } & \multicolumn{2}{c}{{\small{}-16.78}} & \multicolumn{2}{c}{{\small{}-15.56}}\tabularnewline
{\small{}3 } & {\small{}-17.05 } & \multicolumn{2}{c}{{\small{}-17.66}} & \multicolumn{2}{c}{{\small{}-16.45}}\tabularnewline
{\small{}4 } & {\small{}-20.07 } & \multicolumn{2}{c}{{\small{}-20.67}} & \multicolumn{2}{c}{{\small{}-19.46}}\tabularnewline
{\small{}5 } & {\small{}-23.4 } & \multicolumn{2}{c}{{\small{}-24}} & \multicolumn{2}{c}{{\small{}-22.81}}\tabularnewline
{\small{}6 } & {\small{}-27.77 } & \multicolumn{2}{c}{{\small{}-28.36}} & \multicolumn{2}{c}{{\small{}-27.18}}\tabularnewline
{\small{}7 } & {\small{}-30.02 } & \multicolumn{2}{c}{{\small{}-30.6}} & \multicolumn{2}{c}{{\small{}-29.45}}\tabularnewline
{\small{}8 } & {\small{}-31.14 } & \multicolumn{2}{c}{{\small{}-31.73}} & \multicolumn{2}{c}{{\small{}-30.55}}\tabularnewline
\end{tabular}
\end{table}
{\small\par}

{\small{}{} }
\begin{table}
{\small{}\centering{}{}\caption{\textbf{Tuning of $k$-best $\protect\alphabeta$ with Child Batching
and $1$ seconds per action (all games: part I).}\label{tab:Tuning-k-best-1s-with-child-batching-part-2}}
}%
\begin{tabular}{cccccccccccc}
\hline 
{\tiny{}{}$k$}{\small{} } & {\tiny{}{}Amazons}{\small{} } & {\tiny{}{}Arimaa}{\small{} } & {\tiny{}{}Ataxx}{\small{} } & {\tiny{}{}Breaktrhough}{\small{} } & {\tiny{}{}Brazilian}{\small{} } & {\tiny{}{}Canadian}{\small{} } & {\tiny{}{}Clobber}{\small{} } & {\tiny{}{}Connect6}{\small{} } & {\tiny{}{}International}{\small{} } & {\tiny{}{}Chess}{\small{} } & {\tiny{}{}Go 9}\tabularnewline
\hline 
{\tiny{}{}2}{\small{} } & {\tiny{}{}-16 $\pm$ 3}{\small{} } & {\tiny{}{}-12 $\pm$ 2}{\small{} } & {\tiny{}{}-27 $\pm$ 4}{\small{} } & {\tiny{}{}-9 $\pm$ 3}{\small{} } & {\tiny{}{}-3 $\pm$ 1}{\small{} } & {\tiny{}{}-16 $\pm$ 3}{\small{} } & {\tiny{}{}-6 $\pm$ 3}{\small{} } & {\tiny{}{}-39 $\pm$ 5}{\small{} } & {\tiny{}{}-8 $\pm$ 1}{\small{} } & {\tiny{}{}-18 $\pm$ 3}{\small{} } & {\tiny{}{}-1 $\pm$ 3}\tabularnewline
{\tiny{}{}3}{\small{} } & {\tiny{}{}-15 $\pm$ 3}{\small{} } & {\tiny{}{}-12 $\pm$ 2}{\small{} } & {\tiny{}{}-40 $\pm$ 4}{\small{} } & {\tiny{}{}-15 $\pm$ 3}{\small{} } & {\tiny{}{}-4 $\pm$ 1}{\small{} } & {\tiny{}{}-6 $\pm$ 3}{\small{} } & {\tiny{}{}-2 $\pm$ 3}{\small{} } & {\tiny{}{}-42 $\pm$ 5}{\small{} } & {\tiny{}{}-3 $\pm$ 2}{\small{} } & {\tiny{}{}-13 $\pm$ 3}{\small{} } & {\tiny{}{}6 $\pm$ 3}\tabularnewline
{\tiny{}{}4}{\small{} } & {\tiny{}{}-20 $\pm$ 3}{\small{} } & {\tiny{}{}-13 $\pm$ 2}{\small{} } & {\tiny{}{}-47 $\pm$ 4}{\small{} } & {\tiny{}{}-17 $\pm$ 3}{\small{} } & {\tiny{}{}-7 $\pm$ 1}{\small{} } & {\tiny{}{}-9 $\pm$ 3}{\small{} } & {\tiny{}{}-6 $\pm$ 3}{\small{} } & {\tiny{}{}-46 $\pm$ 4}{\small{} } & {\tiny{}{}-6 $\pm$ 2}{\small{} } & {\tiny{}{}-13 $\pm$ 3}{\small{} } & {\tiny{}{}10 $\pm$ 3}\tabularnewline
\end{tabular}
\end{table}
{\small\par}

{\small{}{} }
\begin{table}
{\small{}\centering{}{}\caption{\textbf{Tuning of $k$-best $\protect\alphabeta$ with Child Batching
and $1$ seconds per action (all games: part II).}\label{tab:Tuning-k-best-1s-with-child-batching-part-3}}
}%
\begin{tabular}{cccccccccccc}
\hline 
{\tiny{}{}$k$}{\small{} } & {\tiny{}{}Gomoku}{\small{} } & {\tiny{}{}Havannah}{\small{} } & {\tiny{}{}Hex 11}{\small{} } & {\tiny{}{}Hex 13}{\small{} } & {\tiny{}{}Hex 19}{\small{} } & {\tiny{}{}Lines of A.}{\small{} } & {\tiny{}{}Othello 10}{\small{} } & {\tiny{}{}Othello 8}{\small{} } & {\tiny{}{}Santorini}{\small{} } & {\tiny{}{}Surakarta}{\small{} } & {\tiny{}{}Xiangqi}\tabularnewline
\hline 
{\tiny{}{}2}{\small{} } & {\tiny{}{}-41 $\pm$ 3}{\small{} } & {\tiny{}{}-6 $\pm$ 3}{\small{} } & {\tiny{}{}-10 $\pm$ 3}{\small{} } & {\tiny{}{}-20 $\pm$ 3}{\small{} } & {\tiny{}{}-18 $\pm$ 3}{\small{} } & {\tiny{}{}-17 $\pm$ 3}{\small{} } & {\tiny{}{}-35 $\pm$ 3}{\small{} } & {\tiny{}{}-20 $\pm$ 3}{\small{} } & {\tiny{}{}-10 $\pm$ 5}{\small{} } & {\tiny{}{}-4 $\pm$ 3}{\small{} } & {\tiny{}{}-22 $\pm$ 3}\tabularnewline
{\tiny{}{}3}{\small{} } & {\tiny{}{}-47 $\pm$ 3}{\small{} } & {\tiny{}{}-6 $\pm$ 3}{\small{} } & {\tiny{}{}-15 $\pm$ 3}{\small{} } & {\tiny{}{}-24 $\pm$ 3}{\small{} } & {\tiny{}{}-17 $\pm$ 3}{\small{} } & {\tiny{}{}-22 $\pm$ 3}{\small{} } & {\tiny{}{}-40 $\pm$ 3}{\small{} } & {\tiny{}{}-25 $\pm$ 3}{\small{} } & {\tiny{}{}-19 $\pm$ 5}{\small{} } & {\tiny{}{}-6 $\pm$ 3}{\small{} } & {\tiny{}{}-19 $\pm$ 3}\tabularnewline
{\tiny{}{}4}{\small{} } & {\tiny{}{}-50 $\pm$ 2}{\small{} } & {\tiny{}{}-9 $\pm$ 3}{\small{} } & {\tiny{}{}-22 $\pm$ 3}{\small{} } & {\tiny{}{}-29 $\pm$ 3}{\small{} } & {\tiny{}{}-17 $\pm$ 3}{\small{} } & {\tiny{}{}-24 $\pm$ 3}{\small{} } & {\tiny{}{}-49 $\pm$ 2}{\small{} } & {\tiny{}{}-30 $\pm$ 3}{\small{} } & {\tiny{}{}-33 $\pm$ 5}{\small{} } & {\tiny{}{}-5 $\pm$ 3}{\small{} } & {\tiny{}{}-18 $\pm$ 3}\tabularnewline
\end{tabular}
\end{table}
{\small\par}

{\small{}{} }
\begin{table}
{\small{}\centering{}{}\caption{\textbf{Tuning of $\protect\mcts{}$ without Child Batching and $1$
seconds per action (mean).}\label{tab:Tuning-MCTS-1s-without-child-batching-part-1}}
}%
\begin{tabular}{ccc||cc||c}
\hline 
{\small{}$C$ } & {\small{}mean } & \multicolumn{2}{c}{{\small{}lower bound}} & \multicolumn{2}{c}{{\small{}upper bound}}\tabularnewline
\hline 
{\small{}0.0001 } & {\small{}-95.64 } & \multicolumn{2}{c}{{\small{}-95.73}} & \multicolumn{2}{c}{{\small{}-95.55}}\tabularnewline
{\small{}0.0003 } & {\small{}-95.68 } & \multicolumn{2}{c}{{\small{}-95.77}} & \multicolumn{2}{c}{{\small{}-95.59}}\tabularnewline
{\small{}0.001 } & {\small{}-95.65 } & \multicolumn{2}{c}{{\small{}-95.74}} & \multicolumn{2}{c}{{\small{}-95.56}}\tabularnewline
{\small{}0.003 } & {\small{}-95.61 } & \multicolumn{2}{c}{{\small{}-95.71}} & \multicolumn{2}{c}{{\small{}-95.52}}\tabularnewline
{\small{}0.01 } & {\small{}-95.6 } & \multicolumn{2}{c}{{\small{}-95.69}} & \multicolumn{2}{c}{{\small{}-95.51}}\tabularnewline
{\small{}0.03 } & {\small{}-95.66 } & \multicolumn{2}{c}{{\small{}-95.75}} & \multicolumn{2}{c}{{\small{}-95.57}}\tabularnewline
{\small{}0.1 } & {\small{}-95.53 } & \multicolumn{2}{c}{{\small{}-95.62}} & \multicolumn{2}{c}{{\small{}-95.43}}\tabularnewline
{\small{}0.3 } & {\small{}-95.35 } & \multicolumn{2}{c}{{\small{}-95.45}} & \multicolumn{2}{c}{{\small{}-95.25}}\tabularnewline
{\small{}1 } & {\small{}-95.62 } & \multicolumn{2}{c}{{\small{}-95.71}} & \multicolumn{2}{c}{{\small{}-95.52}}\tabularnewline
{\small{}$\sqrt{2}$ } & {\small{}-95.76 } & \multicolumn{2}{c}{{\small{}-95.85}} & \multicolumn{2}{c}{{\small{}-95.66}}\tabularnewline
{\small{}$*$ } & {\small{}-95.13 } & \multicolumn{2}{c}{{\small{}-95.23}} & \multicolumn{2}{c}{{\small{}-95.02}}\tabularnewline
\end{tabular}
\end{table}
{\small\par}

{\small{}{} }
\begin{table}
{\small{}\centering{}{}\caption{\textbf{Tuning of $\protect\mcts{}$ without Child Batching and $1$
seconds per action (all games).}\label{tab:Tuning-MCTS-1s-without-child-batching-part-2}}
}%
\begin{tabular}{cccccccccccc}
\hline 
{\tiny{}{}$C$}{\small{} } & {\tiny{}{}Amazons}{\small{} } & {\tiny{}{}Arimaa}{\small{} } & {\tiny{}{}Ataxx}{\small{} } & {\tiny{}{}Breaktrhough}{\small{} } & {\tiny{}{}Brazilian}{\small{} } & {\tiny{}{}Canadian}{\small{} } & {\tiny{}{}Clobber}{\small{} } & {\tiny{}{}Connect6}{\small{} } & {\tiny{}{}International}{\small{} } & {\tiny{}{}Chess}{\small{} } & {\tiny{}{}Go 9}\tabularnewline
\hline 
{\tiny{}{}0.0001}{\small{} } & {\tiny{}{}-100 $\pm$ 0}{\small{} } & {\tiny{}{}-24 $\pm$ 1}{\small{} } & {\tiny{}{}-100 $\pm$ 0}{\small{} } & {\tiny{}{}-100 $\pm$ 0}{\small{} } & {\tiny{}{}-98 $\pm$ 0}{\small{} } & {\tiny{}{}-99 $\pm$ 0}{\small{} } & {\tiny{}{}-100 $\pm$ 0}{\small{} } & {\tiny{}{}-98 $\pm$ 1}{\small{} } & {\tiny{}{}-99 $\pm$ 0}{\small{} } & {\tiny{}{}-99 $\pm$ 0}{\small{} } & {\tiny{}{}-99 $\pm$ 0}\tabularnewline
{\tiny{}{}0.0003}{\small{} } & {\tiny{}{}-100 $\pm$ 0}{\small{} } & {\tiny{}{}-24 $\pm$ 1}{\small{} } & {\tiny{}{}-100 $\pm$ 0}{\small{} } & {\tiny{}{}-100 $\pm$ 0}{\small{} } & {\tiny{}{}-98 $\pm$ 0}{\small{} } & {\tiny{}{}-99 $\pm$ 0}{\small{} } & {\tiny{}{}-100 $\pm$ 0}{\small{} } & {\tiny{}{}-98 $\pm$ 1}{\small{} } & {\tiny{}{}-99 $\pm$ 0}{\small{} } & {\tiny{}{}-99 $\pm$ 0}{\small{} } & {\tiny{}{}-99 $\pm$ 0}\tabularnewline
{\tiny{}{}0.001}{\small{} } & {\tiny{}{}-100 $\pm$ 0}{\small{} } & {\tiny{}{}-23 $\pm$ 1}{\small{} } & {\tiny{}{}-100 $\pm$ 0}{\small{} } & {\tiny{}{}-100 $\pm$ 0}{\small{} } & {\tiny{}{}-98 $\pm$ 0}{\small{} } & {\tiny{}{}-100 $\pm$ 0}{\small{} } & {\tiny{}{}-99 $\pm$ 0}{\small{} } & {\tiny{}{}-96 $\pm$ 1}{\small{} } & {\tiny{}{}-99 $\pm$ 0}{\small{} } & {\tiny{}{}-99 $\pm$ 0}{\small{} } & {\tiny{}{}-99 $\pm$ 0}\tabularnewline
\end{tabular}
\end{table}
{\small\par}

{\small{}{} }
\begin{table}
{\small{}\centering{}{}\caption{\textbf{Tuning of $\protect\mctsh{}$ without Child Batching and $1$
seconds per action (mean).}\label{tab:Tuning-MCTSh-1s-without-child-batching-part-1}}
}%
\begin{tabular}{ccc||cc||c}
\hline 
{\small{}$C$ } & {\small{}mean } & \multicolumn{2}{c}{{\small{}lower bound}} & \multicolumn{2}{c}{{\small{}upper bound}}\tabularnewline
\hline 
{\small{}0 } & {\small{}-79.4 } & \multicolumn{2}{c}{{\small{}-79.75}} & \multicolumn{2}{c}{{\small{}-79.05}}\tabularnewline
{\small{}0.00001 } & {\small{}-80.05 } & \multicolumn{2}{c}{{\small{}-80.39}} & \multicolumn{2}{c}{{\small{}-79.71}}\tabularnewline
{\small{}0.00003 } & {\small{}-80.03 } & \multicolumn{2}{c}{{\small{}-80.37}} & \multicolumn{2}{c}{{\small{}-79.69}}\tabularnewline
{\small{}0.0001 } & {\small{}-79.85 } & \multicolumn{2}{c}{{\small{}-80.19}} & \multicolumn{2}{c}{{\small{}-79.5}}\tabularnewline
{\small{}0.0003 } & {\small{}-80.14 } & \multicolumn{2}{c}{{\small{}-80.48}} & \multicolumn{2}{c}{{\small{}-79.8}}\tabularnewline
{\small{}0.001 } & {\small{}-80.31 } & \multicolumn{2}{c}{{\small{}-80.64}} & \multicolumn{2}{c}{{\small{}-79.97}}\tabularnewline
{\small{}0.003 } & {\small{}-80.46 } & \multicolumn{2}{c}{{\small{}-80.79}} & \multicolumn{2}{c}{{\small{}-80.12}}\tabularnewline
\end{tabular}
\end{table}
{\small\par}

{\small{}{} }
\begin{table}
{\small{}\centering{}{}\caption{\textbf{Tuning of $\protect\mctsh{}$ without Child Batching and $1$
seconds per action (all games: part I).}\label{tab:Tuning-MCTSh-1s-without-child-batching-part-2}}
}%
%
\end{table}
{\small\par}

{\small{}{} }
\begin{table}
{\small{}\centering{}{}\caption{\textbf{Tuning of $\protect\mctsh{}$ without Child Batching and $1$
seconds per action (all games: part II).}\label{tab:Tuning-MCTSh-1s-without-child-batching-part-3}}
}%
%
\end{table}
{\small\par}

{\small{}{} }{\small\par}

{\small{}\clearpage}{\small\par}

{\small{}\printcredits}{\small\par}

{\small{}%% Loading bibliography style file

{\small{}% Loading bibliography database
 \bibliographystyle{cas-model2-names}
\bibliography{science_BF}

\begin{thebibliography}{50}
\expandafter\ifx\csname natexlab\endcsname\relax\def\natexlab#1{#1}\fi
\providecommand{\url}[1]{\texttt{#1}}
\providecommand{\href}[2]{#2}
\providecommand{\path}[1]{#1}
\providecommand{\DOIprefix}{doi:}
\providecommand{\ArXivprefix}{arXiv:}
\providecommand{\URLprefix}{URL: }
\providecommand{\Pubmedprefix}{pmid:}
\providecommand{\doi}[1]{\href{http://dx.doi.org/#1}{\path{#1}}}
\providecommand{\Pubmed}[1]{\href{pmid:#1}{\path{#1}}}
\providecommand{\bibinfo}[2]{#2}
\ifx\xfnm\relax \def\xfnm[#1]{\unskip,\space#1}\fi
%Type = Article
\bibitem[{Allis et~al.(1994)Allis, van~der Meulen and Van
  Den~Herik}]{allis1994proof}
\bibinfo{author}{Allis, L.V.}, \bibinfo{author}{van~der Meulen, M.},
  \bibinfo{author}{Van Den~Herik, H.J.}, \bibinfo{year}{1994}.
\newblock \bibinfo{title}{Proof-number search}.
\newblock \bibinfo{journal}{Artificial Intelligence} \bibinfo{volume}{66},
  \bibinfo{pages}{91--124}.
%Type = Article
\bibitem[{Baier and Winands(2018)}]{baier2018mcts}
\bibinfo{author}{Baier, H.}, \bibinfo{author}{Winands, M.H.},
  \bibinfo{year}{2018}.
\newblock \bibinfo{title}{Mcts-minimax hybrids with state evaluations}.
\newblock \bibinfo{journal}{Journal of Artificial Intelligence Research}
  \bibinfo{volume}{62}, \bibinfo{pages}{193--231}.
%Type = Book
\bibitem[{Berlekamp et~al.(2003)Berlekamp, Conway and
  Guy}]{berlekamp2018winning}
\bibinfo{author}{Berlekamp, E.R.}, \bibinfo{author}{Conway, J.H.},
  \bibinfo{author}{Guy, R.K.}, \bibinfo{year}{2003}.
\newblock \bibinfo{title}{Winning Ways for Your Mathematical Plays, Volume 3}.
\newblock \bibinfo{publisher}{AK Peters/CRC Press}.
%Type = Incollection
\bibitem[{Berliner(1981)}]{berliner1981b}
\bibinfo{author}{Berliner, H.}, \bibinfo{year}{1981}.
\newblock \bibinfo{title}{The b* tree search algorithm: A best-first proof
  procedure}, in: \bibinfo{booktitle}{Readings in Artificial Intelligence}.
  \bibinfo{publisher}{Elsevier}, pp. \bibinfo{pages}{79--87}.
%Type = Article
\bibitem[{Bonnet et~al.(2016)Bonnet, Jamain and
  Saffidine}]{bonnet2016complexity}
\bibinfo{author}{Bonnet, {\'E}.}, \bibinfo{author}{Jamain, F.},
  \bibinfo{author}{Saffidine, A.}, \bibinfo{year}{2016}.
\newblock \bibinfo{title}{On the complexity of connection games}.
\newblock \bibinfo{journal}{Theoretical Computer Science}
  \bibinfo{volume}{644}, \bibinfo{pages}{2--28}.
%Type = Article
\bibitem[{Bouzy et~al.(2020)Bouzy, Cazenave, Corruble and
  Teytaud}]{bouzy2020artificial}
\bibinfo{author}{Bouzy, B.}, \bibinfo{author}{Cazenave, T.},
  \bibinfo{author}{Corruble, V.}, \bibinfo{author}{Teytaud, O.},
  \bibinfo{year}{2020}.
\newblock \bibinfo{title}{Artificial intelligence for games}.
\newblock \bibinfo{journal}{A Guided Tour of Artificial Intelligence Research:
  Volume II: AI Algorithms} , \bibinfo{pages}{313--337}.
%Type = Article
\bibitem[{Browne et~al.(2012)Browne, Powley, Whitehouse, Lucas, Cowling,
  Rohlfshagen, Tavener, Perez, Samothrakis and Colton}]{browne2012survey}
\bibinfo{author}{Browne, C.B.}, \bibinfo{author}{Powley, E.},
  \bibinfo{author}{Whitehouse, D.}, \bibinfo{author}{Lucas, S.M.},
  \bibinfo{author}{Cowling, P.I.}, \bibinfo{author}{Rohlfshagen, P.},
  \bibinfo{author}{Tavener, S.}, \bibinfo{author}{Perez, D.},
  \bibinfo{author}{Samothrakis, S.}, \bibinfo{author}{Colton, S.},
  \bibinfo{year}{2012}.
\newblock \bibinfo{title}{A survey of monte carlo tree search methods}.
\newblock \bibinfo{journal}{Transactions on Computational Intelligence and AI
  in games} \bibinfo{volume}{4}, \bibinfo{pages}{1--43}.
%Type = Article
\bibitem[{Campbell et~al.(2002)Campbell, Hoane~Jr and Hsu}]{campbell2002deep}
\bibinfo{author}{Campbell, M.}, \bibinfo{author}{Hoane~Jr, A.J.},
  \bibinfo{author}{Hsu, F.h.}, \bibinfo{year}{2002}.
\newblock \bibinfo{title}{Deep blue}.
\newblock \bibinfo{journal}{Artificial Intelligence} \bibinfo{volume}{134},
  \bibinfo{pages}{57--83}.
%Type = Inproceedings
\bibitem[{Cazenave(2021)}]{cazenave2021monte}
\bibinfo{author}{Cazenave, T.}, \bibinfo{year}{2021}.
\newblock \bibinfo{title}{Monte carlo game solver}, in:
  \bibinfo{booktitle}{Monte Carlo Search: First Workshop, MCS 2020, Held in
  Conjunction with IJCAI 2020, Virtual Event, January 7, 2021, Proceedings 1},
  \bibinfo{organization}{Springer}. pp. \bibinfo{pages}{56--70}.
%Type = Inproceedings
\bibitem[{Chaslot et~al.(2008)Chaslot, Winands and van
  Den~Herik}]{chaslot2008parallel}
\bibinfo{author}{Chaslot, G.M.B.}, \bibinfo{author}{Winands, M.H.},
  \bibinfo{author}{van Den~Herik, H.J.}, \bibinfo{year}{2008}.
\newblock \bibinfo{title}{Parallel monte-carlo tree search}, in:
  \bibinfo{booktitle}{Computers and Games: 6th International Conference, CG
  2008, Beijing, China, September 29-October 1, 2008. Proceedings 6},
  \bibinfo{organization}{Springer}. pp. \bibinfo{pages}{60--71}.
%Type = Inproceedings
\bibitem[{Clark and Storkey(2015)}]{clark2015training}
\bibinfo{author}{Clark, C.}, \bibinfo{author}{Storkey, A.},
  \bibinfo{year}{2015}.
\newblock \bibinfo{title}{Training deep convolutional neural networks to play
  go}, in: \bibinfo{booktitle}{International Conference on Machine Learning},
  pp. \bibinfo{pages}{1766--1774}.
%Type = Article
\bibitem[{Cohen-Solal(2020)}]{2020learning}
\bibinfo{author}{Cohen-Solal, Q.}, \bibinfo{year}{2020}.
\newblock \bibinfo{title}{Learning to play two-player perfect-information games
  without knowledge}.
\newblock \bibinfo{journal}{arXiv preprint arXiv:2008.01188} .
%Type = Article
\bibitem[{Cohen-Solal(2021)}]{cohen2021completeness}
\bibinfo{author}{Cohen-Solal, Q.}, \bibinfo{year}{2021}.
\newblock \bibinfo{title}{Completeness of unbounded best-first game
  algorithms}.
\newblock \bibinfo{journal}{arXiv preprint arXiv:2109.09468} .
%Type = Article
\bibitem[{Cohen-Solal and Cazenave(2021)}]{cohen2021descent}
\bibinfo{author}{Cohen-Solal, Q.}, \bibinfo{author}{Cazenave, T.},
  \bibinfo{year}{2021}.
\newblock \bibinfo{title}{Descent wins five gold medals at the computer
  olympiad}.
\newblock \bibinfo{journal}{ICGA Journal} \bibinfo{volume}{43},
  \bibinfo{pages}{132--134}.
%Type = Article
\bibitem[{Cohen-Solal and Cazenave(2023a)}]{cohen2023athenan}
\bibinfo{author}{Cohen-Solal, Q.}, \bibinfo{author}{Cazenave, T.},
  \bibinfo{year}{2023}a.
\newblock \bibinfo{title}{Athenan wins sixteen gold medals at the computer
  olympiad}.
\newblock \bibinfo{journal}{ICGA Journal} \bibinfo{volume}{45}.
%Type = Article
\bibitem[{Cohen-Solal and Cazenave(2023b)}]{cohen2023minimax}
\bibinfo{author}{Cohen-Solal, Q.}, \bibinfo{author}{Cazenave, T.},
  \bibinfo{year}{2023}b.
\newblock \bibinfo{title}{Minimax strikes back}.
\newblock \bibinfo{journal}{AAMAS} .
%Type = Inproceedings
\bibitem[{Coulom(2007)}]{Coulom06}
\bibinfo{author}{Coulom, R.}, \bibinfo{year}{2007}.
\newblock \bibinfo{title}{Efficient selectivity and backup operators in
  monte-carlo tree search}, in: \bibinfo{booktitle}{Computers and Games, 5th
  International Conference, {CG} 2006, Turin, Italy, May 29-31, 2006. Revised
  Papers}, pp. \bibinfo{pages}{72--83}.
%Type = Article
\bibitem[{Fawzi et~al.(2022)Fawzi, Balog, Huang, Hubert, Romera-Paredes,
  Barekatain, Novikov, R~Ruiz, Schrittwieser, Swirszcz
  et~al.}]{fawzi2022discovering}
\bibinfo{author}{Fawzi, A.}, \bibinfo{author}{Balog, M.},
  \bibinfo{author}{Huang, A.}, \bibinfo{author}{Hubert, T.},
  \bibinfo{author}{Romera-Paredes, B.}, \bibinfo{author}{Barekatain, M.},
  \bibinfo{author}{Novikov, A.}, \bibinfo{author}{R~Ruiz, F.J.},
  \bibinfo{author}{Schrittwieser, J.}, \bibinfo{author}{Swirszcz, G.}, et~al.,
  \bibinfo{year}{2022}.
\newblock \bibinfo{title}{Discovering faster matrix multiplication algorithms
  with reinforcement learning}.
\newblock \bibinfo{journal}{Nature} \bibinfo{volume}{610},
  \bibinfo{pages}{47--53}.
%Type = Article
\bibitem[{Fink(1982)}]{fink1982enhancement}
\bibinfo{author}{Fink, W.}, \bibinfo{year}{1982}.
\newblock \bibinfo{title}{An enhancement to the iterative, alpha-beta, minimax
  search procedure}.
\newblock \bibinfo{journal}{ICGA Journal} \bibinfo{volume}{5},
  \bibinfo{pages}{34--35}.
%Type = Article
\bibitem[{Finkel and Fishburn(1982)}]{finkel1982parallelism}
\bibinfo{author}{Finkel, R.A.}, \bibinfo{author}{Fishburn, J.P.},
  \bibinfo{year}{1982}.
\newblock \bibinfo{title}{Parallelism in alpha-beta search}.
\newblock \bibinfo{journal}{Artificial Intelligence} \bibinfo{volume}{19},
  \bibinfo{pages}{89--106}.
%Type = Incollection
\bibitem[{Greenblatt et~al.(1988)Greenblatt, Eastlake and
  Crocker}]{greenblatt1988greenblatt}
\bibinfo{author}{Greenblatt, R.D.}, \bibinfo{author}{Eastlake, D.E.},
  \bibinfo{author}{Crocker, S.D.}, \bibinfo{year}{1988}.
\newblock \bibinfo{title}{The greenblatt chess program}, in:
  \bibinfo{booktitle}{Computer chess compendium}.
  \bibinfo{publisher}{Springer}, pp. \bibinfo{pages}{56--66}.
%Type = Article
\bibitem[{Guo et~al.(2024)Guo, Yu, Li, Zhang, Wang, Li and
  Dong}]{guo2024retrosynthesis}
\bibinfo{author}{Guo, J.}, \bibinfo{author}{Yu, C.}, \bibinfo{author}{Li, K.},
  \bibinfo{author}{Zhang, Y.}, \bibinfo{author}{Wang, G.}, \bibinfo{author}{Li,
  S.}, \bibinfo{author}{Dong, H.}, \bibinfo{year}{2024}.
\newblock \bibinfo{title}{Retrosynthesis zero: Self-improving global synthesis
  planning using reinforcement learning}.
\newblock \bibinfo{journal}{Journal of Chemical Theory and Computation} .
%Type = Book
\bibitem[{Hsu(1989)}]{hsu1989large}
\bibinfo{author}{Hsu, F.H.}, \bibinfo{year}{1989}.
\newblock \bibinfo{title}{Large-scale parallelization of alpha-beta search: An
  algorithmic and architectural study with computer chess}.
\newblock \bibinfo{publisher}{Carnegie Mellon University}.
%Type = Article
\bibitem[{Kaliszyk et~al.(2018)Kaliszyk, Urban, Michalewski and
  Ol{\v{s}}{\'a}k}]{kaliszyk2018reinforcement}
\bibinfo{author}{Kaliszyk, C.}, \bibinfo{author}{Urban, J.},
  \bibinfo{author}{Michalewski, H.}, \bibinfo{author}{Ol{\v{s}}{\'a}k, M.},
  \bibinfo{year}{2018}.
\newblock \bibinfo{title}{Reinforcement learning of theorem proving}.
\newblock \bibinfo{journal}{Advances in Neural Information Processing Systems}
  \bibinfo{volume}{31}.
%Type = Article
\bibitem[{Kishimoto et~al.(2012)Kishimoto, Winands, M{\"u}ller and
  Saito}]{kishimoto2012game}
\bibinfo{author}{Kishimoto, A.}, \bibinfo{author}{Winands, M.H.},
  \bibinfo{author}{M{\"u}ller, M.}, \bibinfo{author}{Saito, J.T.},
  \bibinfo{year}{2012}.
\newblock \bibinfo{title}{Game-tree search using proof numbers: The first
  twenty years}.
\newblock \bibinfo{journal}{ICGA journal} \bibinfo{volume}{35},
  \bibinfo{pages}{131--156}.
%Type = Article
\bibitem[{Knuth and Moore(1975)}]{knuth1975analysis}
\bibinfo{author}{Knuth, D.E.}, \bibinfo{author}{Moore, R.W.},
  \bibinfo{year}{1975}.
\newblock \bibinfo{title}{An analysis of alpha-beta pruning}.
\newblock \bibinfo{journal}{Artificial Intelligence} \bibinfo{volume}{6},
  \bibinfo{pages}{293--326}.
%Type = Article
\bibitem[{Korf(1985)}]{korf1985depth}
\bibinfo{author}{Korf, R.E.}, \bibinfo{year}{1985}.
\newblock \bibinfo{title}{Depth-first iterative-deepening: An optimal
  admissible tree search}.
\newblock \bibinfo{journal}{Artificial Intelligence} \bibinfo{volume}{27},
  \bibinfo{pages}{97--109}.
%Type = Inproceedings
\bibitem[{Korf and Chickering(1994)}]{korf1994best}
\bibinfo{author}{Korf, R.E.}, \bibinfo{author}{Chickering, D.M.},
  \bibinfo{year}{1994}.
\newblock \bibinfo{title}{Best-first minimax search: Othello results}, in:
  \bibinfo{booktitle}{AAAI}, pp. \bibinfo{pages}{1365--1370}.
%Type = Article
\bibitem[{Korf and Chickering(1996)}]{korf1996best}
\bibinfo{author}{Korf, R.E.}, \bibinfo{author}{Chickering, D.M.},
  \bibinfo{year}{1996}.
\newblock \bibinfo{title}{Best-first minimax search}.
\newblock \bibinfo{journal}{Artificial intelligence} \bibinfo{volume}{84},
  \bibinfo{pages}{299--337}.
%Type = Book
\bibitem[{Mandziuk(2010)}]{mandziuk2010knowledge}
\bibinfo{author}{Mandziuk, J.}, \bibinfo{year}{2010}.
\newblock \bibinfo{title}{Knowledge-free and learning-based methods in
  intelligent game playing}. volume \bibinfo{volume}{276}.
\newblock \bibinfo{publisher}{Springer}.
%Type = Article
\bibitem[{Ma{\'n}dziuk and Osman(2004)}]{mandziuk2004alpha}
\bibinfo{author}{Ma{\'n}dziuk, J.}, \bibinfo{author}{Osman, D.},
  \bibinfo{year}{2004}.
\newblock \bibinfo{title}{Alpha-beta search enhancements with a real-value
  game-state evaluation function}.
\newblock \bibinfo{journal}{ICGA Journal} \bibinfo{volume}{27},
  \bibinfo{pages}{38--43}.
%Type = Article
\bibitem[{Mankowitz et~al.(2023)Mankowitz, Michi, Zhernov, Gelmi, Selvi,
  Paduraru, Leurent, Iqbal, Lespiau, Ahern et~al.}]{mankowitz2023faster}
\bibinfo{author}{Mankowitz, D.J.}, \bibinfo{author}{Michi, A.},
  \bibinfo{author}{Zhernov, A.}, \bibinfo{author}{Gelmi, M.},
  \bibinfo{author}{Selvi, M.}, \bibinfo{author}{Paduraru, C.},
  \bibinfo{author}{Leurent, E.}, \bibinfo{author}{Iqbal, S.},
  \bibinfo{author}{Lespiau, J.B.}, \bibinfo{author}{Ahern, A.}, et~al.,
  \bibinfo{year}{2023}.
\newblock \bibinfo{title}{Faster sorting algorithms discovered using deep
  reinforcement learning}.
\newblock \bibinfo{journal}{Nature} \bibinfo{volume}{618},
  \bibinfo{pages}{257--263}.
%Type = Article
\bibitem[{Marsland and Campbell(1982)}]{marsland1982parallel}
\bibinfo{author}{Marsland, T.A.}, \bibinfo{author}{Campbell, M.},
  \bibinfo{year}{1982}.
\newblock \bibinfo{title}{Parallel search of strongly ordered game trees}.
\newblock \bibinfo{journal}{ACM Computing Surveys (CSUR)} \bibinfo{volume}{14},
  \bibinfo{pages}{533--551}.
%Type = Article
\bibitem[{McAllester(1988)}]{mcallester1988conspiracy}
\bibinfo{author}{McAllester, D.A.}, \bibinfo{year}{1988}.
\newblock \bibinfo{title}{Conspiracy numbers for min-max search}.
\newblock \bibinfo{journal}{Artificial Intelligence} \bibinfo{volume}{35},
  \bibinfo{pages}{287--310}.
%Type = Book
\bibitem[{Millington and Funge(2009)}]{millington2009artificial}
\bibinfo{author}{Millington, I.}, \bibinfo{author}{Funge, J.},
  \bibinfo{year}{2009}.
\newblock \bibinfo{title}{Artificial intelligence for games}.
\newblock \bibinfo{publisher}{CRC Press}.
%Type = Inproceedings
\bibitem[{Pearl(1980)}]{pearl1980scout}
\bibinfo{author}{Pearl, J.}, \bibinfo{year}{1980}.
\newblock \bibinfo{title}{Scout: A simple game-searching algorithm with proven
  optimal properties.}, in: \bibinfo{booktitle}{AAAI}, pp.
  \bibinfo{pages}{143--145}.
%Type = Incollection
\bibitem[{Pearl(2022)}]{pearl2022asymptotic}
\bibinfo{author}{Pearl, J.}, \bibinfo{year}{2022}.
\newblock \bibinfo{title}{Asymptotic properties of minimax trees and
  game-searching procedures}, in: \bibinfo{booktitle}{Probabilistic and Causal
  Inference: The Works of Judea Pearl}, pp. \bibinfo{pages}{61--90}.
%Type = Inproceedings
\bibitem[{Piette et~al.(2020)Piette, Soemers, Stephenson, Sironi, Winands and
  Browne}]{Piette2020Ludii}
\bibinfo{author}{Piette, {\'E}.}, \bibinfo{author}{Soemers, D.J.N.J.},
  \bibinfo{author}{Stephenson, M.}, \bibinfo{author}{Sironi, C.F.},
  \bibinfo{author}{Winands, M.H.M.}, \bibinfo{author}{Browne, C.},
  \bibinfo{year}{2020}.
\newblock \bibinfo{title}{Ludii -- the ludemic general game system}, in:
  \bibinfo{editor}{Giacomo, G.D.}, \bibinfo{editor}{Catala, A.},
  \bibinfo{editor}{Dilkina, B.}, \bibinfo{editor}{Milano, M.},
  \bibinfo{editor}{Barro, S.}, \bibinfo{editor}{Bugarín, A.},
  \bibinfo{editor}{Lang, J.} (Eds.), \bibinfo{booktitle}{Proceedings of the
  24th European Conference on Artificial Intelligence (ECAI 2020)},
  \bibinfo{publisher}{IOS Press}. pp. \bibinfo{pages}{411--418}.
%Type = Article
\bibitem[{Plaat et~al.(1996)Plaat, Schaeffer, Pijls and
  De~Bruin}]{plaat1996best}
\bibinfo{author}{Plaat, A.}, \bibinfo{author}{Schaeffer, J.},
  \bibinfo{author}{Pijls, W.}, \bibinfo{author}{De~Bruin, A.},
  \bibinfo{year}{1996}.
\newblock \bibinfo{title}{Best-first fixed-depth minimax algorithms}.
\newblock \bibinfo{journal}{Artificial Intelligence} \bibinfo{volume}{87},
  \bibinfo{pages}{255--293}.
%Type = Article
\bibitem[{Powley et~al.(1990)Powley, Ferguson and Korf}]{powley1990parallel}
\bibinfo{author}{Powley, C.}, \bibinfo{author}{Ferguson, C.},
  \bibinfo{author}{Korf, R.E.}, \bibinfo{year}{1990}.
\newblock \bibinfo{title}{Parallel heuristic search: Two approaches}.
\newblock \bibinfo{journal}{Parallel algorithms for machine intelligence and
  vision} , \bibinfo{pages}{42--65}.
%Type = Inproceedings
\bibitem[{Ramanujan and Selman(2011)}]{ramanujan2011trade}
\bibinfo{author}{Ramanujan, R.}, \bibinfo{author}{Selman, B.},
  \bibinfo{year}{2011}.
\newblock \bibinfo{title}{Trade-offs in sampling-based adversarial planning.},
  in: \bibinfo{booktitle}{ICAPS}, pp. \bibinfo{pages}{202--209}.
%Type = Article
\bibitem[{Schaeffer(1990)}]{schaeffer1990conspiracy}
\bibinfo{author}{Schaeffer, J.}, \bibinfo{year}{1990}.
\newblock \bibinfo{title}{Conspiracy numbers}.
\newblock \bibinfo{journal}{Artificial Intelligence} \bibinfo{volume}{43},
  \bibinfo{pages}{67--84}.
%Type = Article
\bibitem[{Shoham and Toledo(2002)}]{shoham2002parallel}
\bibinfo{author}{Shoham, Y.}, \bibinfo{author}{Toledo, S.},
  \bibinfo{year}{2002}.
\newblock \bibinfo{title}{Parallel randomized best-first minimax search}.
\newblock \bibinfo{journal}{Artificial Intelligence} \bibinfo{volume}{137},
  \bibinfo{pages}{165--196}.
%Type = Article
\bibitem[{Silver et~al.(2016)Silver, Huang, Maddison, Guez, Sifre, Van
  Den~Driessche, Schrittwieser, Antonoglou, Panneershelvam, Lanctot
  et~al.}]{silver2016mastering}
\bibinfo{author}{Silver, D.}, \bibinfo{author}{Huang, A.},
  \bibinfo{author}{Maddison, C.J.}, \bibinfo{author}{Guez, A.},
  \bibinfo{author}{Sifre, L.}, \bibinfo{author}{Van Den~Driessche, G.},
  \bibinfo{author}{Schrittwieser, J.}, \bibinfo{author}{Antonoglou, I.},
  \bibinfo{author}{Panneershelvam, V.}, \bibinfo{author}{Lanctot, M.}, et~al.,
  \bibinfo{year}{2016}.
\newblock \bibinfo{title}{Mastering the game of go with deep neural networks
  and tree search}.
\newblock \bibinfo{journal}{Nature} \bibinfo{volume}{529},
  \bibinfo{pages}{484}.
%Type = Article
\bibitem[{Silver et~al.(2018)Silver, Hubert, Schrittwieser, Antonoglou, Lai,
  Guez, Lanctot, Sifre, Kumaran, Graepel et~al.}]{silver2018general}
\bibinfo{author}{Silver, D.}, \bibinfo{author}{Hubert, T.},
  \bibinfo{author}{Schrittwieser, J.}, \bibinfo{author}{Antonoglou, I.},
  \bibinfo{author}{Lai, M.}, \bibinfo{author}{Guez, A.},
  \bibinfo{author}{Lanctot, M.}, \bibinfo{author}{Sifre, L.},
  \bibinfo{author}{Kumaran, D.}, \bibinfo{author}{Graepel, T.}, et~al.,
  \bibinfo{year}{2018}.
\newblock \bibinfo{title}{A general reinforcement learning algorithm that
  masters chess, shogi, and go through self-play}.
\newblock \bibinfo{journal}{Science} \bibinfo{volume}{362},
  \bibinfo{pages}{1140--1144}.
%Type = Article
\bibitem[{Silver et~al.(2017)Silver, Schrittwieser, Simonyan, Antonoglou,
  Huang, Guez, Hubert, Baker, Lai, Bolton et~al.}]{silver2017mastering}
\bibinfo{author}{Silver, D.}, \bibinfo{author}{Schrittwieser, J.},
  \bibinfo{author}{Simonyan, K.}, \bibinfo{author}{Antonoglou, I.},
  \bibinfo{author}{Huang, A.}, \bibinfo{author}{Guez, A.},
  \bibinfo{author}{Hubert, T.}, \bibinfo{author}{Baker, L.},
  \bibinfo{author}{Lai, M.}, \bibinfo{author}{Bolton, A.}, et~al.,
  \bibinfo{year}{2017}.
\newblock \bibinfo{title}{Mastering the game of go without human knowledge}.
\newblock \bibinfo{journal}{Nature} \bibinfo{volume}{550},
  \bibinfo{pages}{354}.
%Type = Inproceedings
\bibitem[{Singhal and Sridevi(2019)}]{singhal2019comparative}
\bibinfo{author}{Singhal, S.P.}, \bibinfo{author}{Sridevi, M.},
  \bibinfo{year}{2019}.
\newblock \bibinfo{title}{Comparative study of performance of parallel alpha
  beta pruning for different architectures}, in: \bibinfo{booktitle}{2019 IEEE
  9th international conference on advanced computing (IACC)},
  \bibinfo{organization}{IEEE}. pp. \bibinfo{pages}{115--119}.
%Type = Article
\bibitem[{Tesauro(1995)}]{tesauro1995temporal}
\bibinfo{author}{Tesauro, G.}, \bibinfo{year}{1995}.
\newblock \bibinfo{title}{Temporal difference learning and td-gammon}.
\newblock \bibinfo{journal}{Communications of the ACM} \bibinfo{volume}{38},
  \bibinfo{pages}{58--68}.
%Type = Article
\bibitem[{Van Den~Herik et~al.(2002)Van Den~Herik, Uiterwijk and
  Van~Rijswijck}]{van2002games}
\bibinfo{author}{Van Den~Herik, H.J.}, \bibinfo{author}{Uiterwijk, J.W.},
  \bibinfo{author}{Van~Rijswijck, J.}, \bibinfo{year}{2002}.
\newblock \bibinfo{title}{Games solved: Now and in the future}.
\newblock \bibinfo{journal}{Artificial Intelligence} \bibinfo{volume}{134},
  \bibinfo{pages}{277--311}.
%Type = Inproceedings
\bibitem[{Winands et~al.(2008)Winands, Bj{\"o}rnsson and
  Saito}]{winands2008monte}
\bibinfo{author}{Winands, M.H.}, \bibinfo{author}{Bj{\"o}rnsson, Y.},
  \bibinfo{author}{Saito, J.T.}, \bibinfo{year}{2008}.
\newblock \bibinfo{title}{Monte-carlo tree search solver}, in:
  \bibinfo{booktitle}{International Conference on Computers and Games},
  \bibinfo{organization}{Springer}. pp. \bibinfo{pages}{25--36}.

\end{thebibliography}
}{\small\par}

{\small{}%\vskip3pt}{\small\par}

{\small{}%\bio{}
%Author biography without author photo.
%Author biography. Author biography. Author biography.
%Author biography. Author biography. Author biography.
%Author biography. Author biography. Author biography.
%Author biography. Author biography. Author biography.
%Author biography. Author biography. Author biography.
%Author biography. Author biography. Author biography.
%Author biography. Author biography. Author biography.
%Author biography. Author biography. Author biography.
%Author biography. Author biography. Author biography.
%\endbio}{\small\par}

{\small{}%\bio{figs/cas-pic1}
%Author biography with author photo.
%Author biography. Author biography. Author biography.
%Author biography. Author biography. Author biography.
%Author biography. Author biography. Author biography.
%Author biography. Author biography. Author biography.
%Author biography. Author biography. Author biography.
%Author biography. Author biography. Author biography.
%Author biography. Author biography. Author biography.
%Author biography. Author biography. Author biography.
%Author biography. Author biography. Author biography.
%\endbio}{\small\par}

{\small{}%\vskip3pc}{\small\par}

{\small{}%\bio{figs/cas-pic1}
%Author biography with author photo.
%Author biography. Author biography. Author biography.
%Author biography. Author biography. Author biography.
%Author biography. Author biography. Author biography.
%Author biography. Author biography. Author biography.
%\endbio
{document} }{\small\par}
\end{document}